\documentclass{fairmeta}

\usepackage{amsmath}
\usepackage{amssymb}
\usepackage{url}
\usepackage{hyperref}
\usepackage{booktabs}
\usepackage{multirow}
\usepackage{graphicx}   
\usepackage{wasysym}    
\usepackage{array}
\usepackage{tabularx}
\usepackage{xcolor}

\AtBeginDocument{\hbadness=10000\hfuzz=200pt}
\title{{\setstretch{1.08}\fontsize{17}{20}\selectfont MagicSim: A Unified Infrastructure for Executable Embodied Interaction\par\vspace{0.38em}{\fontsize{11}{13}\selectfont\mdseries One Runtime, One MDP, Three Drivers}}}

\author[1,*,\dag]{Haoran Lu}
\author[5,6*]{Songling Liu}
\author[2,*]{Yue Chen}
\author[1]{Guo Ye}
\author[1]{Mutian Shen}
\author[1]{Shuyang Yu}
\author[1]{Yu Xiao}
\author[1]{Shang Wu}
\author[1]{Jiayi Wang}
\author[1]{Jianshu Zhang}
\author[1]{Jihai Zhao}
\author[4]{Xiangtian Gui}
\author[3]{Chuye Hong}
\author[2]{Yuran Wang}
\author[1]{Maojiang Su}
\author[2,3]{Ruihai Wu}
\author[1]{Zhaoran Wang}
\author[1,\dag]{Han Liu}

\small{
\affiliation[1]{Northwestern~University}
\affiliation[2]{Peking~University}
\affiliation[3]{University~of~California, Berkeley}
\affiliation[4]{ShanghaiTech~University}
\affiliation[5]{Institute~of~Automation,~Chinese~Academy~of~Sciences,~Beijing~100190}
\affiliation[6]{University~of~Chinese~Academy~of~Sciences}
}

\footnotesize{
\contribution[*]{Equal Contribution}
\contribution[\dag]{Corresponding Author}
}

\usepackage{booktabs}
\usepackage{multirow}
\usepackage{array}
\usepackage{graphicx}
\usepackage{adjustbox}
\usepackage{pifont}            
\usepackage{wasysym}           
\usepackage[table]{xcolor}
\usepackage{placeins}

\definecolor{cmGreen}{RGB}{0,150,70}
\definecolor{cmOrange}{RGB}{230,126,20}
\definecolor{cmGray}{RGB}{150,150,150}
\definecolor{cmRowGreen}{RGB}{222,244,229}
\definecolor{cmRowGray}{RGB}{238,238,238}
\newcommand{\Y}{\textcolor{cmGreen}{\ding{51}}}      
\newcommand{\Pt}{\textcolor{cmOrange}{\LEFTcircle}}  
\newcommand{\N}{\textcolor{cmGray}{\ding{55}}}       

\newcolumntype{R}[2]{%
  >{\adjustbox{angle=#1,lap=\width-(#2)}\bgroup}%
  l%
  <{\egroup}%
}
\newcommand*\rot{\multicolumn{1}{R{70}{1em}}}

\abstract{
Robot learning and embodied agents now require simulation to serve as a shared
execution substrate linking control, skills, and planning, not only as a
renderer, controller testbed, or fixed task environment. Existing pipelines
split these layers with ``magic'' actions, disconnected training environments,
or forward-only renders that cannot reproduce, evaluate, and annotate the same
episode. We present MagicSim, an embodied interaction infrastructure built
around one deterministic batched runtime and a shared Markov decision process
(MDP). From YAML-first specifications that decouple contents, placement,
behavior, and agent exposure, MagicSim constructs diverse executable worlds
spanning task families, interaction regimes, physics, layouts, sensors, avatars,
and robot embodiments in one reset-and-step loop. A common execution interface
grounds high-level commands through controllers, atomicskills, planner
primitives, and asynchronous planning, realizing them as robot actions rather
than simulator-side state edits. One task definition supports three capabilities:
benchmark and RL evaluation, an autocollect interface that automatically turns
commands into grounded trajectories, and agent/VLM-facing interaction. For
automatic execution, commands flow through a
Command$\rightarrow$Skill$\rightarrow$Planner$\rightarrow$Robot$\rightarrow$Record
pipeline, while per-environment command, skill, planning, retry, annotation, and
episode states advance independently above the shared physics tick. Successful
rollouts are saved as structured multimodal trajectories aligning language
supervision, action representations, visual/geometric representations, and
task-level status with the executed episode. MagicSim thus unifies diverse world
construction, embodied execution, task evaluation, automatic rollout generation,
and interactive agent interfaces in one planner-in-the-loop runtime.
}

\begin{document}

\maketitle

\begin{figure}[!htb] \centering \includegraphics[width=0.98\linewidth]{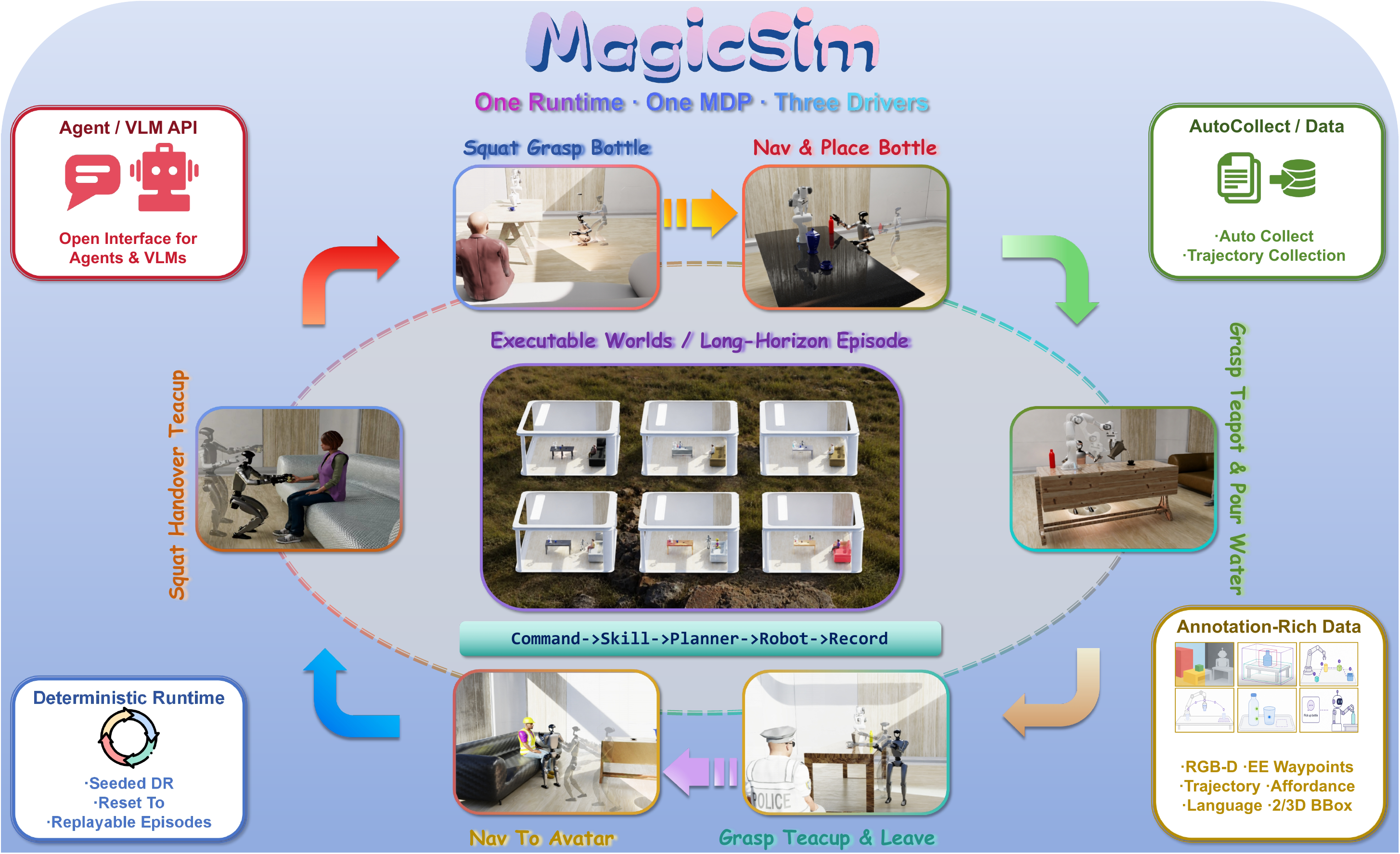} \caption[MagicSim at a glance.]{ \textbf{MagicSim at a glance.} A deterministic batched runtime executes planner-grounded, long-horizon episodes through a shared Markov decision process. Each interaction follows the Command-to-Skill-to-Planner-to-Robot-to-Record pipeline and can be exposed through three drivers: benchmark and reinforcement-learning evaluation, success-gated trajectory collection, and agent/VLM interaction. The same execution produces synchronized multimodal supervision, including RGB-D observations, language traces, end-effector waypoints, object bounding boxes, and trajectory affordances. } \label{fig:magicsim-teaser} \end{figure}
\section{Introduction}
\label{sec:introduction}

\subsection{From Simulators to Embodied Interaction Infrastructure}

Recent progress in robot learning~\cite{Intelligence2026pi07AS,Intelligence202505AV,generalist2025gen0}, vision-language-action (VLA) and world-action models~\cite{Nvidia2025GR00TNA,Ye2026WorldAM,yuan2026fastwamworldactionmodels,Ye2025LearningTF}, and embodied agents~\cite{Wang2025VAGENRW,Wang2025RAGENUS,Fung2025EmbodiedAA} has changed the role of simulation. A simulator is no longer only a place to test controllers, render scenes, or define task environments~\cite{todorov2012mujoco,makoviychuk2021isaacgym,xiang2020sapien}. It is increasingly expected to serve as a shared substrate across different levels of embodied intelligence: low-level robot learning needs RL-ready environments and successful demonstrations~\cite{chen2026learningpartawaredense3d,Shen2025BiAssembleLC,Li2024BroadcastingSR,11128651}; mid-level systems need grounded skill abstractions; and high-level agents need an interactive world where language commands, plans, observations, and consequences unfold over time.

This shift matters because these research levels are often studied in isolation. Low-level robotics focuses on control, manipulation, locomotion, contact, and reinforcement learning, while high-level agent research focuses on planning, task decomposition, language-conditioned decision making, and social interaction. Without a shared execution layer, high-level agents are often evaluated through artificial ``magic'' actions, and low-level policies are trained in environments disconnected from the plans that should invoke them. MagicSim connects these levels by placing them on one shared, replayable runtime and a single MDP. On top of that runtime, the same task can serve as an RL-style benchmark, be driven by planner-backed atomicskills for data collection, or be opened to agents whose language commands invoke those same grounded skills.

Consider one end-to-end demonstration in MagicSim: a Unitree G1 humanoid walks across a room, crouches to pick up a bottle, carries it to a side table, pours water from a teapot, places the objects onto a tray, carries the tray to a seated person, and hands it over Figure.\ref{fig:magicsim-teaser} This single episode combines navigation, whole-body balance, rigid-object manipulation, fluid interaction, placement, carrying, and human-facing interaction in one runnable scene. Rendering such a sequence is not enough. To make this episode reusable for training, evaluation, or agent interaction, the system must reproduce it when the pour goes wrong, plan the crouch and the placement inside the stepping loop rather than offline, tie an affordance label to where the robot actually grasped, and keep the trajectory only when the task succeeds. A render pipeline can produce the \emph{pictures} of this episode, but not these interaction-dependent records, because each requires state that a forward-only pipeline discards.

MagicSim is built around this view. Given a scene specification, a task specification, and an embodiment, it constructs a randomized but replayable world, exposes it through a shared task interface, drives behavior through policies, planners, and atomicskills, captures multimodal observations and runtime annotations, and returns the episode as a benchmark rollout, a successful training trajectory, or an agent-facing interaction. MagicSim is therefore not merely a physics simulator, renderer, task benchmark, or data pipeline. It is an embodied interaction infrastructure that serves as a shared substrate across the control, skill, and planning levels of the embodied stack.

\subsection{What Makes Embodied Interaction Executable?}

The gap above cannot be closed by a richer asset library or a larger task list alone.\cite{garmentlab,UniGarmentAU, geng2025roboverse} The hard part is making several system pressures coexist inside one running simulator. An embodied interaction substrate must support heterogeneous physics, deterministic replay, multiple robot embodiments, planner-in-the-loop control, and supervision that captures what happened during interaction rather than only what was rendered.

\textbf{Heterogeneous physical worlds.}
A long-horizon task may involve rigid bodies, articulated objects, particle fluids, deformables, terrains, sensors, avatars, and robots in the same episode. These elements have different simulation backends and lifecycle constraints, yet they must coexist inside one reset and step loop. The simulator must therefore define a common contract for what exists in the world, where it is placed, how it behaves, and how it can be acted on.

\textbf{Deterministic large-scale runtime.}
Executable worlds must also be reproducible at scale. In MagicSim, many sub-environments share one simulator instance: one agent may be navigating, another may be waiting for a planner result, another may be manipulating an object, and another may have failed and reset. Randomization is useful only if these parallel worlds can be reset, snapshotted, and replayed from the same initial condition. Deterministic reset ordering, state snapshotting, disjoint seeded randomization streams, and \texttt{reset\_to} replay are therefore part of the runtime contract rather than post-hoc logging utilities.

\textbf{Multi-embodiment execution.}
A shared substrate cannot assume a single robot form or action space. Low-level robotics may require single-arm manipulation, dual-arm coordination, mobile manipulation, dexterous hands, humanoid loco-manipulation, quadruped locomotion, or camera-only motion. These embodiments differ in kinematics, controllers, sensors, and action spaces, but they must still be exposed through a consistent task and execution interface so that benchmark policies, scripted collectors, and agent drivers can operate over the same world distribution.

\textbf{Planner-in-the-loop interaction.}
Long-horizon embodied tasks cannot be reduced to offline trajectories or instantaneous state edits. Skills such as crouching to grasp, placing an object on a tray, navigating around furniture, or handing an object to a person require IK, motion generation, local navigation, and controller feedback inside the stepping loop. This is especially challenging in batched simulation: one slow planning call should not freeze every other environment. An executable runtime must therefore let skills submit planner requests, continue stepping the batch, and consume results when they are ready.

\textbf{Annotation-rich data generation.}
Finally, the system must preserve the interaction structure that learning and evaluation need. A rendered frame alone does not say where the robot actually grasped, which skill phase produced an action, which planner target was selected, whether the task succeeded, or how the scene should be described in language. Executable interaction requires a capture layer that records observations, actions, task states, skill and planner traces, object grounding, affordance targets, end-effector waypoints, language, and success-gated trajectories.

Together, these challenges define MagicSim's premise: one batched reproducible runtime, one shared MDP, and multiple drivers for benchmarking, data collection, and agent interaction.

\subsection{Design Philosophy}

MagicSim is designed as usable infrastructure rather than a fixed benchmark or a closed data-generation script. Its first principle is configuration over code: scenes, layouts, assets, robots, cameras, tasks, planners, skills, randomization, and recording options are exposed through structured YAML-style specifications. This lets researchers build and modify environments without rewriting simulator internals, and also makes the system easier for coding agents to inspect, generate, and edit programmatically.

Usability is paired with efficiency as a runtime property. MagicSim runs many sub-environments inside one simulator instance, while each environment maintains its own state and lifecycle. Expensive operations such as IK and motion planning are handled asynchronously: a skill can submit a planning request, the batched simulation continues stepping, and the result is consumed when ready. This design keeps the hot path from blocking on a single slow solve and treats time and compute efficiency as part of how the environment runs, not as an optimization added after the fact.

Finally, MagicSim pursues breadth without detaching abstraction from physical grounding. The system is designed to support diverse physical interactions, sensors, avatars, robot embodiments, and task families, so the same framework can serve robot learning, data generation, and agent-facing use cases. At the same time, high-level skills are exposed only insofar as they can be realized through planners, controllers, and robot actions inside the simulator. The result is an extensible embodied interaction framework that is easy to specify, efficient to run, broad in coverage, and grounded in stateful simulation.

\subsection{Contributions}

This report presents MagicSim as an embodied interaction infrastructure that connects configurable world construction, reproducible simulation, grounded robot execution, planner-in-the-loop interaction, structured annotation, and multiple research drivers within one framework. Its main contributions are:

\begin{enumerate}
    \item \textbf{A configurable substrate for heterogeneous physical worlds.}
    MagicSim provides a YAML-first environment interface that decouples what exists, where it is placed, and how it behaves. Scenes can combine diverse object families, articulated structures, deformables, fluids, terrains, rooms, sensors, avatars, and robot embodiments under one specification contract, allowing users to define not only what a world contains, but also what can be acted on, sensed, randomized, and evaluated.

    \item \textbf{A deterministic large-scale batched runtime.}
    MagicSim organizes simulation through a manager-based runtime in which many sub-environments share one simulator instance while maintaining independent per-environment state. The runtime supports deterministic reset ordering, seeded domain randomization with disjoint streams, state snapshotting, \texttt{reset\_to} replay, and dynamic object lifecycle handling across P1/P2/P3 modes. This makes reproducibility a runtime contract for randomized, heterogeneous embodied worlds.

    \item \textbf{A multi-embodiment execution stack.}
    MagicSim connects robot actions, closed-loop controllers, navigation, humanoid whole-body control, dexterous control, avatar interaction, and camera motion through a common execution interface. This allows the same runtime to support single-arm robots, dual-arm systems, mobile manipulators, dexterous hands, humanoids, quadrupeds, avatars, and camera-only agents without reducing them to a single action model.

    \item \textbf{Planner-in-the-loop interaction over one MDP and three drivers.}
    MagicSim exposes tasks as a shared MDP with observations, actions, rewards, terminations, and success conditions. The same Gym-compatible task can be used as an RL benchmark, driven through atomicskills and planners for scripted data collection, or exposed through agent-facing interfaces where language-level commands are grounded into robot behaviors. Its planning stack, including cuRobo-based IK and motion generation, supports asynchronous microbatch solving so planner-backed skills can run inside a batched simulation loop without globally blocking all environments.

    \item \textbf{An annotation-rich data generation and serving system.}
    MagicSim separates asset-level world priors from runtime trajectory supervision. Asset annotations describe objects, rooms, layouts, and semantic metadata before interaction, while runtime annotators record what happens during the episode. The system combines camera streams, Omni Replicator annotations, language outputs, skill and planner traces, and MagicSim-native annotations such as end-effector waypoints, object bounding boxes, and affordance targets, then saves successful interactions as structured multimodal trajectories for local or served use.
\end{enumerate}
\FloatBarrier
\section{Related Work}

\begin{table*}[htbp]
\centering
\small
\setlength{\tabcolsep}{4pt}
\renewcommand{\arraystretch}{1.25}
\begin{adjustbox}{max width=\textwidth,center}
\begin{tabular}{l ccccc ccccc ccccc cccccc}
\toprule
\multirow{2}{*}{\textbf{System}}
 & \multicolumn{5}{c}{\textbf{World, phenomena \& actors}}
 & \multicolumn{5}{c}{\textbf{Robot embodiments}}
 & \multicolumn{5}{c}{\textbf{Deterministic execution runtime}}
 & \multicolumn{6}{c}{\textbf{Data, tasks \& agent interfaces}} \\
\cmidrule(lr){2-6}\cmidrule(lr){7-11}\cmidrule(lr){12-16}\cmidrule(lr){17-22}
 & \rot{Rigid / artic.}
 & \rot{Soft / cloth / rope}
 & \rot{Fluid / granular}
 & \rot{Flow / fire / dust}
 & \rot{Avatar / human}
 & \rot{Single / dual arm}
 & \rot{Dexterous hand}
 & \rot{Wheeled mobile}
 & \rot{Quadruped}
 & \rot{Humanoid}
 & \rot{Parallel envs}
 & \rot{Seeded DR}
 & \rot{Replay / \texttt{reset\_to}}
 & \rot{Planner-grounded skills}
 & \rot{Async planning}
 & \rot{Benchmark MDP}
 & \rot{Success-gated data-gen}
 & \rot{Native annot.}
 & \rot{Tactile sensing}
 & \rot{Language / task traces}
 & \rot{Agent / VLM API} \\
\midrule
\rowcolor{cmRowGray} Isaac Lab    & \Y&\Pt&\N&\N&\N & \Y&\Y&\Pt&\Y&\Y & \Y&\Y&\Pt&\Pt&\N & \Y&\Pt&\Pt&\Pt&\N&\N \\
ManiSkill3   & \Y&\Pt&\N&\N&\N & \Y&\Y&\Y&\Y&\Y & \Y&\Y&\Y&\Y&\N & \Y&\Y&\N&\N&\N&\Pt \\
\rowcolor{cmRowGray} RoboCasa     & \Y&\N&\N&\N&\N & \Y&\N&\Y&\Pt&\Pt & \Pt&\Y&\Y&\Pt&\N & \Y&\Y&\N&\N&\Y&\Pt \\
RoboTwin 2.0 & \Y&\N&\N&\N&\N & \Y&\N&\N&\N&\N & \Pt&\Y&\Pt&\Y&\N & \Y&\Y&\Y&\N&\Y&\Pt \\
\rowcolor{cmRowGray} RLBench      & \Y&\N&\N&\N&\N & \Pt&\N&\N&\N&\N & \N&\Pt&\Pt&\Y&\N & \Y&\Y&\Pt&\N&\Y&\N \\
BEHAVIOR-1K  & \Y&\Y&\Y&\Y&\N & \Y&\N&\Y&\N&\Pt & \Pt&\Pt&\Y&\Y&\N & \Y&\Pt&\Pt&\N&\Y&\Pt \\
\rowcolor{cmRowGray} Genesis      & \Y&\Y&\Y&\Y&\Pt & \Y&\N&\N&\Y&\N & \Y&\Pt&\Pt&\Y&\N & \Pt&\N&\N&\Pt&\N&\N \\
Habitat 3.0  & \Y&\N&\N&\N&\Y & \Pt&\N&\Y&\Pt&\Pt & \Y&\Pt&\Y&\Y&\N & \Y&\Pt&\N&\N&\Y&\Y \\
\midrule
\rowcolor{cmRowGreen} \textbf{MagicSim (ours)} & \Y&\Y&\Y&\Y&\Y & \Y&\Y&\Y&\Y&\Y & \Y&\Y&\Y&\Y&\Y & \Y&\Y&\Y&\Y&\Y&\Pt \\
\bottomrule
\end{tabular}
\end{adjustbox}
\caption{Capability comparison of MagicSim against representative robotics / embodied-AI simulators and benchmarks, judged by each system's official documentation or papers (June 2026).
\textcolor{cmGreen}{\ding{51}}~=~full support,\quad \textcolor{cmOrange}{\LEFTcircle}~=~partial / limited,\quad \textcolor{cmGray}{\ding{55}}~=~not supported.
``Seeded DR'' = seeded domain randomization; ``Native annot.'' = native end-effector-waypoint / 3D-box / affordance annotations.}
\label{tab:comparison}
\end{table*}

\textbf{Robotics and physics simulators.}
Modern robot-learning systems rely on mature simulation substrates for physics, rendering, and parallel environment execution. Isaac Gym and Isaac Lab emphasize GPU-parallel robot learning workflows \cite{makoviychuk2021isaacgym,isaacLab2025}, while MuJoCo remains a widely used engine for control and robotics \cite{todorov2012mujoco}. SAPIEN and ManiSkill further support articulated-object interaction and manipulation benchmarks \cite{xiang2020sapien,maniskill2_2023}. MagicSim is complementary to these simulators: it is not a new physics engine or renderer, but a manager-centric layer for deterministic world construction, heterogeneous scene objects, planner-in-the-loop execution, synchronized annotation, data collection, replay, and serving.

\textbf{Embodied AI and robot-learning benchmarks.}
Embodied AI and robot-learning benchmarks define what agents should solve. Habitat established a modular platform for embodied AI in 3D environments \cite{habitat2019}; iGibson and BEHAVIOR extended this direction toward interactive household scenes and everyday activities \cite{igibson2020,li2024behavior1k}. In manipulation, RLBench provides diverse task demonstrations \cite{rlbench2019}, while CALVIN and RoboCasa focus on long-horizon, language-conditioned household manipulation \cite{calvin2021,robocasa2024}. MagicSim adopts explicit task definitions, but treats each task MDP as a reusable runtime object shared by benchmarking, scripted collection, replay, and remote inference.

\textbf{Synthetic data and demonstration generation.}
Recent robot learning increasingly depends on scalable, diverse, and standardized trajectory data. Open X-Embodiment aggregates cross-robot datasets for RT-X-style generalist policies \cite{openX2023}, while BridgeData V2 and DROID show the value of broad real-robot data for generalization \cite{bridgedata2023,droid2024}. In simulation, VIMA studies multimodal-prompt manipulation \cite{vima2022}, and MimicGen and RoboCasa demonstrate scalable synthetic demonstration generation \cite{mandlekar2023mimicgen,robocasa2024}. MagicSim contributes at the simulator-production layer: it generates success-gated trajectories with synchronized actions, observations, cameras, annotations, and language streams.

\textbf{Motion planning and task-and-motion planning.}
Motion planning provides the execution backbone for high-level robot commands. OMPL and MoveIt made reusable planning infrastructure broadly available \cite{ompl2012,moveit2014}, while task-and-motion planning studies the coupling of symbolic task structure with continuous feasibility \cite{kaelbling2011,tampSurvey2024}. cuRobo pushes motion generation toward massively parallel GPU-based collision checking and trajectory optimization \cite{curobo2023}. MagicSim builds on this lineage, but makes planning a first-class runtime service: planner requests, skill execution, and environment state are handled within the same manager abstraction rather than through an external planning script.

\textbf{VLM/VLA and agent interfaces to simulation.}
Multimodal robot-learning systems motivate simulators that expose more than low-level states. CALVIN and VIMA use language or multimodal prompts for long-horizon manipulation \cite{calvin2021,vima2022}; RT-1 and RT-2 train visuomotor policies from large-scale robot data \cite{rt1_2022,rt2_2023}; and SayCan and PaLM-E ground language models through robot skills, perception, and affordances \cite{saycan2022,palme2023}. MagicSim is positioned as the simulator-side interface layer for such agents, exposing visual streams, structured scene facts, language-conditioned tasks, skill APIs, replay, and remote serving while leaving closed-loop VLM/VLA planning as downstream use.
\FloatBarrier
\section{System Overview: Parallel Episode Runtime for Executable Interaction Data}
\label{sec:system-overview}

MagicSim treats an \emph{episode}, rather than a rendered frame, static scene, or
single controller call, as the unit of system design. A frame records appearance;
an episode preserves the interaction-dependent state that makes a simulated world
reusable: replayable initial conditions, task success state, planner and skill
history, grounded affordances, and the trajectory that is retained only when the
interaction succeeds.

This episode view makes parallelism a runtime requirement. MagicSim does not
scale by assigning one Isaac Sim~\cite{makoviychuk2021isaacgym, mittal2023orbit}
process to every agent, planner request, or task attempt. Instead, many
sub-environments share one batched simulator process, while each environment
maintains its own randomized world state~\cite{tobin2017domain}, command state,
planner futures, task status, and recording buffer. Physics time is synchronized
across the batch; semantic time is asynchronous across agents, so at any instant
different environments occupy different control stages while the same task MDP is
used for RL, scripted collection, and planned inference or replay.

\subsection{Episodes, Not Frames}
\label{sec:overview-episodes-not-frames}

A conventional rendering pipeline can emit images whether or not an interaction
worked. MagicSim instead preserves the causal structure around those images. An
episode couples five axes: \emph{world} specifies what exists; \emph{runtime}
specifies what can be reproduced; \emph{action} specifies what can be executed;
\emph{supervision} specifies what can be observed and labeled; and
\emph{exposure} specifies how the result is used as a benchmark rollout, dataset
trajectory, or agent-facing interaction.

These axes are not separate pipelines. They are different views of the same
running episode. A grasping trajectory, for example, is simultaneously a world
state transition, a robot-control process, a planner-backed skill execution, a
source of affordance supervision, and a candidate data record. The system
overview therefore focuses on how MagicSim keeps these views aligned while
running many episodes in parallel.

\subsection{Parallelism as a Runtime Contract}
\label{sec:overview-parallel-runtime}

MagicSim's primary unit of scalability is the batched simulator process. A single
Isaac Sim instance hosts many sub-environments, and managers own
per-environment slices of the shared stage. Scene, layout, terrain, robot,
planner, camera, capture, navigation, avatar, language, and record managers each
own one part of the runtime state, while the environment wrapper sequences them
through a common reset and step lifecycle.

Parallelism appears at several levels. Physics and rendering are batched over
sub-environments. Resets and replay operate on selected \texttt{env\_ids}. Each
environment may receive a different action, command, skill phase, or planner
request. Planner calls are submitted asynchronously and solved off the main
simulator loop. Each environment also owns its own success gate and trajectory
buffer, so one successful episode can be written while other episodes continue
running.

\begin{table}[htbp]
\centering
\small
\begin{tabularx}{\linewidth}{>{\bfseries}p{0.18\linewidth} p{0.24\linewidth} X}
\toprule
Stage & Parallel unit & Runtime contract \\
\midrule
World state &
Sub-environment &
Managers own per-env slices of the shared Isaac Sim stage. \\

Reset and replay &
\texttt{env\_ids} &
Reset, randomization, snapshot, and \texttt{reset\_to} operate on selected
environments. \\

Physics stepping &
Batched simulator tick &
Actions are applied before a synchronized simulation step. \\

Agent progress &
Per-env command state &
Different environments may be running different tasks, skills, planner phases,
or reset states. \\

Planner compute &
Planner request &
Planner calls return futures and are solved asynchronously off the main loop
(\S\ref{sec:async-solve-farm}). \\

Data recording &
Per-env trajectory &
Each environment has its own buffers and writes only after success. \\
\bottomrule
\end{tabularx}
\caption{Parallelism in MagicSim is staged across world state, reset/replay,
physics stepping, agent progress, planner compute, and data recording.}
\label{tab:overview-parallel-stages}
\end{table}

The important point is that MagicSim does not require one simulator process per
agent. Multiple simulator processes can still be used as an outer serving or
cluster-scaling layer, but the core systems contribution is inside a single
batched runtime: many agents, tasks, planner requests, and trajectory buffers
coexist within one simulator instance. This is what allows both RL and data
collection to scale without turning each agent into a separate Isaac Sim process.

\subsection{Runtime Topology and the MDP Boundary}
\label{sec:overview-topology}

\begin{figure*}[htbp]
\centering
\includegraphics[width=\textwidth]{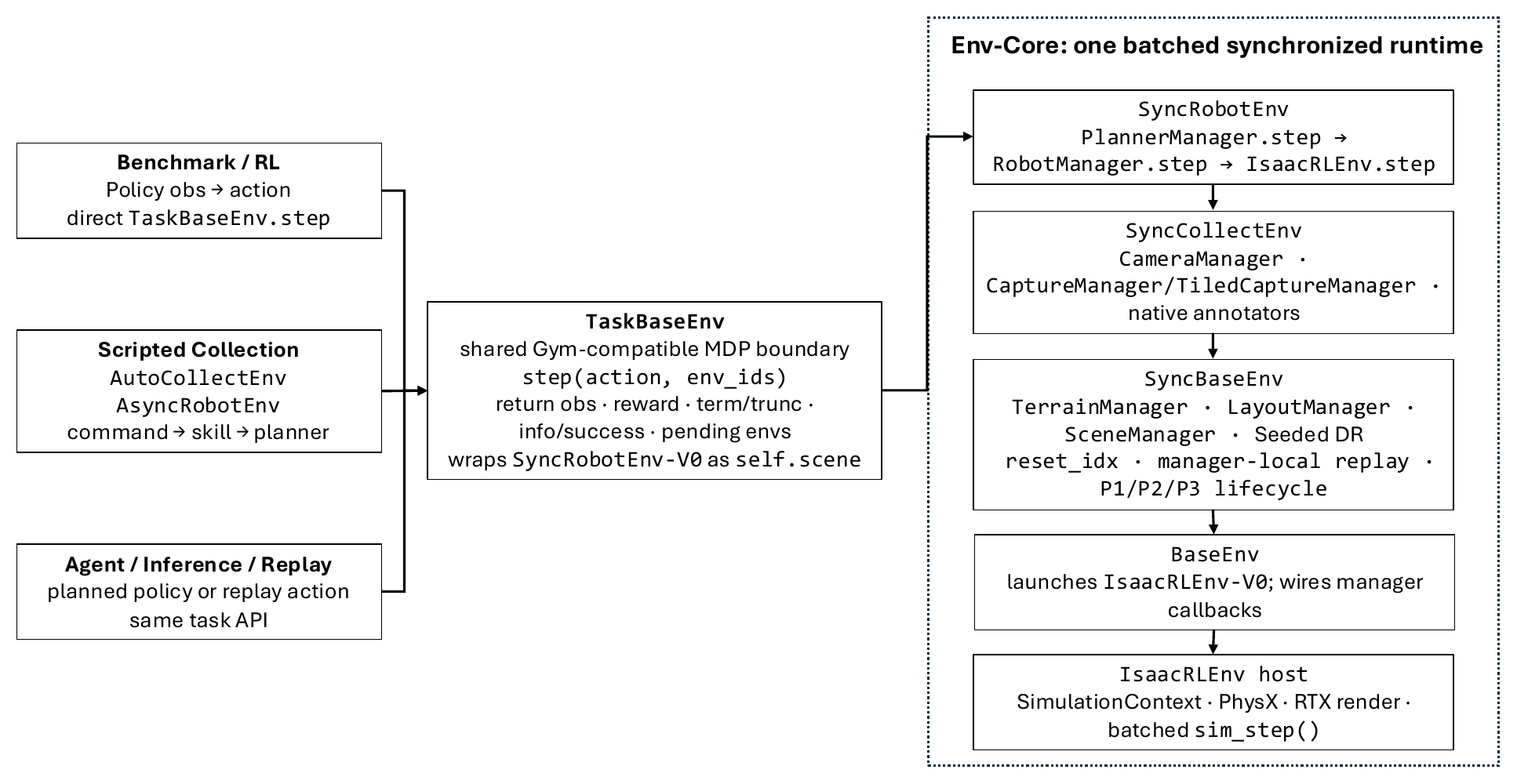}
\caption{Runtime topology and the MDP boundary. Env-Core constructs a batched
physical runtime; \texttt{TaskBaseEnv} exposes it as a shared MDP; different
drivers step the same task interface.}
\label{fig:overview-runtime-topology}
\end{figure*}

The runtime topology has two boundaries (Figure~\ref{fig:overview-runtime-topology}). The first boundary separates Env-Core
from the task layer. Env-Core managers and synchronized wrappers construct and
step the physical world; in robot environments, this core is exposed through
\texttt{SyncRobotEnv}. The second boundary is the MDP boundary.
\texttt{TaskBaseEnv} wraps \texttt{SyncRobotEnv} and defines the observation
space, action space, reward, termination, and success signal.

This topology is deliberately asymmetric. RL training is an external loop that
directly steps \texttt{TaskBaseEnv}. Scripted data collection is implemented as
wrappers above the same task: \texttt{AsyncRobotEnv} adds AtomicSkill and
GlobalPlanner managers, and \texttt{AutoCollectEnv} adds command sampling and
recording, and drives the Env-Core language annotation. Inference and replay are
planned as an external driver over the same MDP, using the RobotManager action
space rather than the Collect layer's skill and planner hierarchy.

The MDP boundary is therefore the point where different research modes meet.
Below it, the same world, robot, sensors, planners, and runtime determinism are
reused. Above it, a user can attach an RL policy, a scripted data-collection
stack, or a policy/replay driver. MagicSim can therefore support benchmark
rollouts, data trajectories, and agent-facing interactions without defining
three separate simulators.

\subsection{One MDP, Three Drivers}
\label{sec:overview-one-mdp-three-drivers}

The headline claim of the system is \emph{one MDP, three drivers}. A task is
defined once as a Gym-compatible~\cite{brockman2016openai} MDP with observations,
actions, rewards, terminations, and success conditions. That same MDP can then be
stepped by three different drivers.

\begin{table}[htbp]
\centering
\small
\begin{tabularx}{\linewidth}{>{\bfseries}p{0.21\linewidth} p{0.15\linewidth} X}
\toprule
Driver & Status & Action path \\
\midrule
RL training &
current &
A policy directly steps \texttt{TaskBaseEnv}. The action is interpreted by the
task and RobotManager action space, and the environment returns batched
observations, rewards, terminations, and success information. \\

Scripted AutoCollect &
current &
\texttt{AutoCollectEnv} wraps the same task through \texttt{AsyncRobotEnv}. The
path is Command $\rightarrow$ Skill $\rightarrow$ Planner $\rightarrow$ Robot
$\rightarrow$ Record, with per-env state machines driving scripted
demonstrations. \\

Inference / Replay &
planned &
\texttt{InferenceRunner} drives the same \texttt{TaskBaseEnv} with a policy or
replayed action stream. The action path uses the RobotManager action space and
does not go through the Collect layer's AtomicSkill or GlobalPlanner stack. \\
\bottomrule
\end{tabularx}
\caption{The three drivers step the same task MDP rather than defining separate
environments.}
\label{tab:overview-three-drivers}
\end{table}

Drivers and exposure surfaces are different axes. RL training, AutoCollect, and
Inference/Replay describe \emph{how the MDP is stepped}. Benchmark rollouts,
dataset trajectories, and agent-facing interactions describe \emph{how the
resulting episode is exposed}. Keeping these two axes separate prevents the
system from becoming a collection of special-purpose pipelines. The driver
changes, but the batched runtime does not.

\subsection{Agent-Level Asynchrony}
\label{sec:overview-agent-level-async}

\begin{figure*}[htbp]
\centering
\includegraphics[width=\textwidth]{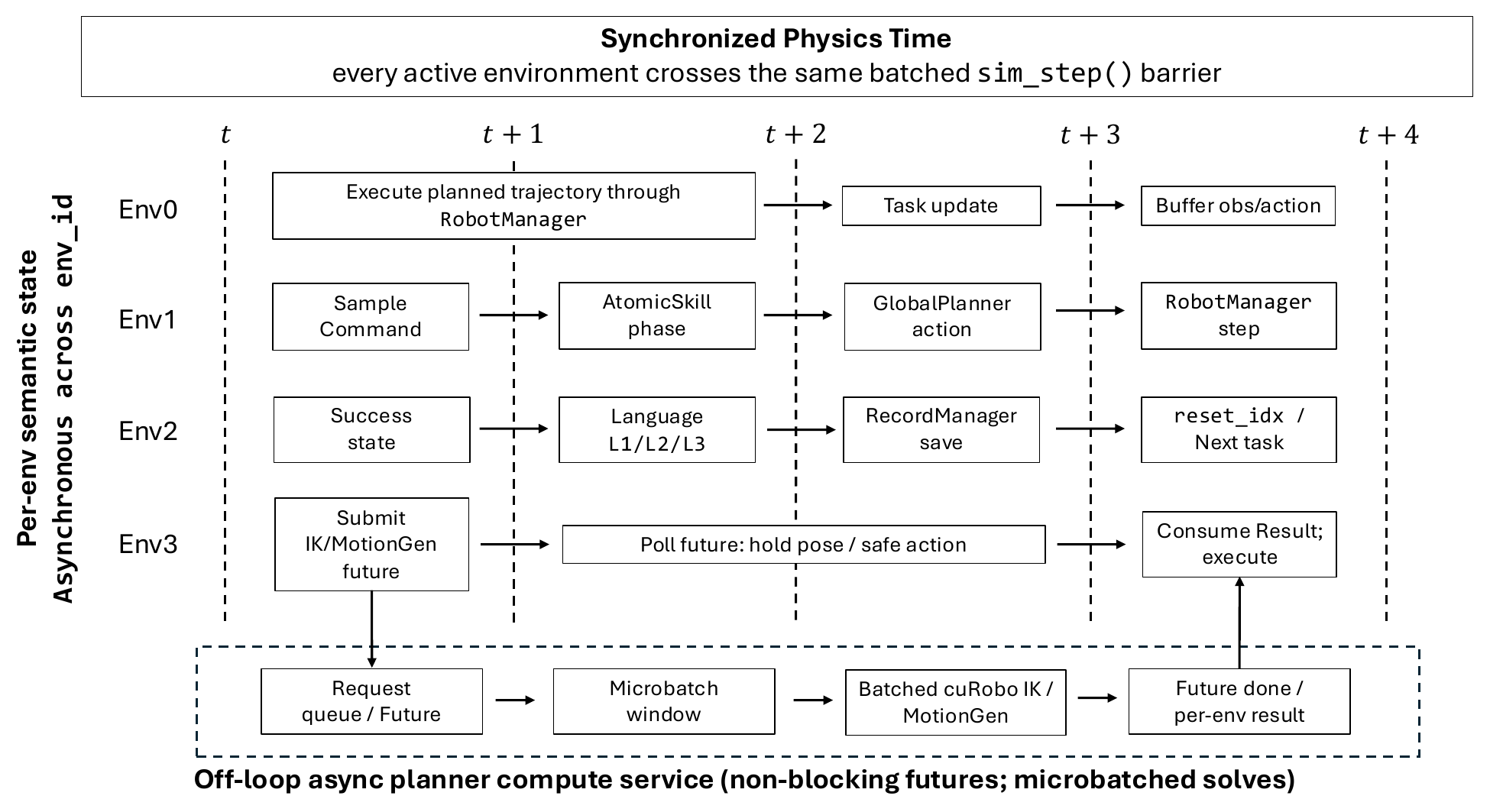}
\caption{Agent-level asynchrony. Environments share a synchronized physics tick
but progress through semantic control states independently.}
\label{fig:overview-agent-async}
\end{figure*}

The most important form of parallelism in MagicSim is not merely that multiple
environments step together. It is that multiple \emph{agents} can be at different
semantic stages while the simulator remains batched
(Figure~\ref{fig:overview-agent-async}).

This matters because planner-backed behavior is not constant-time. If every
planning request were executed synchronously inside the main loop, a single slow
environment would stall the entire batch. MagicSim instead exposes a
submit-and-poll contract: a skill or globalplanner submits a request, receives a
future, lets the batch continue stepping, and consumes the result when it is
ready. The planner service realizes the compute side of this abstraction as an
asynchronous microbatch solve-farm~\cite{curobo2023}; the mechanism
is deferred to \S\ref{sec:async-solve-farm}. For the overview, two properties
matter: the simulator loop is never blocked by a slow solve, and per-agent
requests are aggregated into larger batched solves rather than issued as many
isolated calls.

Agent-level asynchrony is also what connects high-level commands to low-level
execution without introducing ``magic'' actions. A high-level command does not
teleport the world into a goal state. It instantiates a task command, advances
an AtomicSkill state machine, submits planner requests, executes robot actions,
observes what happened, and updates the task state. The agent may reason at the
level of commands, but the runtime records only data produced by grounded actions
inside the simulator.

The same asynchronous, batched, and agent-addressable runtime positions MagicSim
as a rollout substrate for high-level agent learning. A vision-language model or
other planning policy can emit task or skill commands against many parallel
simulated worlds rather than only replaying pre-recorded trajectories. We treat
closed-loop reinforcement learning of such high-level planners---for example
through RL post-training infrastructures such as \texttt{verl}~\cite{sheng2024hybridflow}
or \texttt{RLinf}~\cite{yu2025rlinf}---as a planned use of this interface rather
than a completed capability. The current VLM layout hook is an integration point
(\S\ref{sec:limitations}), while the agent-facing serving interfaces are described
later in the report (\S\ref{sec:agent-capability}).

\subsection{How Interaction Becomes Data}
\label{sec:overview-interaction-to-data}

\begin{figure*}[htbp]
\centering
\includegraphics[width=\textwidth]{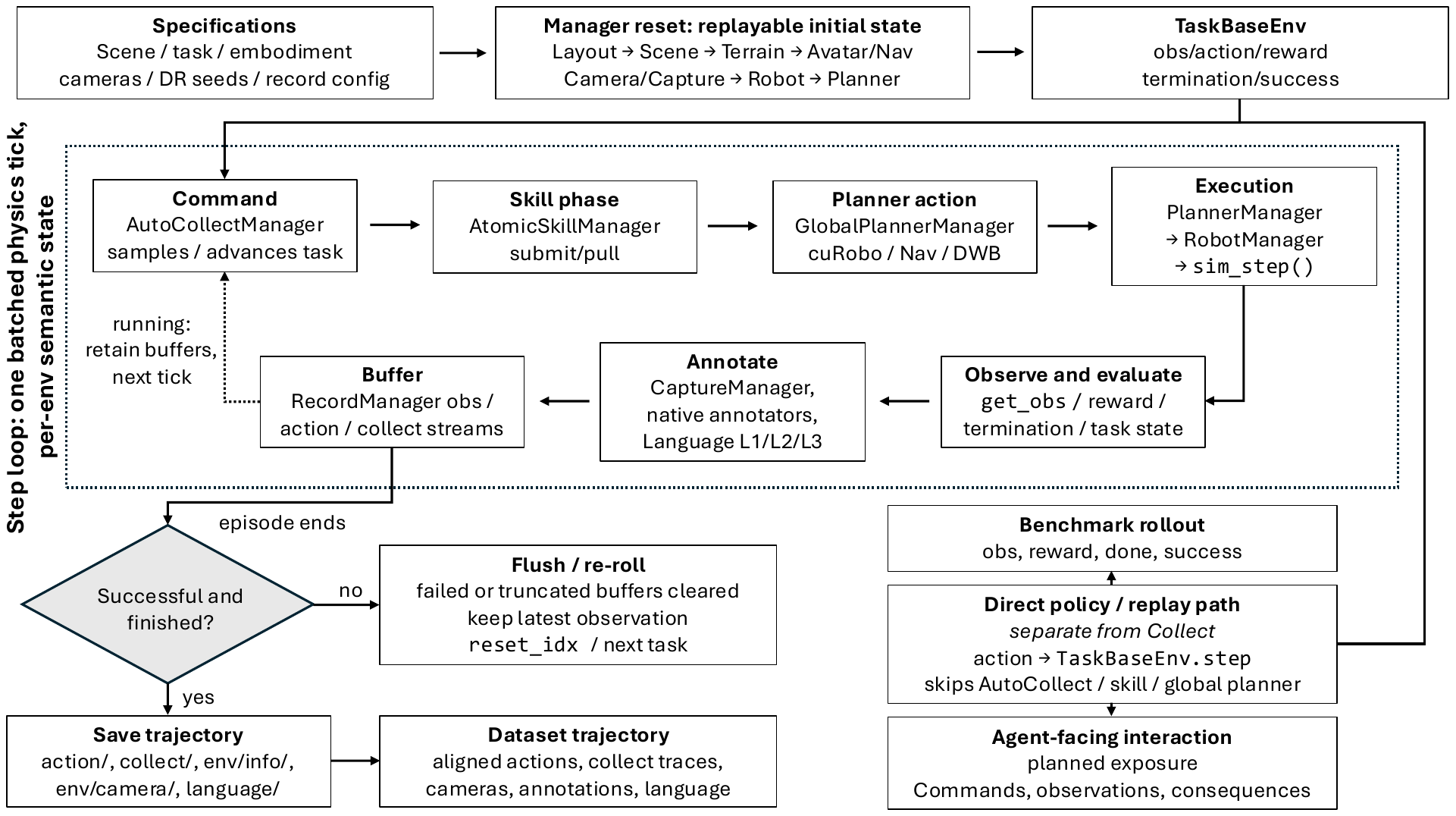}
\caption{Episode lifecycle. MagicSim turns interaction into data only after
execution, observation, annotation, and success gating.}
\label{fig:overview-lifecycle}
\end{figure*}

A MagicSim episode becomes data only after execution and validation (Figure~\ref{fig:overview-lifecycle}). The system
does not save every attempted rollout. Instead, each environment accumulates
observations, actions, task states, skill states, planner states, camera streams,
annotations, and language into per-env buffers. The task success signal then
serves two roles: it is an evaluation signal for the MDP and a write gate for
data collection.

In MagicSim, interaction is not data until it succeeds and is recorded. This is
the payoff of treating episodes rather than frames as the unit of design. A
render pipeline emits images whether or not the robot grasped the object,
reached the target, followed the plan, or satisfied the task. MagicSim instead
records the interaction structure around the frame: what command was active,
which skill phase produced the action, which planner target was selected, which
robot action was applied, which observation and annotation streams were
produced, and whether the task state justified saving the episode.

The same lifecycle supports all three exposure surfaces. For benchmarking, the
episode is a rollout with task observations, rewards, terminations, and success.
For data collection, the episode is a success-gated multimodal trajectory with
action, collect, camera, annotation, and language streams. For agent interaction,
the episode is exposed through an environment or serving interface where
commands, observations, and consequences unfold over time.

\subsection{Architectural Commitments and Report Roadmap}
\label{sec:overview-roadmap}

These choices amount to five architectural commitments. First, MagicSim decouples
\emph{what} exists, \emph{where} it is placed, and \emph{how} it can be acted on.
Second, it treats determinism as a runtime contract rather than a logging
convention. Third, it makes planning a first-class runtime component rather than
an offline preprocessor. Fourth, it treats data as a runtime product, produced
by execution, annotation, and success gating. Finally, it exposes one task
interface across evaluation, collection, and interaction, so the same episode
substrate can serve RL training, data generation, and agent-facing use.

The remaining chapters unpack this overview by following the executable verbs summarized in Table~\ref{tab:overview-roadmap}. MagicSim first builds heterogeneous worlds, then makes them reproducible, executes embodied actions, annotates runtime interaction, evaluates tasks, composes commands into skills, and finally collects or exposes the resulting data.

\begin{table}[htbp]
\centering
\small
\setlength{\tabcolsep}{4pt}
\renewcommand{\arraystretch}{1.08}
\begin{tabularx}{\linewidth}{>{\bfseries}p{0.17\linewidth} p{0.18\linewidth} X}
\toprule
Verb & Section & Role in the system \\
\midrule
Build
& \S\ref{sec:engine}
& Construct heterogeneous multi-physics worlds, including rigid and articulated objects, deformables, fluids, granular media, ropes, effects, avatars, world carriers, and embodiments. \\

Reproduce
& \S\ref{sec:runtime}
& Make randomized batched worlds deterministic through manager-owned state, seeded streams, reset/reset\_to, snapshots, and object lifecycle handling. \\

Execute
& \S\ref{sec:control-planning}
& Run embodied actions through robot channels, closed-loop controllers, navigation, motion planning, asynchronous planner services, flying cameras, and avatars. \\

Sense
& \S\ref{sec:sensors}
& Simulate synchronized observation streams, including camera/RGB-D capture, LiDAR/IMU, tactile contact, occupancy, navmesh queries, and frame tracking. \\

Annotate
& \S\ref{sec:assets-annotations}
& Compile offline asset priors and record runtime supervision, including Omni annotations, native end-effector/object/affordance labels, and L1/L2/L3 language. \\

Evaluate
& \S\ref{sec:tasks-benchmarks}
& Define the task MDP contract, task families, observations, rewards, terminations, success status, and benchmark protocol. \\

Compose
& \S\ref{sec:atomicskills}
& Lower commands into AtomicSkills and planner primitives with typed outcomes, retries, backend execution, and success gates. \\

Collect / Serve
& \S\ref{sec:collection}
& Run AutoCollect, maintain per-environment asynchronous collection state, save success-gated trajectories, and expose the runtime through serving APIs. \\

Support
& \S\ref{sec:supported-capabilities}
& Summarize user-facing capability views: benchmark evaluation, data collection, and agent/VLM interaction, including current and planned status boundaries. \\

Position
& \S\ref{sec:downstream}
& Position MagicSim as executable infrastructure for robotics, physical reasoning, and embodied VLM/VLA or agent learning. \\

Discuss
& \S\ref{sec:discussion}
& State limitations and future directions on simulation fidelity, learned skills, and more complex long-horizon tasks. \\
\bottomrule
\end{tabularx}
\caption{Report roadmap. Each row points to the section that implements or evaluates one part of MagicSim's executable embodied-interaction stack.}
\label{tab:overview-roadmap}
\end{table}
\FloatBarrier
%

\section{MagicSim Simulation Engine: A Unified Multi-Physics Backbone}
\label{sec:engine}

\subsection{Engine Overview}

MagicSim is built around a unified simulation engine rather than a collection of
object-specific simulators assembled per task. Within one stage, rigid and
articulated bodies, finite-element soft solids and surfaces, particle-based cloth,
fluids, granular media, constraint chains, force-field interactions, volumetric
effects, and animated human avatars can be instantiated, advanced, and observed
together. The engine is layered on Isaac Sim and IsaacLab, but its role is not
merely to expose individual solvers. Its contribution is to make heterogeneous
simulation methods \emph{co-executable}: different object families share the same
world context, appear in the same episode, and, where supported by the underlying
physics, interact across solver boundaries.

This section focuses on what the engine can simulate and how these simulation
methods compose. The deterministic mechanisms that create, randomize, reset,
step, and replay these entities belong to the manager runtime
(Section~\ref{sec:runtime}). This separation is important: the engine defines
the multi-physics capability surface, while the runtime defines the lifecycle,
seeding, reset, and replay contracts.

\subsection{Programmatic Import and Parameterized Simulation Entities}
\label{sec:engine-programmatic-import}

A central design choice in MagicSim is that simulated entities are not treated as
monolithic USD props with fixed appearance and fixed dynamics. A USD file,
primitive, generated particle set, field source, or animation clip is only the
source representation. Before an entity becomes part of the executable world, the
engine imports or generates it through code, assigns it to a simulation family,
binds visual and physical materials, resolves its pose and scale, attaches
solver-specific state, and exposes observation hooks. This programmatic import
path is what makes pose randomization, scale randomization, visual
randomization, physics randomization, and multi-physics coupling possible in the
same world.

We distinguish three identifiers. Let \(c \in \mathcal{C}\) denote a
user-defined logical category, such as a mug, tray, shirt, cabinet, rice pile, or
human avatar. Let
\[
    \mathcal{F} = \{f_1,\ldots,f_K\}
\]
denote the set of simulation families exposed by the engine, where
\(K=|\mathcal{F}|\) is a variable family count rather than a fixed part of the
engine definition. An entity instance is identified by its environment, logical
category, and instance index,
\[
    e=(n,c,i),
\]
while its simulation behavior is determined by a family assignment
\[
    f(e)\in\mathcal{F}.
\]
Thus, \(c\) describes what the entity means in the scene, whereas \(f(e)\)
describes how it is simulated.

The final simulation entity can be written abstractly as
\[
    x_e =
    \Big(
        a_e,\,
        f(e),\,
        T_e,\,
        s_e,\,
        \theta^{\mathrm{vis}}_e,\,
        \theta^{\mathrm{phys}}_e,\,
        \theta^{\mathrm{solver}}_e,\,
        h_e
    \Big),
\]
where \(a_e\) is the selected asset or generated source, \(T_e\) is the world
transform, \(s_e\) is scale, \(\theta^{\mathrm{vis}}_e\) are visual parameters,
\(\theta^{\mathrm{phys}}_e\) are physical parameters,
\(\theta^{\mathrm{solver}}_e\) are solver-specific parameters, and \(h_e\)
denotes observation and annotation hooks. Pose, scale, visual appearance,
physical material, and solver parameters are therefore first-class construction
axes rather than post-hoc edits to a static scene.

This parameterized construction is what enables domain randomization and
multi-physics coupling. Visual randomization can vary color, texture, material
binding, emission, or effect presets. Pose and scale randomization can vary
object placement, support relations, and collision opportunities. Physical
randomization can vary mass, density, friction, restitution, damping, contact
offsets, stiffness, viscosity, cohesion, pressure, field strength, or flow
direction, depending on the simulation family. The runtime mechanisms that sample,
seed, reset, and replay these parameter values are described in
Section~\ref{sec:runtime}; here we emphasize the engine-level consequence:
MagicSim objects are configurable simulation entities, not fixed USD props.

\begin{table}[!htb]
\centering
\small
\setlength{\tabcolsep}{4pt}
\renewcommand{\arraystretch}{1.08}
\caption{Parameterized construction axes for a MagicSim simulation entity.
The simulation-family count \(K=|\mathcal{F}|\) is a variable. The current
implementation used in this report instantiates \(K=15\), but the engine
abstraction is written in terms of \(\mathcal{F}\) so that new families can be
added without changing the architecture.}
\label{tab:engine-parameterization-axes}
\begin{tabular}{p{0.22\linewidth}p{0.30\linewidth}p{0.38\linewidth}}
\toprule
Axis & Examples & Engine role \\
\midrule
Logical identity
& \(e=(n,c,i)\): env id, logical category, instance id
& Gives each entity a stable user-facing identity independent of its physical
stage prim path. \\
\midrule
Simulation family
& \(f(e)\in\mathcal{F}\): \texttt{Rigid}, \texttt{Fluid}, \texttt{Garment},
\texttt{Avatar}, etc.
& Selects the backend representation, solver state, and method-specific
parameters. \\
\midrule
Source representation
& USD file, USD folder, primitive geometry, generated particles, field source,
animation clip
& Provides the raw asset or generated source that the engine imports or binds to
a solver-specific representation. \\
\midrule
Layout parameters
& Position, orientation, scale
& Determine spatial relations, support, collision opportunities, and task
geometry. \\
\midrule
Visual parameters
& Color, texture, USD/MDL material, emissive preset, effect preset
& Support visual domain randomization without changing semantic identity or
simulation family. \\
\midrule
Physical parameters
& Mass, density, friction, restitution, damping, contact offset, rest offset
& Control contact behavior and dynamics for rigid, articulated, deformable, and
particle-based entities. \\
\midrule
Solver-specific parameters
& Joint limits, FEM stiffness, cloth constraints, particle spacing, viscosity,
cohesion, surface tension, pressure, field strength, flow direction
& Connect the user-facing family to the simulation method governing state
evolution. \\
\midrule
Observation hooks
& Semantic label, instance id, bbox, particle state, joint state, animation state
& Make the entity observable, annotatable, and replayable as part of an
executable episode. \\
\bottomrule
\end{tabular}
\end{table}

\subsection{Simulation Methods and Supported Families}
\label{sec:engine-supported-families}

MagicSim organizes its simulation support into a small number of method families.
\emph{Rigid and articulated dynamics} cover movable rigid bodies, jointed objects
such as drawers and cabinets, and large instanced rigid populations.
\emph{Finite-element simulation} covers volumetric soft bodies and surface cloth.
\emph{Particle-based and position-based systems} cover garments, fluids,
granular materials, fine-grain media, and pressurized soft shells.
\emph{Constraint-based chains} model rope-like objects. Beyond contact dynamics,
MagicSim also supports \emph{force-field interactions} for magnetic effects,
\emph{volumetric and emissive effects} for fire and flow, and
\emph{animation-driven entities} for human avatars.

These families do not collapse into one monolithic solver. Instead, they share a
common multi-physics engine boundary: they can be placed into the same world,
advanced as part of the same simulation episode, and exposed through the same
observation and annotation path. Physical parameters remain method-specific. For
example, rigid bodies expose mass, density, collision shape, friction,
restitution, and contact offsets; articulations expose joint state and joint
limits; finite-element bodies expose mesh resolution, stiffness, damping, density,
and contact parameters; particle systems expose particle spacing, contact offset,
density, viscosity, cohesion, surface tension, drag, and lift; flow effects
expose source pose, direction, and magnitude. How these parameters are sampled,
seeded, and replayed is handled by the runtime in Section~\ref{sec:runtime}.

\begin{figure*}[!htb]
\centering
\includegraphics[width=\textwidth]{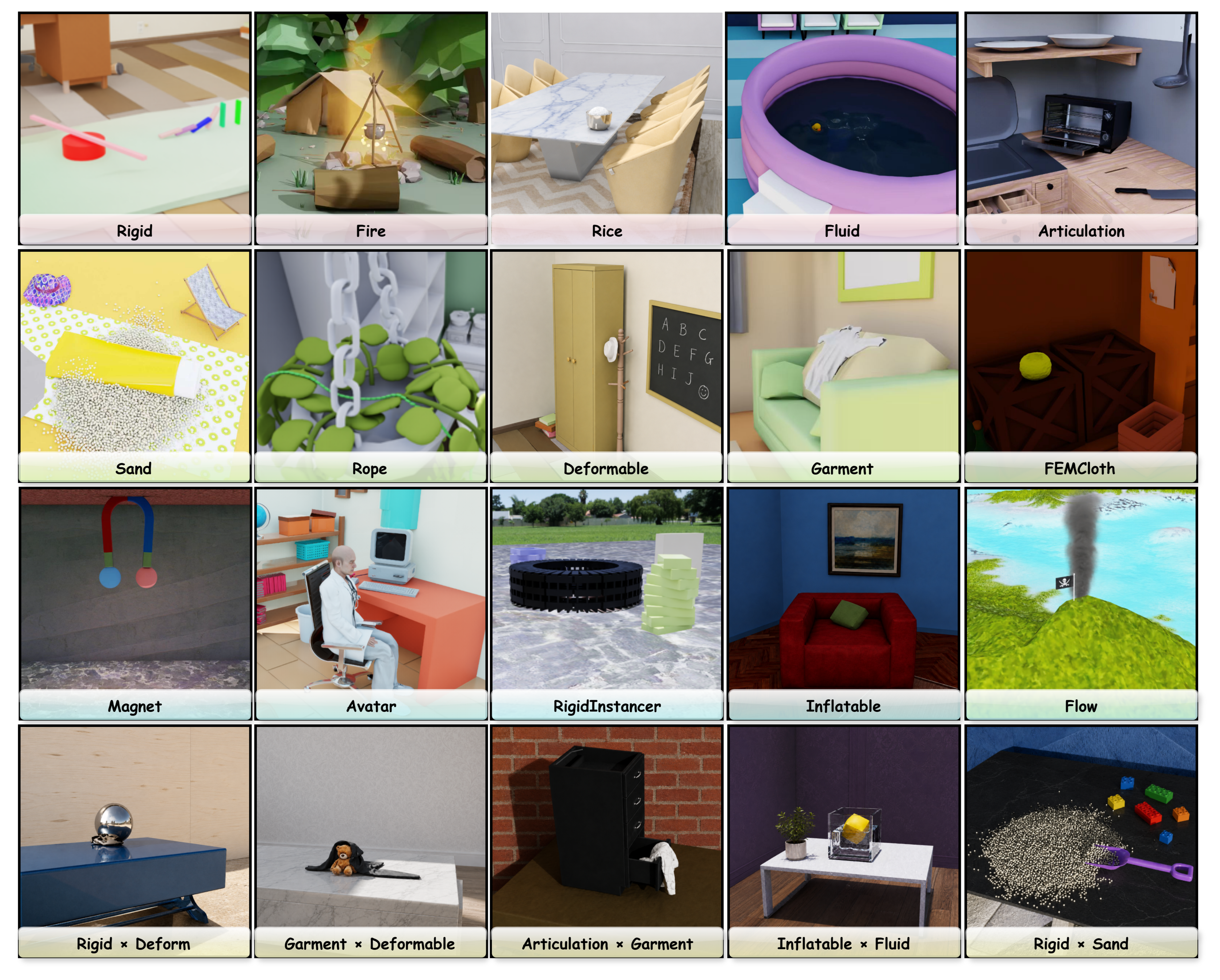}
\caption{MagicSim simulation families and coupled multi-physics interactions.
The first three rows show the current implementation of the family set
\(\mathcal{F}\), where \(K=15\), spanning rigid and articulated dynamics,
finite-element soft bodies and cloth, particle-based fluids and granular media,
force fields, volumetric effects, and animated avatars. The last row is the
coupled multi-physics evidence row: it shows representative cross-family
interactions in which entities from different solver families are co-executed
and physically interact in a single scene.}
\label{fig:engine-families}
\end{figure*}

MagicSim exposes a user-facing simulation-family set
\(\mathcal{F}=\{f_1,\ldots,f_K\}\). The family count \(K\) is deliberately
written as a variable: it is a property of the current implementation, not a
hard-coded ontology of the engine. The implementation described in this report
instantiates \(K=15\), summarized in Table~\ref{tab:engine-object-families} and
illustrated in Figure~\ref{fig:engine-families}.

These families are a capability taxonomy rather than a one-to-one mirror of the
internal manager registry. Internally, they are realized through multiple runtime
managers and specialized backends, including physics objects, artifacts and
effects, instancing and field backends, and the animation subsystem. To the user,
however, they present a unified simulation surface: each family can be specified
as part of a world, observed during an episode, and composed with other families
in executable scenes.\footnote{Static collision geometry is handled as an
internal scene-geometry substrate rather than as one of the user-facing
simulation families in \(\mathcal{F}\), and is omitted from
Figure~\ref{fig:engine-families}.}

\begin{table*}[htbp]
\centering
\footnotesize
\setlength{\tabcolsep}{3pt}
\renewcommand{\arraystretch}{1.08}
\caption{Current instantiation of MagicSim's simulation-family set
\(\mathcal{F}\), with \(K=15\) user-facing families. The table groups families
by simulation method and lists how each family is represented in simulation.
Detailed domain-randomization sampling, seeding, reset, and replay mechanisms
belong to the manager runtime rather than this engine taxonomy.}
\label{tab:engine-object-families}
\begin{tabularx}{\textwidth}{
    p{0.18\textwidth}
    p{0.13\textwidth}
    p{0.34\textwidth}
    X
}
\toprule
Method family & Family & How it is simulated & State and representative parameters \\
\midrule

Rigid and articulated dynamics
& \texttt{Rigid}
& PhysX rigid-body dynamics with collision geometry and rigid-body material.
& SE(3) pose, linear/angular velocity, mass or density, collision approximation,
friction, restitution, contact/rest offsets. \\

Rigid and articulated dynamics
& \texttt{Articulation}
& PhysX articulation for jointed rigid structures such as cabinets, drawers,
doors, and appliances.
& Root pose, joint positions/velocities, joint limits, drives, link collision,
rigid material parameters. \\

Rigid and articulated dynamics
& \texttt{RigidInstancer}
& Instanced rigid-body population for large numbers of repeated rigid entities.
& Per-instance transforms and velocities, shared or per-instance rigid material,
collision scale, density/friction settings. \\

\midrule
Finite-element simulation
& \texttt{Deformable}
& FEM volumetric soft body represented by a simulation tetrahedral mesh.
& Node positions/velocities, tetrahedral mesh resolution, density, Young's
modulus, Poisson ratio, damping, dynamic friction, contact/rest offsets. \\

Finite-element simulation
& \texttt{FEMCloth}
& FEM surface cloth for sheet-like deformable objects.
& Surface vertices, cloth mesh topology, stiffness/damping, density, thickness or
contact offset, friction and collision settings. \\

\midrule
Particle-based and position-based systems
& \texttt{Garment}
& Particle-based cloth for garments and fabric-like objects.
& Cloth particles, stretch/bend constraints, particle contact offset, rest offset,
solver iterations, density, friction, damping, drag/lift-style material terms. \\

Particle-based and position-based systems
& \texttt{Fluid}
& Particle-based fluid, optionally coupled with a container.
& Fluid particles, density, viscosity, cohesion, surface tension, drag, lift,
solid/fluid rest offsets, contact offset, solver iterations. \\

Particle-based and position-based systems
& \texttt{Sand}
& Granular particle medium for piles, scooping, pushing, and contact-rich
interaction.
& Granular particles, particle radius or contact offset, density, friction,
cohesion-like material terms, solver iterations, settling/contact parameters. \\

Particle-based and position-based systems
& \texttt{Rice}
& Fine-grain granular medium represented as small particle-like or grain-like
solid elements.
& Grain positions, grain size or particle spacing, density, friction, contact
offsets, material and collision parameters. \\

Particle-based and position-based systems
& \texttt{Inflatable}
& Pressurized soft-shell or particle-shell object.
& Shell particles or surface nodes, pressure-like constraint, stiffness, damping,
contact offsets, friction, deformation state. \\

\midrule
Constraint chains
& \texttt{Rope}
& Constraint-based capsule or chain representation for rope-like objects.
& Segment poses, capsule radius/length, chain constraints, bending/stretch
response, collision and friction parameters. \\

\midrule
Force fields
& \texttt{Magnet}
& Field-based interaction that applies non-contact forces to nearby compatible
bodies.
& Field source pose, range, strength, polarity or attraction/repulsion mode, and
affected-body filters. \\

\midrule
Volumetric and emissive effects
& \texttt{Fire}
& Volumetric or emissive effect backend for flame-like visual and semantic
context.
& Source pose, effect preset, color/emission, scale, intensity, temporal
variation; generally co-observed rather than contact-coupled. \\

Volumetric and emissive effects
& \texttt{Flow}
& Flow-like volumetric effect for smoke, steam, dust, or wind-style context.
& Source pose, flow direction, magnitude, scale, density or visual preset;
generally co-executed as context rather than universal reaction-force coupling. \\

\midrule
Animation-driven entities
& \texttt{Avatar}
& Animation-driven human entity controlled by skeletal animation or animation
graph variables, not by the robot controller stack.
& Root pose, skeletal joint transforms, animation state, command queue, gaze or
motion clip state; used for human-scene and human-robot context. \\

\bottomrule
\end{tabularx}
\end{table*}

\subsection{Coupled Multi-Physics Composition}
\label{sec:engine-coupled-composition}

The engine's defining property is coupled breadth, and it is useful to separate
two senses in which families compose. The first is \emph{co-execution}: all
families in \(\mathcal{F}\) can be instantiated in the same stage, advanced under
one simulation episode, and observed as a single scene, regardless of which
solver or backend governs each entity. Co-execution alone distinguishes the
engine from a collection of method-specific simulators, because a single episode
need not be confined to one solver's world.

The second and stronger sense is \emph{physical coupling}: entities governed by
different dynamics solvers do not merely coexist but interact across solver
boundaries. The bottom row of Figure~\ref{fig:engine-families} shows five
representative cross-family couplings in the current \(K=15\) instantiation:
rigid bodies against FEM soft solids
(Rigid\,$\times$\,Deformable), particle cloth against FEM solids
(Garment\,$\times$\,Deformable), articulated structures against particle cloth
(Articulation\,$\times$\,Garment), pressurized shells against fluids
(Inflatable\,$\times$\,Fluid), and rigid bodies against granular media
(Rigid\,$\times$\,Sand). These pairs span rigid-body--FEM, FEM--particle, and
particle--particle interaction, so the engine supports bidirectional contact
coupling across its dynamics solvers rather than co-locating independent
simulations that never touch.

Not every family participates in physical coupling in the same way. Volumetric
and emissive effects such as fire and flow, and animation-driven avatars, are
co-executed and co-observed within the scene but are not, in general, two-way
contact-coupled to the dynamics solvers; they contribute visual, semantic, and
interactive context rather than universal reaction forces. This distinction keeps
the engine claim precise: MagicSim supports broad co-execution across
\(\mathcal{F}\), and demonstrates physical coupling on representative
solver-family pairs where contact dynamics are supported.

\subsection{Heterogeneous Sub-Environments, One Engine}
\label{sec:engine-heterogeneous-subenvs}

The simulation-family set \(\mathcal{F}\) defines the interactive content that
MagicSim can instantiate, but the engine is designed for heterogeneity not only
across object types, but also across sub-environments. In a batched run,
different sub-envs may contain different assets, different logical categories,
different simulation families, different poses and scales, and different visual
and physical parameters. One sub-env may contain a rigid container, a fluid pour,
and an animated human recipient; another may contain an articulation interacting
with garment cloth; another may contain a rigid tool pushing granular media.
These sub-envs are not separate simulators. They are heterogeneous instances of
the same multi-physics engine, advanced under one simulation boundary and exposed
through a common observation and annotation surface.

This design follows from the parameterized import model described above. A USD
asset, primitive, generated particle set, field source, or animation clip is only
the source representation. At construction time, each entity is assigned a
logical identity \(e=(n,c,i)\), a simulation family \(f(e)\in\mathcal{F}\), a
world transform \(T_e\), a scale \(s_e\), visual parameters
\(\theta^{\mathrm{vis}}_e\), physical parameters
\(\theta^{\mathrm{phys}}_e\), and solver-specific parameters
\(\theta^{\mathrm{solver}}_e\). Because these quantities are resolved
programmatically rather than baked into a static USD scene, different sub-envs
can instantiate different mixtures of families and parameters while remaining
part of the same batched episode. This is what makes pose randomization, scale
randomization, visual randomization, physics randomization, and multi-physics
coupling compatible rather than separate features.

The engine therefore treats heterogeneity as a first-class property of the world
batch. The same user-facing simulation family can appear with different assets or
physical parameters across sub-envs, and different families can be combined in
different proportions across the batch. For example, \(\texttt{Rigid}\) objects
may differ in mass, friction, restitution, collision shape, and scale;
\(\texttt{Fluid}\) instances may differ in particle spacing, viscosity, cohesion,
and container geometry; \(\texttt{Garment}\) instances may differ in cloth
resolution, contact offset, stiffness, and material appearance; and
\(\texttt{Avatar}\) instances may differ in character asset, root pose, and
animation state. The sampling, seeding, reset, and replay mechanisms that make
this heterogeneity deterministic are part of the manager runtime in
Section~\ref{sec:runtime}; the engine-level point is that these parameters are
exposed as configurable simulation axes rather than hidden inside fixed assets.

Complete MagicSim worlds also include carriers, embodiments, observability, and
task structure. Terrains and rooms provide the world carrier: they define floors,
room-scale geometry, traversable regions, and navigation-relevant context, but
they are not counted as user-facing simulation families in \(\mathcal{F}\).
Terrain lifecycle semantics, including when terrain geometry is locked and what
is re-sampled on reset, belong to the deterministic manager runtime in
Section~\ref{sec:runtime}. Room-level world priors such as occupancy maps,
free-point sampling, room bounding boxes, and boundaries belong to the asset and
annotation system in Section~\ref{sec:annotation}. Navigation consumption by
navmesh queries, runtime occupancy maps, and DWB-style local planning belongs to
the control and planning stack in Section~\ref{sec:control-planning}.

Similarly, robots are not ordinary scene objects in this taxonomy. They are
controlled embodiments that act on the world through joint, base, end-effector,
navigation, and whole-body controllers. Avatars, by contrast, are
animation-driven human entities that provide human-scene context, interaction
targets, and social grounding. Sensors and annotators observe each heterogeneous
sub-env through camera, depth, segmentation, tactile, occupancy, navmesh,
language, and native trajectory annotations. Thus, MagicSim worlds are not static
asset arrangements: they are heterogeneous, parameterized, multi-physics
sub-environments that can be acted on, coupled, observed, annotated, and replayed
by the runtime described next.

For example, a long-horizon household episode can combine a navigable room, a
humanoid embodiment, rigid containers, a fluid pour, a tray, and an animated
human recipient. The room supplies the carrier, the humanoid supplies embodied
action, the rigid and fluid families supply coupled physical content, the avatar
supplies human context, and the sensor stack records the episode as a coherent
trajectory. This is the role of the simulation engine: it turns heterogeneous,
parameterized families into co-executable physical content that later chapters
can control, annotate, benchmark, and collect.
\FloatBarrier

\section{Deterministic Manager Runtime}
\label{sec:runtime}

Section~\ref{sec:engine} described what MagicSim can simulate: heterogeneous
objects, effects, and avatars can co-exist in one world. This section describes
how those worlds are made reproducible while many episodes share one continuously
stepping simulator. The key constraint is that the batched simulator is not
globally stopped or rebuilt when a single environment resets. Each
sub-environment therefore needs local reset, local randomization, local layout,
and local lifecycle decisions, even though all environments share one stage and
one simulator clock.

\begin{figure*}[htbp]
\centering
  \includegraphics[width=\linewidth]{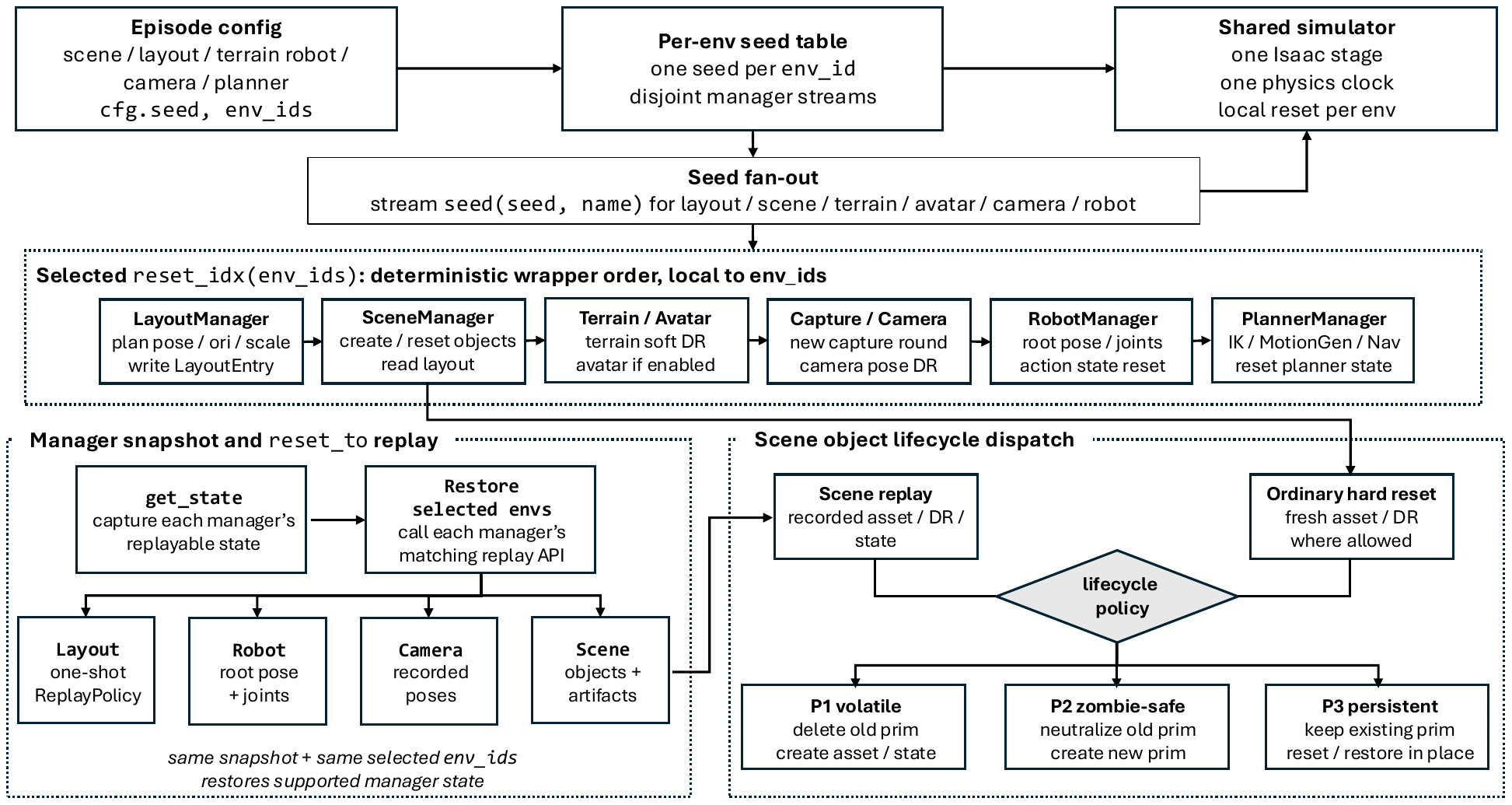}
  \caption{Deterministic manager runtime. A global seed fans out into disjoint
  per-environment, per-manager streams; \texttt{reset\_idx} drives a fixed
  manager reset order; \texttt{reset\_to} restores manager-owned snapshots; and
  hard resets dispatch each object through its P1/P2/P3 lifecycle mode.}
  \label{fig:runtime}
\end{figure*}

The central claim is that \emph{determinism is a runtime contract, not a property of
any single solver}.
MagicSim achieves it structurally: it assigns state to
managers, addresses operations by \texttt{env\_ids}, fixes the order in which
runtime components are invoked, isolates random draws into manager-specific
streams, and maps backend import/delete limitations into a small number of
lifecycle modes. The result is a
runtime in which one environment can reset without perturbing another, one
randomizer can change without reordering another, and many object families can
share the same reset scheduler (Figure~\ref{fig:runtime}).

\subsection{SimulationContext Host and Launch}
\label{sec:simulationcontext-host}

The shared stage is hosted by \texttt{IsaacRLEnv}. It owns the
\texttt{SimulationContext}, the batched scene, and the physics and rendering
settings under which the engine runs. This host is the execution context for the
solvers of Section~\ref{sec:engine}; it is not a second physics layer. The engine
defines the capability surface, while the runtime host provides one stage, one
clock, and one synchronized set of simulator services.

Before any manager creates stage content, MagicSim launches Isaac Sim from a
launch file. The launch file acts as a runtime manifest: it selects the
application mode, enables the required Isaac Sim extensions, and fixes
process-level settings that must exist before USD prims, physics views, cameras,
or annotators are constructed. This step is global and happens once. World
construction, placement, reset, and randomization are local and happen repeatedly
through managers.

Version-sensitive GPU dependencies are also fixed during launch. We keep that
detail at the runtime boundary only to preserve ordering: process-wide
dependencies must be pinned before managers and planners import CUDA-facing
modules. 
\subsection{Managers as State Owners}
\label{sec:manager-runtime}

Runtime state is partitioned across managers. Each manager owns one slice of the
batched stage and communicates with other managers through explicit contracts
rather than by mutating their internal state. The purpose is not just modularity;
it is determinism. If every subsystem owns its state and exposes the same
lifecycle, the environment wrapper can drive the world through one ordered
sequence instead of accumulating object-, sensor-, or robot-specific reset paths.

All managers follow the same lifecycle shape: construct from configuration and
per-environment seeds, initialize against the \texttt{SimulationContext}, bind
runtime views once the stage is live, reset either the full batch or selected
\texttt{env\_ids}, and snapshot or restore their own state. State restoration
through \texttt{reset\_to} is therefore part of the lifecycle contract: it
restores manager-owned initial conditions, not a separate log replay.

The user-facing object families of Section~\ref{sec:engine} are realized under
this manager contract rather than by a one-to-one registry. Most physical objects
and effects are owned by \texttt{SceneManager}; specialized instancing, field,
and grain backends cover families such as \texttt{RigidInstancer},
\texttt{Magnet}, and \texttt{Rice}; and animated humans are owned by
\texttt{AvatarManager}, a sibling of \texttt{SceneManager}. Robot, planner,
navigation, camera, capture, and tactile managers are sequenced by the same
runtime, but their control and observation semantics belong to
Sections~\ref{sec:control-planning} and~\ref{sec:annotation}.

The synchronized Env-Core wrapper chain simply assembles these manager groups:
\texttt{BaseEnv} launches the host, \texttt{SyncBaseEnv} adds world-core managers,
\texttt{SyncCollectEnv} adds the capture layer, and the final branch exposes
either \texttt{SyncRobotEnv} or \texttt{SyncCameraEnv}. Task wrappers,
asynchronous collection wrappers, language annotation drivers, and record
managers sit above this Env-Core boundary and are described later with the task,
annotation, and collection stack.

\subsection{LayoutManager as the Placement Backbone}
\label{sec:layout-manager}

A randomized world is not executable until objects have places to go. MagicSim
therefore separates object identity from object placement. \texttt{SceneManager}
owns what an entity is: its logical identity, asset choice, behavior, physics
state, and lifecycle. \texttt{LayoutManager} owns where that entity goes. Reset
begins with layout because placement must be decided before scene objects can be
created, reset, or rebound to physical prims.

The layout interface is shared across three placement backbones.

\paragraph{Declarative layout.}
The YAML backbone is the controlled path. Users specify placement ranges, room
regions, support surfaces, relations, sampling ratios, and replayable
randomization policies in configuration. This mode is useful when an experiment
needs an exact distribution over object poses, distractors, target regions, or
room-conditioned placements.

\paragraph{VLM-backed layout.}
The VLM backbone lets language or vision-language systems propose semantically
grounded arrangements, such as placing a cup near a table setting or populating a
room from a natural-language description. This follows recent scene-generation
systems such as SceneSmith, which builds simulation-ready indoor scenes from
language prompts through agentic VLM interactions~\cite{scenesmith2026}. In
MagicSim, the VLM is not allowed to bypass the runtime contract: it proposes a
layout plan, while \texttt{LayoutManager} still owns validation, seeding, replay,
and reset ordering.

\paragraph{Heuristic and constraint-based layout.}
The heuristic backbone uses geometric and task priors: room bounds, object
support surfaces, collision checks, free-space samples, spatial relations, and
affordance constraints. This is closest in spirit to systems such as Holodeck,
which uses language to generate embodied 3D environments and optimizes object
placement under spatial constraints~\cite{yang2024holodeck}. In MagicSim, these
heuristics are another route to the same output: a per-environment pose plan over
logical objects.

Across all three backbones, the invariant is the same. A logical object identity,
such as \texttt{(env\_id, cat\_name, inst\_id)}, is stable across episodes, while
its physical \texttt{prim\_path} is only a runtime handle. Tasks, policies,
annotations, and replay state refer to the logical object. The lifecycle
scheduler may later reallocate prims, but layout remains expressed over logical
identities and planned poses.

\begin{figure}[htbp] \centering \includegraphics[ width=0.98\linewidth ]{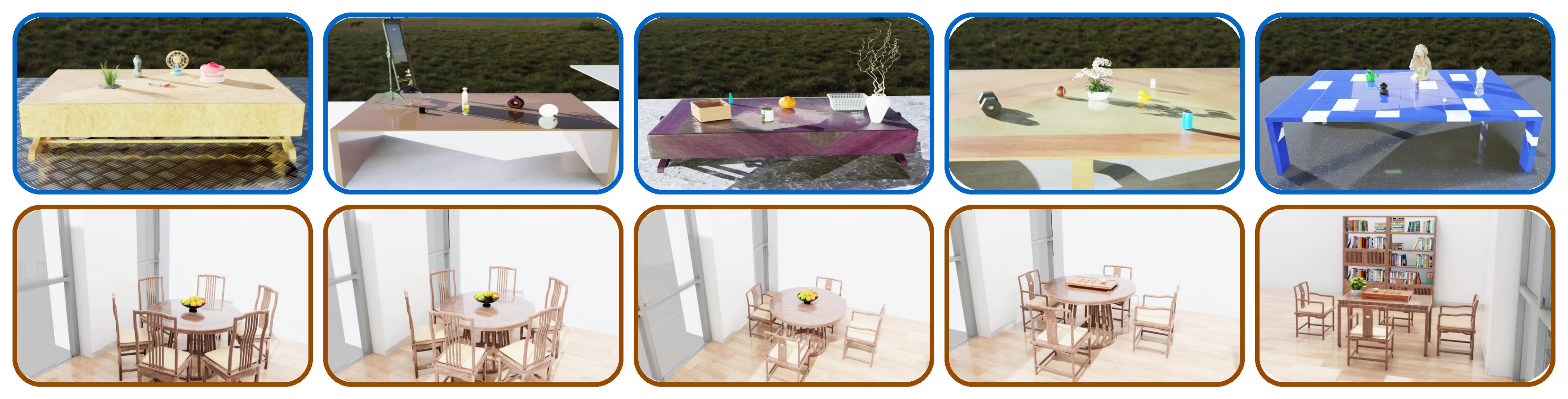} \caption[Randomized layouts at object and room scales.]{ \textbf{Representative randomized layouts at object and room scales in MagicSim.} The upper row shows object-level layouts in which support surfaces, object selections, positions, orientations, scales, and distractor arrangements can vary across environments. The lower row shows room-level layouts with different spatial arrangements of tables, chairs, furniture, and scene assets. Each layout is represented as a per-environment pose plan over logical object identities and can be generated through declarative configuration, VLM-backed proposals, or geometric and task-conditioned constraints. Layout sampling remains integrated with MagicSim's seeded reset and replay contract, allowing a sampled arrangement to be reconstructed without coupling it to unrelated scene, robot, camera, or physics randomization. } \label{fig:randomized-layouts} \end{figure}

Figure~\ref{fig:randomized-layouts} visualizes representative outputs of the shared layout interface at both object and room scales. Although the declarative, VLM-backed, and heuristic or constraint-based backbones differ in how they propose an arrangement, they produce the same runtime abstraction: a per-environment pose plan over stable logical object identities. 

\begin{figure}[htbp] \centering \includegraphics[width=0.98\linewidth]{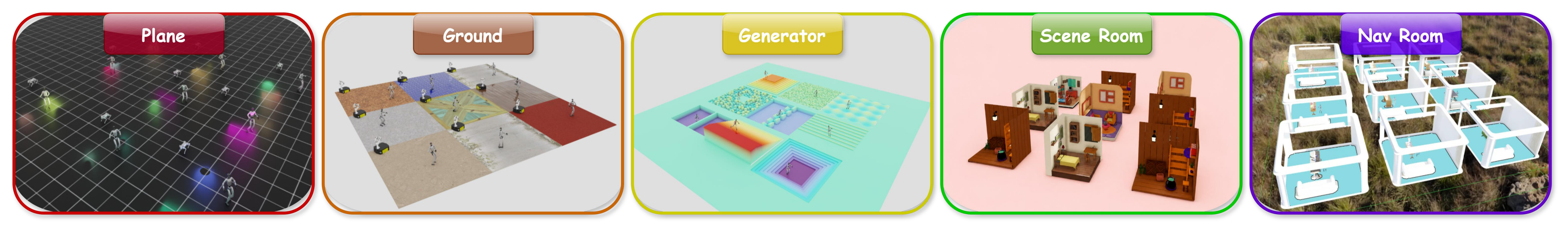} \caption[World carriers and room-scale environment backbones.]{ \textbf{World carriers and room-scale environment backbones in MagicSim.} Representative configurations include a flat plane, textured ground, procedurally generated terrain, scene-room layouts, and navigation-ready rooms. These carriers define floor geometry, traversable regions, and room-scale context while remaining integrated with the deterministic construction, reset, randomization, and replay runtime. Terrains and rooms serve as world carriers and are therefore not counted among the user-facing simulation families. } \label{fig:world-carriers} \end{figure}

Figure~\ref{fig:world-carriers} summarizes the world-carrier and room-scale environment configurations currently exposed by MagicSim. These configurations provide the geometric and navigational substrate on which heterogeneous simulation entities, robots, sensors, and tasks are instantiated. 

\subsection{Domain Randomization as Controlled Variation}
\label{sec:dr-seed-streams}

Domain randomization matters because MagicSim does not only randomize where
objects are placed; it randomizes how the world looks and how it behaves.
Appearance randomization covers object materials, colors, textures, texture
scales, terrain and floor appearance, lighting, background context, and visual
effects. These changes alter the observation stream while preserving the logical
task state.

Physics randomization covers parameters exposed by the engine families. Rigid and
articulated objects may vary mass or density, friction, restitution, collision
scale, and contact offsets. Deformable bodies, cloth, garments, fluids, sand,
rice, ropes, and inflatables may vary family-specific parameters such as
stiffness, damping, particle spacing, viscosity, cohesion, pressure, solver
iterations, or chain response. A \texttt{ratio} field is not itself a physical
coefficient; it is a policy gate that decides whether a perturbation, distractor,
material change, or physics variation is active for a given environment or reset.

Making this deterministic is difficult because the simulator does not stop.
Different environments reset at different semantic times, and random draws occur
inside layout, scene objects, terrain, robots, and cameras. MagicSim prevents
these draws from coupling by separating value parsing from random-stream
ownership. Values use one shared shape rule: constants stay fixed, ranges are
sampled uniformly, and vector ranges are sampled component-wise. The random
source is then isolated by environment and by manager.

The unit of isolation is the manager stream, not the episode. A global seed
expands into per-environment seeds, and each manager derives its own stream, such
as \texttt{layout}, \texttt{scene}, \texttt{terrain}, \texttt{robot}, or
\texttt{camera}. Adding a new texture randomizer cannot change object physics;
adding a new mass draw cannot change lighting; resetting environment 3 does not
advance environment 4. Stream isolation is therefore stronger than logging:
logging records what happened, while stream isolation prevents unrelated random
choices from becoming coupled in the first place.

\begin{figure}[!htb] \centering \includegraphics[ width=0.96\linewidth, height=0.78\textheight, keepaspectratio ]{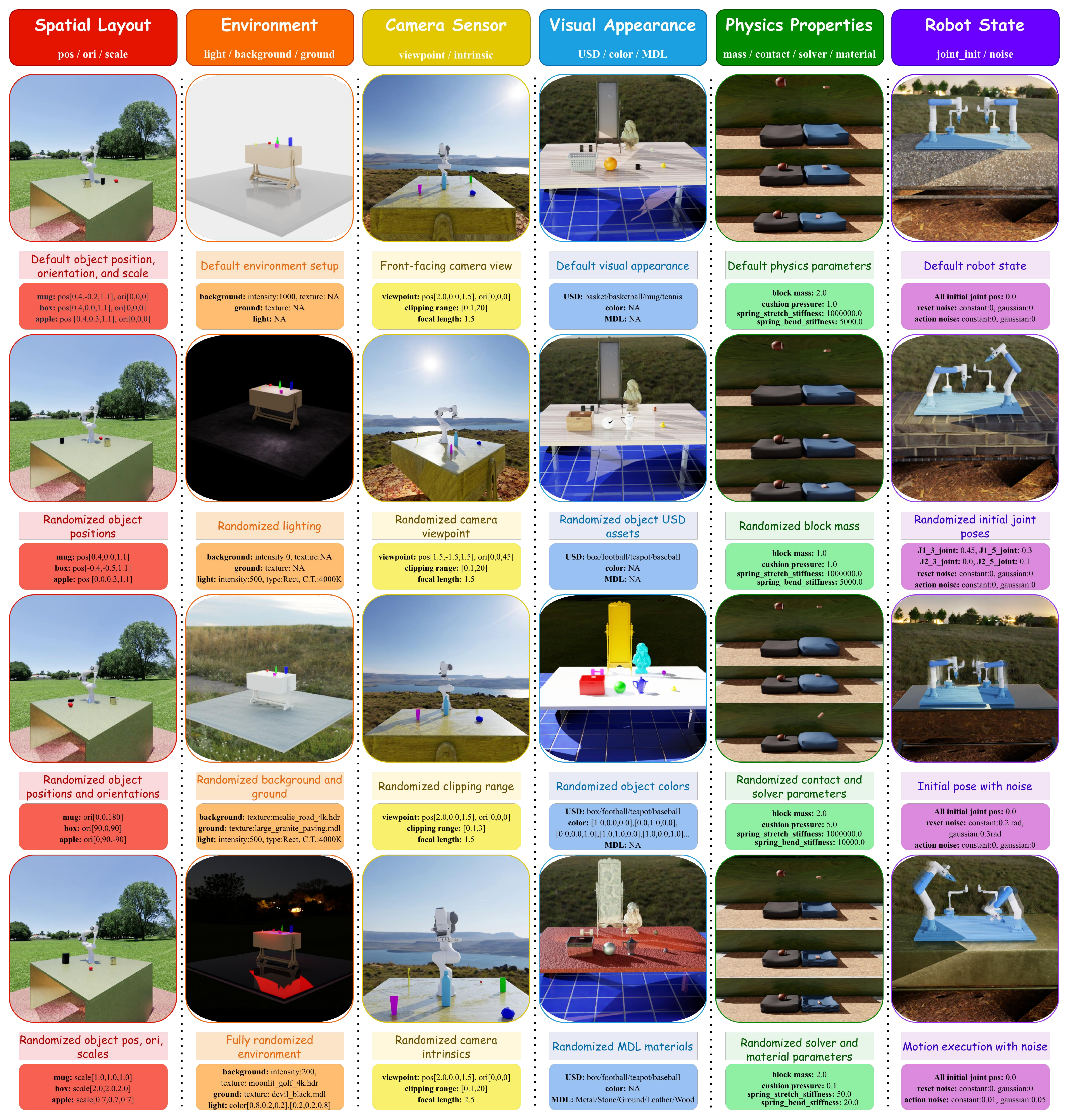} \caption[Domain-randomization axes supported by MagicSim.]{ \textbf{Representative domain-randomization axes supported by MagicSim.} From left to right, MagicSim randomizes spatial layout through object position, orientation, and scale; environment conditions through lighting, background, and ground appearance; camera sensing through viewpoint and intrinsic parameters; visual appearance through asset selection, color, texture, and MDL material; physical properties through mass, contact, material, and solver parameters; and robot state through initial joint configurations and execution noise. Each column contrasts a default configuration with representative independent or combined perturbations. Random draws are isolated by environment and manager, allowing any axis to be varied without changing unrelated random choices. } \label{fig:domain-randomization} \end{figure}

Figure~\ref{fig:domain-randomization} summarizes the principal randomization axes exposed by MagicSim. Together, these axes vary the geometry, visual observations, physical dynamics, sensing conditions, and embodiment state of an episode while preserving logical object identities, task definitions, and reproducible manager-local random streams. 

\subsection{Object Lifecycle Modes: P1 / P2 / P3}
\label{sec:lifecycle-modes}

The non-stopping simulator has its sharpest consequence during hard reset. A reset
enters the runtime through a fixed call order---layout plans poses, scene binds
objects to those poses, terrain refreshes only mutable appearance, and robot,
camera, and capture managers follow---but the hard-reset behavior of each object
is determined by its lifecycle mode. Because environments reset at different
times on a shared stage that keeps stepping, MagicSim cannot tear down and
rebuild the entire world between episodes.

Once \texttt{SimulationContext.reset()} has locked the USD stage, whether an
object can still be dynamically imported or dynamically deleted is a backend
capability, not a high-level runtime preference. A uniform ``delete everything and
rebuild'' path is therefore unavailable. MagicSim collapses each object's two
backend capabilities---dynamic import and dynamic delete---into three lifecycle
modes, and the scheduler dispatches on the mode rather than on the object type.

In \textbf{P1}, both import and delete are available. A hard reset can destroy the
current prim, create a replacement, and optionally choose a different asset. In
\textbf{P2}, import is available but delete is not. The old prim cannot be
removed, so the runtime retires it: gravity and collision are disabled, the prim
is hidden and moved aside, and the replacement is created at a fresh
\texttt{prim\_path}. This is why logical identity must be separate from physical
prim path. In \textbf{P3}, neither import nor delete is available after the stage
lock. The object must be built once before the lock and can only be reset in
place afterward.

This mode collapse is the runtime simplification. Heterogeneous simulation
families do not require one reset algorithm per family; each object declares its
backend lifecycle capabilities, and the scheduler maps them to P1, P2, or P3.
State restoration inherits the same limits: \texttt{reset\_to} can restore a P1
or P2 snapshot with a different selected asset because the prim can be rebuilt or
reallocated, whereas a P3 object can only restore pose and randomized state onto
the existing asset. Terrain is not a fourth mode: it is a locked carrier whose
reset-time changes are soft randomization of mutable appearance, so it honors the
deterministic contract without entering the hard-reset scheduler.

Step ordering is intentionally simpler: actions are applied, the batched
simulator advances one physics tick, and observation follows. Physics coupling,
control, and sensing are specified in their own sections; the runtime only fixes
the order in which they are invoked.
\FloatBarrier
%

\section{Embodiment Control and Motion Planning}
\label{sec:control}
\label{sec:control-planning}
\label{sec:planning-engine}
\label{sec:async-planning}

A MagicSim sub-environment can host several robots of different
morphologies at once---a humanoid and a wheeled manipulator sharing one
scene, each commanded independently---because the platform is built for
multi-agent embodied settings. That requirement shapes everything in this
section: a control stack written per robot class would not survive such
mixtures. Instead, every robot---seven categories, roughly thirty-three
registered embodiments---is reduced at configuration time to a uniform
channel interface, \texttt{base}, \texttt{arm}, \texttt{eef}, and driven
through one three-level stack: closed-loop control at the bottom, motion
planning in the middle, target supply at the top. Every layer above
addresses channels rather than robots, and the combined action space simply
concatenates the channels of whatever robots an environment declares, so a
heterogeneous team is configured no differently from a single arm. The
stack's planning level is powered by a forked cuRobo
engine~\cite{curobo2023}, rebuilt for multi-tool-frame,
per-environment-heterogeneous, asynchronous batched solving. One stack for
every embodiment; one engine behind the stack.

The running example previews both halves. Commanded to walk to the table and grasp
the tray with both hands, the G1 splits the command into a base path---a
route from the navigation mesh, a local planner converting it into velocity
commands, a learned whole-body controller realizing them---and an arm
path---a grasp skill submitting candidate end-effector poses to an
asynchronous planning service while every parallel environment keeps
stepping, the trajectory landing a few frames later. Part~A
(\S\ref{sec:control-robotmanager}--\S\ref{sec:control-high}) walks the stack
bottom-up after a framing map; Part~B
(\S\ref{sec:curobo-capabilities}--\S\ref{sec:curobo-fork}) opens the engine:
what the forked solver can solve, how it runs without blocking, what changed
at source level. Part~C closes with the two movers that are not robots. This
section owns the mechanisms; skills and the collection loop consume them
(Sections~\ref{sec:skills} and~\ref{sec:collection}).


\subsection{RobotManager and the Embodiment Taxonomy}
\label{sec:control-robotmanager}

\begin{figure}[htbp] \centering \includegraphics[width=0.98\linewidth]{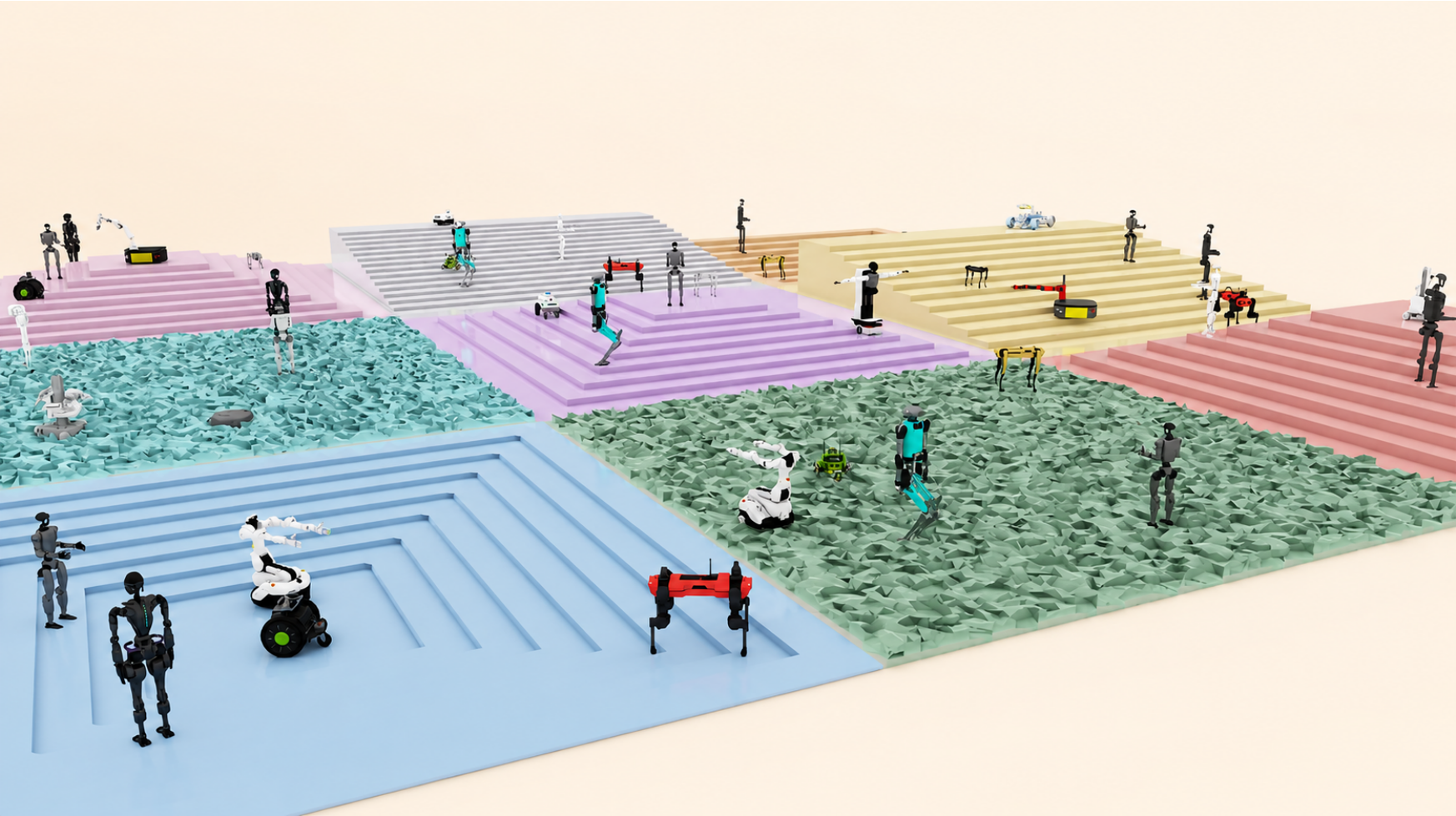} \caption[Representative robot embodiments supported by MagicSim.]{ \textbf{Representative robot embodiments supported by MagicSim.} The registered embodiment collection spans single-arm manipulators, dual-arm manipulators, dexterous-hand systems, mobile bases, mobile manipulators, humanoids, and quadrupeds. The figure shows a representative subset of the approximately thirty-three registered robots. Despite their heterogeneous morphologies and controllers, all embodiments are exposed to higher-level planners, skills, and task drivers through the shared base-arm-end-effector channel interface. } \label{fig:robot-embodiments} \end{figure}

\texttt{RobotManager} spawns every robot, builds its articulation and
per-robot action and observation managers, assembles one combined Gym space
across robots, and implements the deterministic-runtime contracts of
Section~\ref{sec:runtime} (seeding, noise, snapshot and replay). Seven
embodiment categories are supported---single-arm manipulators, dual-arm
manipulators, dexterous-hand manipulators, mobile bases, mobile
manipulators, humanoids, and quadrupeds---covering roughly thirty-three
registered robots, from tabletop arms such as the Franka and xArm to the
Unitree G1 and Go2; new embodiments enter through an import pipeline
without touching anything above. Figure~\ref{fig:robot-embodiments} visualizes a representative subset of the registered embodiments and the morphological breadth handled by RobotManager. 

What makes one stack serve all of them is the channel interface. At
configuration time every robot reduces to at most three action
channels---\texttt{base}, \texttt{arm}, \texttt{eef}---each filled by an
action term from a shared library (\S\ref{sec:control-low}) or absent where
it does not apply: a tabletop manipulator has no \texttt{base}, a wheeled
base no \texttt{arm}, and the G1 fills \texttt{base} with a learned
whole-body controller rather than a wheel model. Planners, skills, and the
three drivers of Section~\ref{sec:overview-one-mdp-three-drivers} all
address channels, so adapting to a new embodiment means declaring which
term fills each channel---no layer above changes.

\subsection{The Three-Level Control Stack}
\label{sec:control-stack}

\begin{figure}[htbp]
\centering
\includegraphics[width=\textwidth]{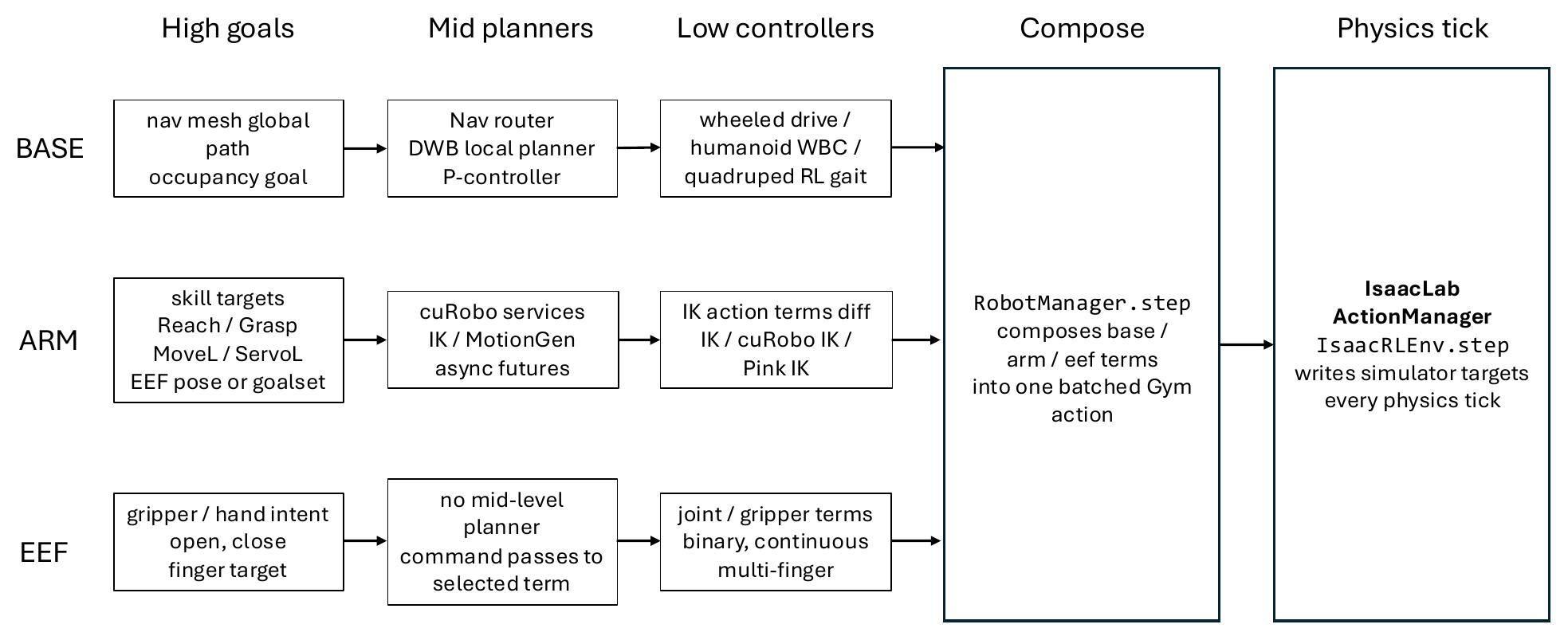}
\caption{The three-level control stack. Commands flow left to right per channel
and converge in the action manager; cuRobo appears as low-level action term,
as mid-level services, and as the shared engine behind both (Part~B).}
\label{fig:control-stack}
\end{figure}

Figure~\ref{fig:control-stack} is the map of Part~A. Horizontally, three
levels: low-level closed-loop controllers write the simulator every physics
tick; mid-level planners turn goals into commands (a local navigation
planner for \texttt{base}, the cuRobo services for \texttt{arm}); the high
level supplies the goals themselves. Vertically, the channel decomposition
of \S\ref{sec:control-robotmanager}; everything reconverges in the IsaacLab
action manager each tick. Three properties anchor the map. The high level is
symmetric in role but not in location: navigation belongs to this section,
while arm-side targets come from the skill layer
(Section~\ref{sec:skills}). Collision awareness is a \emph{level} property:
the mid-level services plan against per-environment scene collision, the
low-level cuRobo term checks only self-collision and joint limits. And
cuRobo appears three times---low-level action term, mid-level services, and
the shared engine of Part~B behind both. The figure reads left to right, as a
command flows; the prose proceeds right to left.

\subsection{Low Level: Closed-Loop Controllers}
\label{sec:control-low}

Low-level controllers share one contract---consume a command, write the
simulator every tick---and differ in where their control knowledge comes
from: direct mapping, vehicle modeling, optimization, or reinforcement
learning. Joint and gripper terms map commands directly into joint limits
and span binary, continuous, and multi-finger end effectors. Drive-model
terms encode vehicle kinematics---differential drive, Ackermann steering,
and holonomic bases---turning a body-frame velocity command into wheel
actuation; the quadruped term shares the same $(v,\omega)$ interface, but
its backend is a learned gait, so the stack above cannot tell wheels from
legs. Optimization enters through three inverse-kinematics architectures:
Jacobian differential IK, fast and collision-blind, whose dual-arm variant
stacks two independent chains; cuRobo batched IK, which solves all
environments in one call of the forked engine, tracks every declared tool
frame so a single term drives single-arm, dual-arm, and humanoid upper
bodies alike, and is the only term with built-in self-collision and
joint-limit handling (scene-aware collision lives at the service level);
and Pink IK~\cite{pink}, a QP over weighted frame and posture
tasks and the route for \emph{coupled} dual-arm and upper-body solving,
whose humanoid subclass fills a shared arm-reference buffer without writing
the simulator. A fourth cuRobo mode, trajectory generation, is not a
per-tick term and belongs to the mid level. Reinforcement learning closes
the family: the quadruped's (\texttt{go2}) locomotion policy is trained
with IsaacLab's built-in RL workflows~\cite{mittal2023orbit}, and the G1's
whole-body controllers are selected through a policy
factory---\texttt{homie\_v2} following HOMIE~\cite{ben2025homie};
\texttt{sonic\_v1}, which realizes the upper-body reference the Pink term
proposes; and an Agile lower-body
variant~\cite{zhao2026agilecomprehensiveworkflowhumanoid}. The family thus
ends where its storyline was headed: optimization proposing, a learned
policy realizing, inside one robot.

\subsection{Mid Level: Motion Planning}
\label{sec:control-mid}

\texttt{PlannerManager} spans two problem families behind one combined
action space---base navigation and manipulation---composing planners the
way \texttt{RobotManager} composes action terms, from the robot's
\texttt{planner:} block. The base side carries the mechanisms. A Dynamic
Window planner~\cite{fox1997dwa} samples velocity candidates within limits,
rolls each forward, scores the rollouts against goal, path, and an obstacle
cost read from the occupancy map, and emits the best command, with a
turn-only proportional term near the goal; four variants---holonomic,
differential, humanoid, quadruped---differ only in output width and padding,
matching the \texttt{base}-channel consumers of \S\ref{sec:control-low}. A
P-controller handles direct navigate-to-pose and in-place turning, and the
\texttt{Nav} router places both behind one channel, a trailing mode flag
selecting the controller or forwarding waypoints to the dynamic-window
planner. On the arm side only the interface appears here: four services per
robot---\texttt{IKServer}, \texttt{DualIKServer}, \texttt{MotionGenServer},
\texttt{DualMotionGenServer}---accept submissions and resolve futures to
\texttt{(success, goalset\_index, env\_ids)} or
\texttt{(actions, success, env\_ids)}. The engine behind them is Part~B.

\subsection{High Level: Navigation and End-Effector Targets}
\label{sec:control-high}

The high level supplies goals and introduces no mechanism. On the base
path, a baked navigation mesh answers global-path queries and the route is
handed to the \texttt{Nav} router, while the occupancy map supplies the
obstacle model; both are consumed here, their construction belonging to the
engine, runtime, and annotation chapters
(Sections~\ref{sec:engine}, \ref{sec:runtime}, \ref{sec:annotation}). On
the arm path, skills such as \emph{Reach}, \emph{Grasp}, \emph{MoveL}, and
\emph{ServoL} emit target end-effector poses or goalsets, a hand selector
designating which arm of a bimanual robot moves; their state machines are
the subject of Section~\ref{sec:skills}. Part~A thus closes as it opened:
global path $\rightarrow$ router $\rightarrow$ velocity command
$\rightarrow$ drive model or whole-body controller; end-effector target
$\rightarrow$ asynchronous solve $\rightarrow$ trajectory $\rightarrow$ IK
term. Every embodiment of \S\ref{sec:control-robotmanager} instantiates a
subset, and nothing above the channels knows which.


\subsection{cuRobo Solver Capabilities}
\label{sec:curobo-capabilities}

The mid level holds only because the arm-side planner does what stock
cuRobo cannot. Stock cuRobo solves for one end effector, one goal, a
homogeneous batch; MagicSim's worlds are \emph{heterogeneous}---many
morphologies, candidate grasps rather than single poses, batches in which
every environment wants something different---and its tasks demand
\emph{precision}, from bimanual grasps that must be jointly reachable to
trajectories that must not wander. The fork's additions divide along
exactly these two axes, turning cuRobo from a single-end-effector planning
library into a multi-embodiment, multi-tool-frame, asynchronous planning
engine for parallel simulation. How the additions run without blocking is
\S\ref{sec:curobo-async}; where they live in source is
\S\ref{sec:curobo-fork}.

On the heterogeneity axis, the pose objective generalizes from one end
effector to an ordered set of named tool frames, so a single solver serves
every morphology of \S\ref{sec:control-robotmanager}; goals extend to
per-frame \emph{goalsets} of candidate poses; criteria become assignable
per batch slot, a NaN-marked frame simply dropping out, so one batched
solve moves different arms in different environments; mobile manipulators
get both problem formulations---base locked, or base free through virtual
planar joints---as the ``dual'' services of \S\ref{sec:control-mid}; and a
retargeter on the same objective carries motion across embodiments.

\begin{figure}[htbp] \centering \includegraphics[ width=0.98\linewidth ]{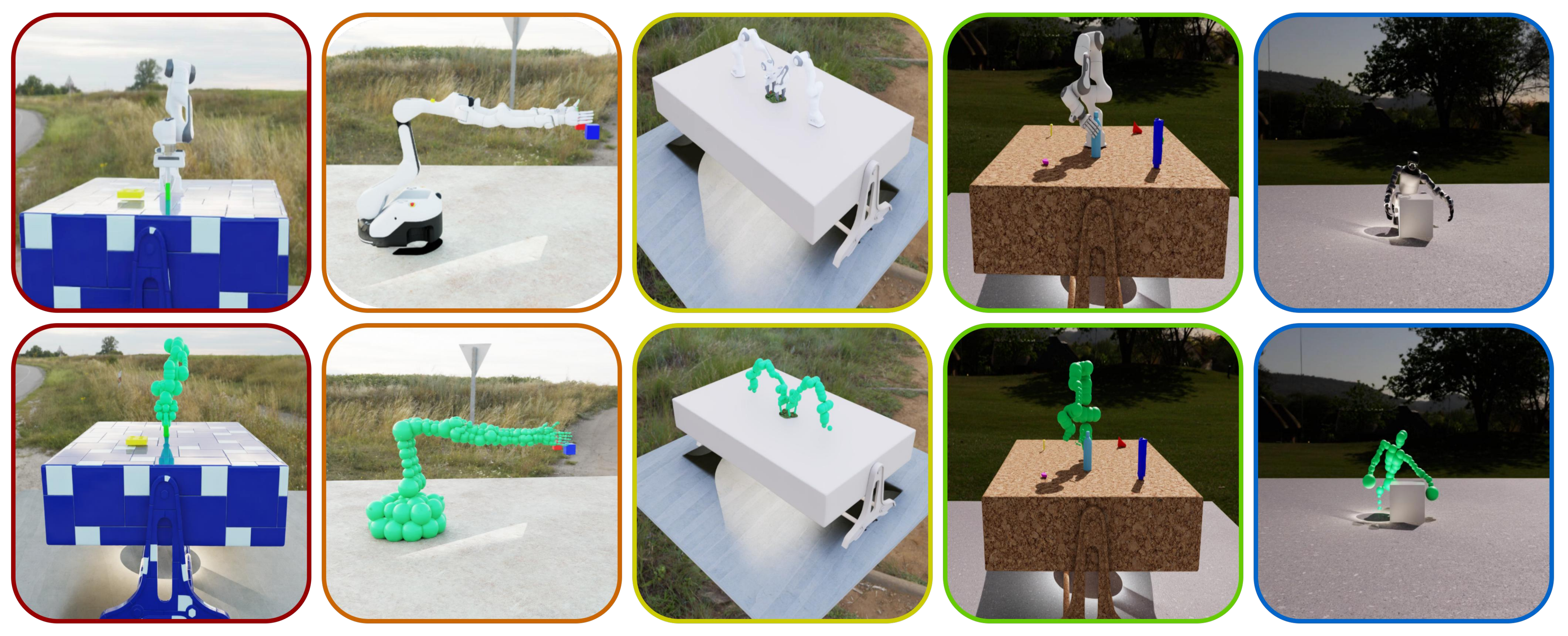} \caption[cuRobo collision-sphere models across robot embodiments.]{ \textbf{Representative cuRobo collision-sphere models across robot embodiments.} The upper row shows the rendered robot geometries, while the lower row visualizes the corresponding link-attached collision spheres used by MagicSim's cuRobo configuration. The examples span heterogeneous morphologies, including stationary manipulators, mobile manipulators, humanoid or whole-body systems, and other articulated embodiments. The overlapping spheres provide a compact approximation of the robot's occupied volume for GPU-parallel collision-distance queries during inverse kinematics and motion generation. At the low-level action-term interface, cuRobo uses the robot model for self-collision and joint-limit handling; at the mid-level service interface, the same representation is additionally evaluated against per-environment scene obstacles for collision-aware trajectory generation. } \label{fig:curobo-collision-spheres} \end{figure}

A common robot-side collision representation is required before the same planning engine can serve heterogeneous embodiments. Figure~\ref{fig:curobo-collision-spheres} shows representative link-attached collision-sphere models used by the cuRobo planning stack. These compact geometric approximations make collision-distance queries suitable for batched GPU execution while preserving the articulated structure and relevant occupied volume of each robot. 

On the precision axis, \emph{paired} selection---the default for
IK---makes all frames share one jointly evaluated candidate, so a bimanual
grasp is reachable as a pair rather than per arm in isolation; and
\emph{seed anchoring} clusters every seed around the live configuration,
eliminating the visible detours that workspace-scattered seeds produce,
enabled by default exactly where detours hurt most.

\subsection{Async Microbatch Solve-Farm}
\label{sec:curobo-async}
\label{sec:async-solve-farm}

\begin{figure}[!htb]
\centering
\includegraphics[width=\textwidth]{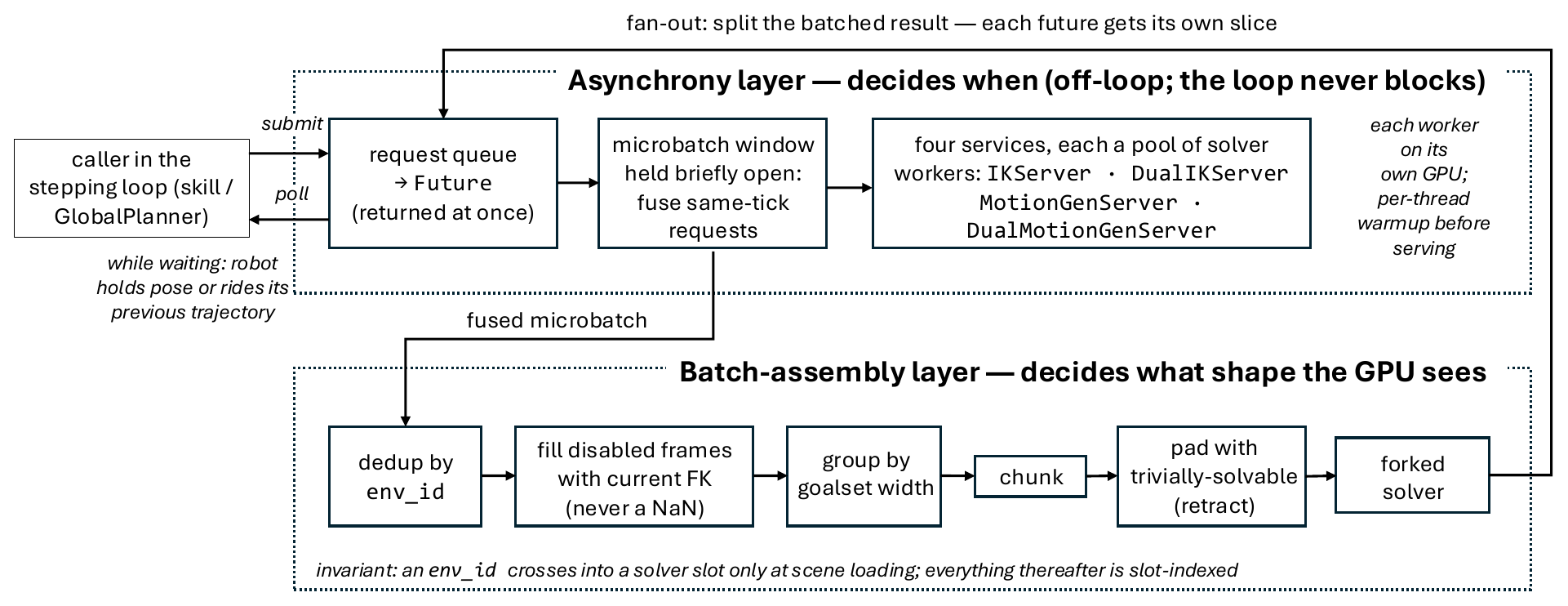}
\caption{One request's path through the solve-farm: the asynchrony layer
decides \emph{when} solving happens without blocking the loop; the
batch-assembly layer decides \emph{what shape} the GPU sees.}
\label{fig:curobo-request-flow}
\end{figure}

These capabilities run through four in-process services sharing one runtime
(Figure~\ref{fig:curobo-request-flow}). The design takes its cue from how
SGLang~\cite{zheng2024sglang} serves rollout generation in LLM
reinforcement learning: planning, like generation, becomes an inference
service---submitted to asynchronously, microbatched across callers, and
free to live on its own GPUs---rather than a library call inside the
stepping loop. The runtime answers two independent questions.
\emph{Time}: a solve takes many milliseconds, the loop steps all
environments each tick, and one synchronous solve would stall the batch.
\emph{Shape}: requests arrive ragged---different environments, frames, and
goalset sizes---while the GPU wants one fixed-shape problem. This
subsection is the sole owner of both mechanisms; the collection loop
consumes them only as submit, keep stepping, harvest
(Section~\ref{sec:collection-async}).

On the time side, callers never block: a submission returns a future at
once, the robot holds pose or rides out its previous trajectory, and the
result is consumed when polling finds it ready. Each service is a pool of
solver workers, each bound to a GPU of its own---the solvers need not share
a card with the simulator, and multiple instances spread across multiple
GPUs---and a worker holds its microbatch window briefly open so
that requests fired by other environments on the same tick fuse into one
batched solve. One detail is load-bearing for correctness rather than
throughput: every worker warms its solver up on its own thread before
serving, because CUDA contexts are per-thread and a late first touch would
invalidate graphs the main thread had already captured---including the
action-term IK graph of \S\ref{sec:control-low}.

On the shape side, a normalization pipeline turns the ragged harvest into
the fixed-shape solve: requests are deduplicated by environment, disabled
frames are filled in with their current forward kinematics so the solver
never observes a NaN, the batch is grouped by goalset width, chunked, and
padded with trivially solvable problems, and the result fans back out, each
future receiving exactly its slice. A single invariant keeps the
bookkeeping safe: scene loading is the only place an environment id crosses
into a solver slot; everything thereafter is slot-indexed.

\subsection{Fork Source and Isaac Sim Integration}
\label{sec:curobo-fork}

None of this exists upstream: the capabilities are source-level additions
to a vendored fork, recorded in Table~\ref{tab:curobo-fork-map}. Beyond these
features, the fork is kept viable as an engineering artifact. It tracks
cuRobo's v2 API and installs as an editable workspace member; the
robot-configuration schema gains tool frames, FK-only links, and virtual
base joints; 

\begin{table}[htbp]
\centering
\small
\begin{tabular}{l l}
\toprule
Capability (\S\ref{sec:curobo-capabilities}) & Implementation site in the fork \\
\midrule
Multi-frame tool poses       & tool-pose types, cost terms, Warp kernels \\
Per-env criteria \& disable  & per-slot criterion hooks on the IK solver \\
Paired selection             & paired-mode flag; paired per-env kernel \\
Per-frame goalsets           & $(B,L,G,7)$ goal/cost path; kernels per width \\
Seed IK \& anchoring         & seed-solver module; motion-planner patch \\
Cross-embodiment retargeting & retargeter module on the same objective \\
Robot-config extensions      & tool frames, FK-only links, base joints, exclusions \\
\bottomrule
\end{tabular}
\caption{Capability-to-source map for the cuRobo fork; semantics live in
\S\ref{sec:curobo-capabilities}.}
\label{tab:curobo-fork-map}
\end{table}


\subsection{Flying-Camera Planning}
\label{sec:flying-camera}

Two movers in MagicSim are not robots. The first is the flying camera:
planning without physics, used by the camera-only environment to smooth
capture motion between target poses through an eased linear model and a
gimbal model that orbits or tracks a subject. Beyond smooth capture, this
is the natural instrument for two data regimes: free-viewpoint exploration
footage for spatial agents that must reason about a scene from limited
views~\cite{wang2025mindcube}, and controlled multi-view recordings of
physical events for physics-consistent world
modeling~\cite{lu2026phys4d}. The planner decides the pose
sequence, the camera manager writes it, the capture manager records through
it; the planners remain here because they implement the same step interface
that \texttt{PlannerManager} composes (\S\ref{sec:control-mid}). Their use
in producing camera-trajectory data is described in
Section~\ref{sec:collection-camera}.

\begin{figure}[htbp] \centering \includegraphics[width=0.98\linewidth]{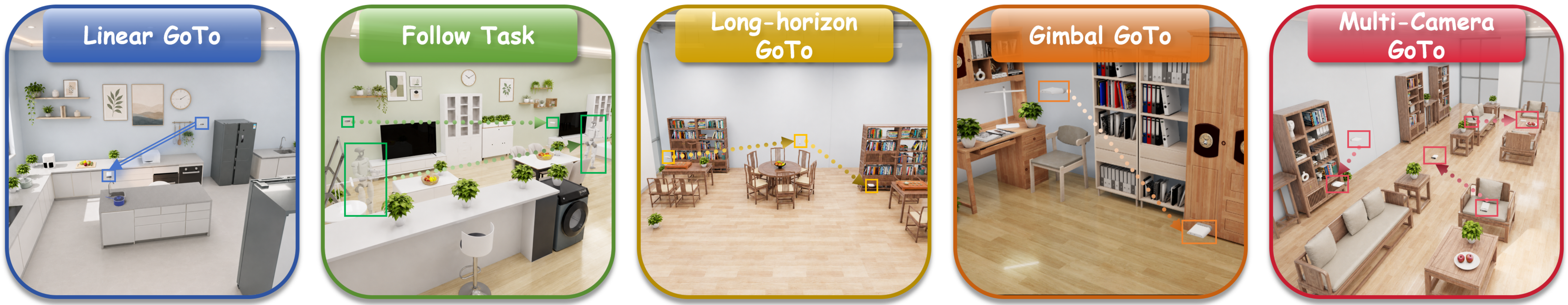} \caption[Flying-camera planning modes in MagicSim.]{ \textbf{Flying-camera planning modes in MagicSim.} Camera-only embodiments support eased linear GoTo motion, task-following trajectories, long-horizon waypoint sequences, gimbal-based target tracking or orbiting, and coordinated multi-camera GoTo. The planner generates the camera-pose sequence, CameraManager applies the poses during simulation, and CaptureManager records the resulting visual trajectory. These modes support free-viewpoint scene exploration and controlled multi-view recording of physical events. } \label{fig:flying-camera} \end{figure}

Figure~\ref{fig:flying-camera} summarizes the camera-motion modes implemented through the common planner interface. 

\subsection{Avatar Animation and Control}
\label{sec:avatar-control}

The second is the animated human avatar: a world entity of the simulation
engine (Section~\ref{sec:engine}), driven by an animation graph rather than
physics, to which none of Part~A applies. Both control routes exist.
Avatars play \emph{externally supplied motion}---skeleton-animation clips
behind a registry of actions (idle, sit, talk, go-to, custom commands),
driven through a per-avatar action queue---and they are \emph{directly
controllable through IK}: four modes span full-clip animation, an
IK-over-locomotion control mode whose upper--lower split at the spine lets
the avatar reach, look, and bend while a motion-matched lower body keeps
walking, a path-walking mode, and a pose-provider mode that streams
arbitrary joint targets for retargeted motion. The stack was validated end
to end by a teleoperated whole-body animation demonstration. The contrast
with \S\ref{sec:control-low} is the point of the placement: the G1's
whole-body controllers are physics-level robot control inside the action
manager loop, while avatars are animation-graph puppetry---two
animated-agent stacks kept apart precisely so they are never confused.

\begin{figure}[htbp] \centering \includegraphics[width=0.98\linewidth]{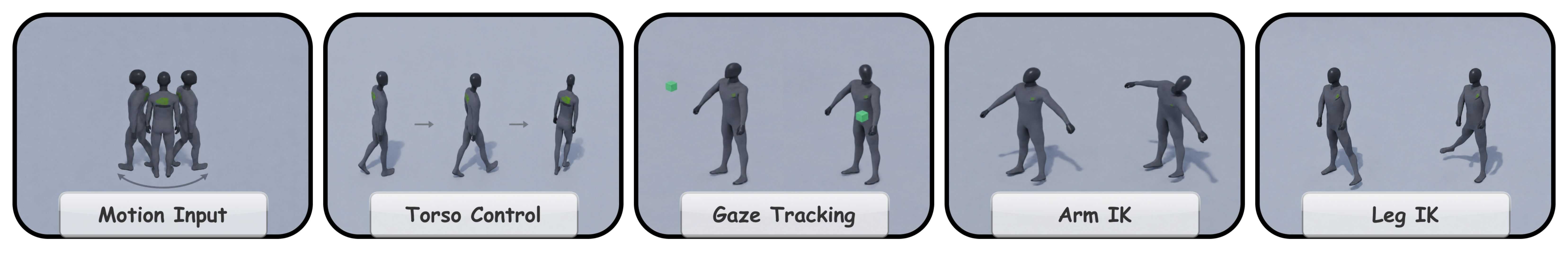} \caption[Avatar control and IK modes in MagicSim.]{ \textbf{Avatar control and IK modes in MagicSim.} The figure summarizes representative control modes supported for animated human avatars. \emph{Motion Input} plays externally supplied motion or action-queue commands through the animation graph. \emph{Torso Control} supports upper-body steering over locomotion, allowing the avatar to bend or reorient the torso while the lower body continues walking. \emph{Gaze Tracking} lets the avatar look toward a designated target. \emph{Arm IK} controls upper-limb reaching through inverse kinematics. \emph{Leg IK} adjusts lower-limb pose targets for stance adaptation and motion retargeting. Together, these modes provide a controllable avatar stack that is distinct from physics-level robot control: the avatar is an environment-side animated actor driven by an animation graph and IK interfaces, rather than by robot action-space commands. } \label{fig:avatar-ik-control} \end{figure}

Figure~\ref{fig:avatar-ik-control} summarizes the principal control surfaces exposed by MagicSim for animated avatars. These interfaces cover both motion-driven and inverse-kinematics-driven control, allowing avatars to be used as environment-side actors in HRI scenes, scenario scripting, teleoperation demonstrations, and avatar-conditioned task generation. 
\FloatBarrier
%

\section{Sensor Simulation and Multimodal Observation}
\label{sec:sensors}

MagicSim treats sensing as part of the runtime. A first-class sensor is not just
a USD prim attached to a robot; it is a manager-owned stream that is synchronized
with reset, batched across environments, and exposed to controllers, planners,
tasks, or the data engine through the same runtime contract as the rest of the
simulator.

This gives the chapter a simple organization. MagicSim inherits broad sensor
availability from the Isaac/USD stage, but it deeply integrates the sensing
families that the system actually uses: visual streams, LiDAR/IMU embodiment
sensing, tactile contact, and the geometric signals that planners and annotators
consume. Radar and other autonomous-driving-style or vendor-specific sensors
remain stage-compatible, but they are not the focus of this manipulation-first
report.

\subsection{Sensor Stack Overview}
\label{sec:sensors-stack}

Table~\ref{tab:sensors-stack} summarizes the stack by sensor family. The table
does not ask whether a device can be mounted on the stage; it asks what MagicSim
owns as a runtime stream.

\begin{table}[htbp]
\centering
\small
\caption{MagicSim sensor families. Manager-owned channels participate in the
runtime contract; inherited channels remain attachable through the underlying
USD/Isaac stage but are not the focus of this report.}
\label{tab:sensors-stack}
\begin{tabularx}{\linewidth}{@{}p{0.22\linewidth}p{0.27\linewidth}X@{}}
\toprule
\textbf{Family} & \textbf{Manager-owned channels} & \textbf{Role in MagicSim} \\
\midrule
Visual sensing &
Camera / tiled capture &
Batched RGB, depth, normal, and related image streams for data collection,
policy observation, and annotation. \\

Range and inertial sensing &
RTX LiDAR / IMU &
Manager-owned embodiment sensors used by the G1 locomotion stack for range and
body-state sensing. \\

Contact and tactile sensing &
Taxel contact, probe tactile, visuotactile &
Contact-rich manipulation signals on robot links: force, pressure, deformation,
privileged contact fields, and tactile images. \\

Geometric support sensing &
Frame tracking, runtime occupancy, navmesh query &
Geometry used by IK, navigation, annotation, and language grounding. \\

Inherited stage sensors &
RTX Radar, effort, proximity, joint-force, PhysX range/light-beam, ultrasonic,
vendor assets &
Attachable through the USD/Isaac stage. Radar is intentionally out of focus
because MagicSim does not claim an autonomous-driving perception stack. \\
\bottomrule
\end{tabularx}
\end{table}

One boundary is important. Joint positions, velocities, and other articulation
observations belong to the robot and task observation space
(Section~\ref{sec:tasks}); MagicSim does not treat them as a separate sensor
module. Kinematic frame sensing, however, is part of the sensor stack and is
covered in Section~\ref{sec:sensors-geometric}.

\subsection{Visual, Range, and Inertial Sensing}
\label{sec:sensors-visual-range}

The visual path is MagicSim's main imaging substrate. A camera viewpoint is
world state: it can be fixed in the scene, mounted on a moving embodiment,
randomized per environment from the seeded streams of
Section~\ref{sec:dr-seed-streams}, restored during replay, or flown along
planner-generated trajectories for camera-only collection
(Section~\ref{sec:flying-camera}). Camera placement therefore participates in
the same reproducibility story as objects, robots, and layouts.

MagicSim separates pose ownership from render ownership. The camera manager owns
camera prims and their poses. The capture manager turns those prims into image
streams. This split lets mounting, randomization, and replay remain independent
from render-product construction. It also makes tiled capture a batching
strategy rather than a different sensor API: the same camera configuration can
be rendered environment-by-environment or as a tiled batch while exposing the
same per-environment outputs downstream.

The image streams produced here are the substrate for later supervision, not the
annotation system itself. Raw RGB, depth, normals, and motion-like channels live
in this chapter; object boxes, end-effector trails, affordance labels, and saved
annotation schemas live in Section~\ref{sec:annotation}. This keeps the sensor
chapter focused on observation streams and leaves semantic packaging to the
asset and annotation chapter.

The same runtime also owns range and inertial sensing for locomotion. On the G1
embodiment, RTX LiDAR and IMU are manager-owned sensor streams used by the
locomotion stack. Their role is different from the manipulation camera/tactile
data engine: LiDAR provides ray-traced range sensing for navigation scenes, and
the IMU provides body-state information for humanoid locomotion. Both belong to
MagicSim's integrated sensor stack; Radar remains inherited and out of focus
because the report does not claim an automotive Radar perception pipeline.

Where possible, the hot path is GPU-enabled. Tiled capture uses batched GPU
rendering buffers, RTX cameras and LiDAR share the ray-tracing backbone, and the
downstream tactile and occupancy streams below keep their high-rate data in
device-resident tensors.

\subsection{Contact Backends and Surface Queries}
\label{sec:sensors-contact}

Touch begins as geometry. Before a tactile image, force field, or pressure map
exists, the runtime must answer a taxel-level question: is this point on the
sensor surface in contact, how deep is the contact, and along which local
direction should the signal be measured? MagicSim factors this into a reusable
contact layer shared by probe tactile and visuotactile sensors.

Every tactile surface is first sampled into taxels. Each taxel carries its own
local frame: an undeformed point, an outward normal, and tangent directions.
The contact backend then fills the same interface regardless of how contact is
computed. MagicSim provides three backends:

\begin{table}[htbp]
\centering
\small
\caption{Contact backends. All backends expose the same taxel-level interface;
the choice is about geometry correctness and runtime constraints.}
\label{tab:contact-backends}
\begin{tabularx}{\linewidth}{@{}p{0.18\linewidth}p{0.30\linewidth}X@{}}
\toprule
\textbf{Backend} & \textbf{When to use it} & \textbf{Why} \\
\midrule
\texttt{sdf\_warp} &
Portable flat or general contact &
Queries a Warp mesh signed-distance field. It needs no pre-baked field and is
the default general-purpose path. \\

\texttt{sdf\_physx} &
High-throughput CUDA SDF contact &
Uses a pre-baked PhysX signed-distance representation. It is the faster SDF
path when the asset and device support it. \\

\texttt{raycast} &
Curved tactile skins &
Casts along each taxel's own outward normal. This is the geometry-correct path
for curved sensors, where SDF contact follows the object normal and can produce
wrong shear or deformation directions. \\
\bottomrule
\end{tabularx}
\end{table}

The key distinction is not speed alone; it is the contact direction. SDF queries
are useful and efficient for flat pads and many privileged contact signals, but
they naturally reason from the object's closest surface and its normal. On a
flat tactile pad this is often acceptable. On a curved fingertip it is not: the
elastomer should compress along the sensor taxel's own normal, not along the
object normal. Otherwise the simulator can fabricate spurious tangential motion
or shear on the curved skin. For this reason, MagicSim treats ray casting as the
canonical backend for curved tactile sensors.

This also explains why ray casting is not presented as a standalone sensor type.
It is a geometric primitive used by multiple systems: tactile contact uses it at
the taxel level, occupancy and nav queries use geometric tests for planning, and
affordance annotation uses ray hits to connect planner intent to object geometry.
The sensor claim lives in the manager-owned streams that consume the primitive.
Body-level rigid contact reporting remains available through the underlying
physics stack; MagicSim's contribution here is the taxel-level contact interface
above it.

\subsection{Tactile Transduction}
\label{sec:sensors-tactile}

Tactile hardware looks diverse: pressure pads, force probes, flat
GelSight-style sensors, and curved visuotactile fingertips. MagicSim organizes
them into a small number of routes. A tactile sensor is configured on a robot
link, owned by the tactile manager, and emits per-environment batched outputs
each step. The difference between sensors is how contact geometry becomes a
signal.

MagicSim supports two broad tactile families.

The first is the \emph{probe} family. A probe does not need a camera: any
pose-trackable link can become a taxelized contact surface. The same contact
query can be transduced into per-taxel force, scalar pressure, penetration
depth, or deformation maps. This route is important not only for deployable
observations, but also for privileged information. In simulation, probe tactile
can expose dense contact fields that are difficult or impossible to measure on
hardware, making it useful for asymmetric actor-critic training, reward
debugging, contact-rich task analysis, and dataset annotation. Depending on the
task, these signals can be logged as privileged state, used to shape rewards, or
distilled into policies that observe only camera or tactile images.

The second family is \emph{visuotactile}. Here, contact geometry becomes an
image. Flat GelSight-family sensors reuse the camera substrate: an
in-elastomer camera observes deformation, and a calibrated optical renderer
turns it into a tactile RGB image, following the GelSight/Taxim line of work
\cite{yuan2017gelsight,si2022taxim}. Curved visuotactile sensors use the same
manager and contact stack, but their geometry is sampled on a curved surface and
resolved through the taxel-frame contact path above. ShARPA is one example of
this curved family, not the definition of it: the general capability is curved
tactile support through per-taxel frames, raycast contact, and a renderer that
maps the resulting deformation into a raw-like tactile image.

The routes can be summarized as follows:

\begin{table}[htbp]
\centering
\small
\caption{Tactile routes in MagicSim. The routes share the same manager runtime
but differ in how contact becomes a sensor signal.}
\label{tab:tactile-routes}
\begin{tabularx}{\linewidth}{@{}p{0.24\linewidth}p{0.30\linewidth}X@{}}
\toprule
\textbf{Route} & \textbf{Input} & \textbf{Output} \\
\midrule
Probe tactile &
Taxel contact from SDF or raycast backends &
Force, pressure, penetration, or deformation maps; often useful as privileged
simulation information. \\

Flat visuotactile &
Camera-observed elastomer deformation &
GelSight-style tactile RGB through a calibrated optical renderer. \\

Curved visuotactile &
Curved-surface taxels and deformation maps, typically using raycast contact &
Raw-like tactile images or geometric tactile state for curved fingers and pads. \\
\bottomrule
\end{tabularx}
\end{table}

This compositional view\cite{Ye2025LearningTF} is the tactile story of MagicSim. TacSL, Taxim,
FlexiTac, Tacmap, and ShARPA-style sensors are not separate silos; they are
different routes through the same manager-owned tactile stack
\cite{akinola2024tacsl,si2022taxim,huang2026flexitac,su2026tacmap}. The system
therefore supports both simple privileged contact probes and image-like tactile
sensors without changing runtime abstractions.

\begin{figure}[htbp]
\centering
\includegraphics[width=0.98\linewidth]{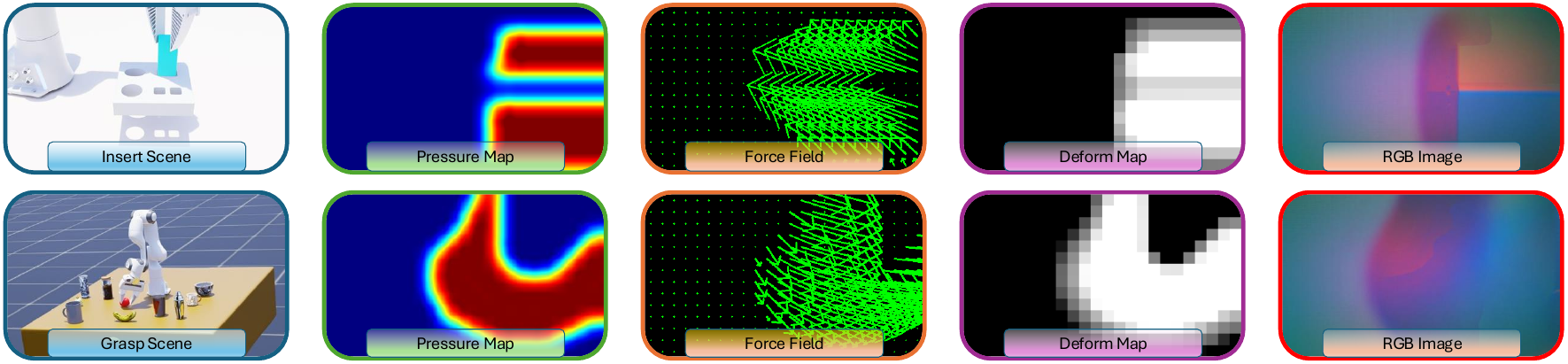}
\caption[Tactile sensing routes in MagicSim.]{
\textbf{Tactile sensing routes in MagicSim.}
MagicSim converts taxel-level contact geometry into both probe tactile fields and visuotactile images through a shared manager-owned runtime stack. Flat tactile pads can use SDF-style contact queries, while curved fingertips use raycast contact along each taxel's local normal to preserve the correct deformation direction. The resulting force, pressure, penetration, deformation, and tactile-image streams are synchronized with batched simulation and can be consumed by policies, planners, task predicates, and annotation pipelines.
}
\label{fig:tactile-sensor-routes}
\end{figure}

The boundary is deliberate. MagicSim does not place full finite-element
elastomer simulation in the hot path. For the data engine, the preferred
tradeoff is fast geometric contact plus scalar, deformation, or optical
transduction. More expensive soft-body tactile simulation is left outside the
main runtime path (Section~\ref{sec:limitations}).

\subsection{Geometric and Kinematic Sensors for Planning and Annotation}
\label{sec:sensors-geometric}

Not every sensor is meant to be rendered as an image or saved as a policy
observation. Some sensors exist so that the runtime can act and annotate.
MagicSim uses three manager-owned geometric channels as the eyes of planners,
annotators, and language grounding: frame tracking, runtime occupancy, and
navmesh query.

Frame tracking reports the world pose of named frames, especially end effectors.
The same stream feeds inverse-kinematics targets
(Section~\ref{sec:control-mid}), end-effector trail annotations, and the spatial
facts used by the language and annotation systems
(Section~\ref{sec:annotation}). This is kinematic sensing rather than
proprioception: the robot articulation still owns joint positions and velocities,
but the sensor stack owns the world-frame facts that other systems need.

Navigation uses two complementary geometric representations. The navmesh is the
global representation. During scene setup, MagicSim bakes the traversable parts
of the room or open world into a navigation mesh using the agent's physical
constraints, such as height, radius, step height, and slope limit. This converts
dense scene geometry into a sparse walkable manifold. The NavMeshManager then
queries that manifold: it can snap a target to the nearest valid point, test
whether a point is navigable, and compute a shortest path over the baked mesh.
Those paths provide global waypoints for mobile bases and humanoid locomotion.

The occupancy scan is the local representation. At runtime, OccupancyManager
scans a bounded region around each environment into a three-valued grid:
occupied, unoccupied, or unknown. This grid is not a static room asset; it is a
live observation of the current scene state. It is also the GPU-enabled hot path
for local obstacle reasoning: the multi-environment occupancy output is kept as
device-resident tensors and consumed by the DWB planner as the obstacle cost
field. DWB then rolls out candidate velocity commands and scores them against
the global path, the goal, and the local occupancy cost
(Section~\ref{sec:control-high}).

The distinction is therefore simple. The navmesh answers global questions:
where can the robot go, and what path should it roughly follow? The runtime
occupancy grid answers local questions: what obstacles are present around the
robot now, and which short-horizon command is safe? A static room occupancy map,
when available, belongs to the asset and annotation system as a world prior
(Section~\ref{sec:annotation}); the runtime occupancy scan belongs to sensing.
In short, the static map says what the world is expected to be, while the
runtime scan says what the robot sees now.

Taken together, these channels define MagicSim's sensor story. The platform
inherits breadth from USD/Isaac stage compatibility, but spends its deep runtime
integration where the system needs synchronized, batched, and GPU-ready
observations: camera and LiDAR streams, IMU state, taxel-level contact, tactile
transduction, runtime occupancy, and the geometry that lets planners and
annotators understand manipulation scenes.

\begin{figure}[htbp] \centering \includegraphics[ width=0.98\linewidth ]{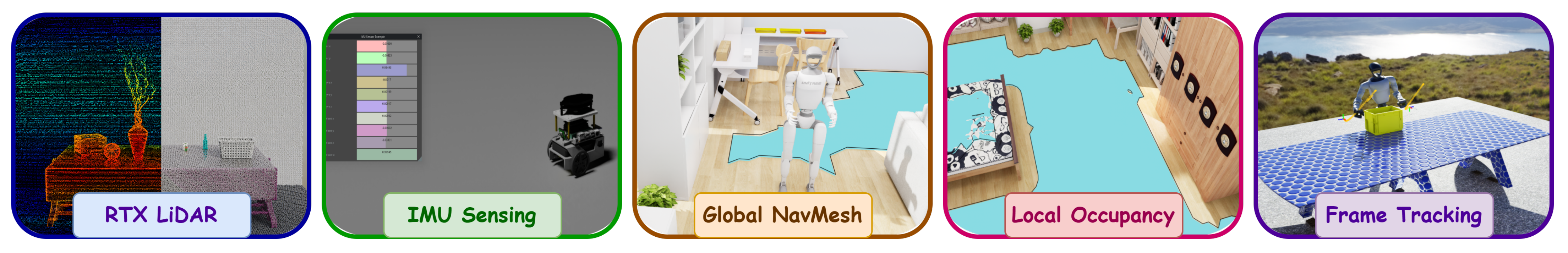} \caption[Range, inertial, and geometric support sensing in MagicSim.]{ \textbf{Range, inertial, and geometric support sensing in MagicSim.} From left to right, RTX LiDAR provides ray-traced range observations for navigation and locomotion; IMU sensing provides body-state information for humanoid control; the global navigation mesh represents traversable scene geometry and supports waypoint and shortest-path queries; runtime local occupancy records the current free, occupied, and unknown regions around an embodiment for short-horizon obstacle avoidance; and frame tracking reports world-frame poses of named robot and object frames for inverse kinematics, motion planning, trajectory annotation, and spatial grounding. These channels are synchronized with the batched simulation lifecycle and are exposed to planners, controllers, tasks, and annotation systems through manager-owned runtime interfaces. } \label{fig:sensor-streams} \end{figure}

Figure~\ref{fig:sensor-streams} summarizes the range, inertial, and geometric support channels that MagicSim integrates as manager-owned runtime streams. These channels complement the visual and tactile sensing paths described earlier in this section: they provide embodiment state, global traversability, current local obstacles, and world-frame geometry for locomotion, navigation, motion planning, annotation, and language grounding. 
\FloatBarrier

\section{Assets and Annotations: From Passive Geometry to Executable Supervision}
\label{sec:annotation}
\label{sec:assets-annotations}

A simulator full of USD files is not yet a robot-learning data engine.
Raw geometry tells a robot what an object looks like, but not which part
is functional, which contact is useful, which motion is executable, or
which action was actually taken during a rollout. MagicSim therefore
uses two coupled annotation systems. Offline, the asset annotation stack,
powered by AnnotateAnything~\cite{annotateanything2026}, compiles
sim-ready assets into language, visual, and physics-validated interaction
priors. Online, the runtime annotation stack records camera evidence,
skill and planner state, selected targets, and language traces during
successful episodes. The offline compiler says what \emph{can be done};
the runtime record says what \emph{was done}.

A cabinet illustrates the lifecycle. As a raw asset, it is only geometry
and joints. As a MagicSim asset, it receives stable logical identity,
semantic labels, and a sim-ready physics package. The visual--language
stage proposes handles, drawer fronts, and reachable free space as
interaction anchors. The physics stage validates handle grasps,
drawer-pull waypoints, and interaction-ready base poses as candidate
banks. At runtime, skills query the banks, planners select feasible
candidates under randomized robot--object configurations, and annotators
record the object box, end-effector trail, ray-cast target point,
commanded orientation, camera evidence, and narration. Repeated failures
flow back into the bank. The asset becomes a living supervision
interface.

\begin{table}[htbp]
\centering
\small
\resizebox{\linewidth}{!}{%
\begin{tabular}{lllll}
\toprule
System & Sub-type & Serves & Timing & Purpose \\
\midrule
Asset annotation & Visual--language (\S\ref{sec:annotation-vl}) &
  physics annotation; segmentation; language &
  offline, per asset & grounded priors \\
Asset annotation & Physics (\S\ref{sec:annotation-bank}) &
  atomicskills; planners; navigation &
  offline; read at reset and skill time & executable priors \\
Runtime annotation & --- (\S\ref{sec:annotation-capture}--\ref{sec:annotation-language}) &
  frames; skill segments; trajectories &
  during episodes; success-time flush & learning supervision \\
\bottomrule
\end{tabular}%
}
\caption{Two annotation systems, one supervision stack. Visual--language
annotation proposes \emph{where} to engage, physics annotation validates
\emph{how} to engage, and runtime annotation records \emph{what
happened}.}
\label{tab:annotation-taxonomy}
\end{table}

The term \emph{affordance} is used in three stages. It is first
\emph{proposed} by visual--language grounding, then \emph{validated} by
physics and stored in candidate banks, and finally \emph{executed} when
a runtime annotator records the planner-induced ray-cast target point and
commanded grasp orientation for one segment. This chapter owns the schemas and MagicSim integration of those
products; AnnotateAnything owns the offline prompting, grounding,
optimization, validation, and augmentation pipelines that produce them.
Non-camera sensors such as tactile are produced by Section~\ref{sec:sensors};
when recorded, they appear as trajectory sidecars owned by the collection
layer rather than being re-explained here.

\begin{figure}[htbp]
\centering
\includegraphics[width=\linewidth]{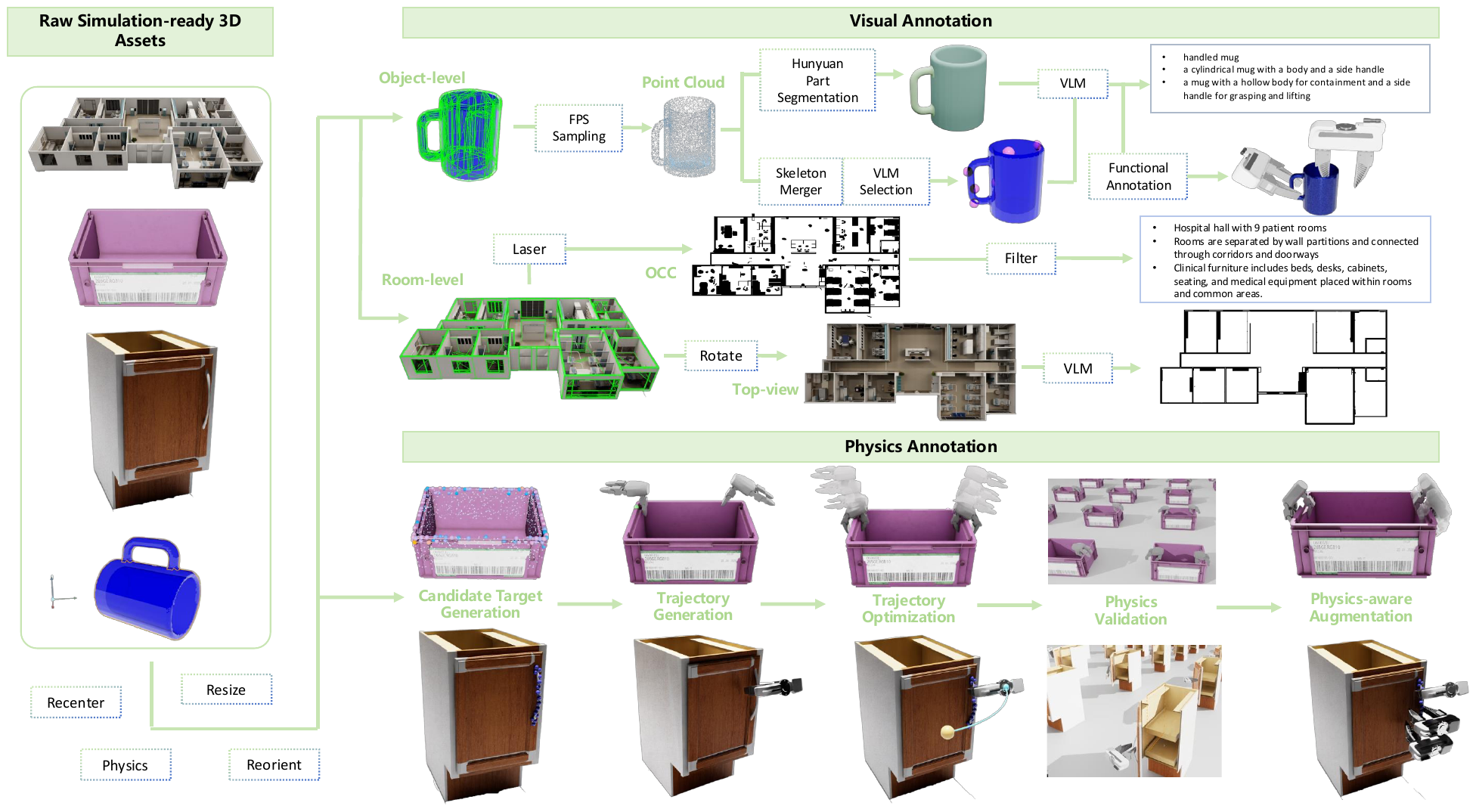}
\caption{Offline asset compilation produces grounded anchors and
physics-validated candidate banks. Runtime skills and planners consume
those banks while camera, native, action, and language annotators record
execution evidence. Execution failures prune or down-weight the bank.}
\label{fig:annotation-integration}
\end{figure}

\subsection{Sim-Ready Assets: Package, Identity, and Scene Binding}
\label{sec:annotation-asset}

Before an asset can be annotated, it must be addressable. A sim-ready
asset packages renderable geometry, collision representation, a declared
physics type, material parameters, canonical pose and scale, and the
identity and semantic hooks to which annotations attach. Normalization
recenters, rescales, reorients, and prepares physics metadata so later
stages can assume a consistent object frame and metadata surface.

The annotated pool spans 17{,}005 assets from nine source families,
grouped into four annotation families
(Table~\ref{tab:annotation-asset-pool})~\cite{annotateanything2026}. This
is an annotation-coverage taxonomy, not a solver taxonomy: the object
families of Section~\ref{sec:engine} describe simulation method, while
the families here describe which supervision labels apply.

\begin{table}[htbp]
\centering
\small
\begin{tabular}{lrrl}
\toprule
Annotation family & \#Assets & \#Ann./obj./skill & Main skill families \\
\midrule
Rigid objects          & 12{,}538 & 869 & Grasp, DexGrasp, BiGrasp, Insertion, Hanging \\
Articulated objects    &  2{,}094 & 304 & Articulation, Grasp, Navigation / Approach \\
Deformable / garments  &  2{,}167 &  83 & BiGrasp, BiDexGrasp, Deformable, Hanging \\
Room-scale scenes      &     206  & 527 & Navigation / Approach, mobile manipulation \\
\midrule
Total                  & 17{,}005 & --- & \\
\bottomrule
\end{tabular}
\caption{Annotated asset pool by annotation family. These families are
orthogonal to the simulation-method families of Section~\ref{sec:engine}.}
\label{tab:annotation-asset-pool}
\end{table}

Scene binding remains deliberately thin. A scene category points to one
USD or a folder of USDs; shared defaults compose with per-instance
overrides; selection can be random or deterministic; visual material
follows the priority color $>$ material $>$ none; and semantic labels are
written into USD semantics for runtime segmentation. Annotation follows
the \emph{logical object}, not the transient prim path, so it survives
re-instancing, hard reset, and per-environment asset diversity. Solver
semantics live in Section~\ref{sec:engine}; sampling, seeding, and reset
lifecycle live in Section~\ref{sec:runtime}; this subsection owns only
the asset-side packaging contract.

\subsection{Asset Annotation I: Hierarchical Visual--Language Annotation}
\label{sec:annotation-vl}

The visual--language stage proposes where interaction might be meaningful.
For an asset $A$, it produces grounded anchors
\begin{equation}
H(A) = P(A) \cup K(A) \cup R_{\mathrm{aff}}(A) \cup R_{\mathrm{scene}}(A),
\label{eq:anchors}
\end{equation}
where $P$ are part regions, $K$ semantic keypoints,
$R_{\mathrm{aff}}$ affordance regions, and $R_{\mathrm{scene}}$
scene-level regions. Each anchor $h$ carries a functional prior
$\varphi(h)$, such as grasp-for-use, grasp-for-opening, insertion,
hanging, articulation, or navigation approach. These are priors; they do
not become executable affordances until physics validation.

At the object level, multi-view renderings are sent to a
vision--language model~\cite{bai2025qwen3vl,qwen2026qwen35}, producing a
semantic phrase, a functional sentence, and a part-aware paragraph. A
fused RGB-D point cloud is then used for 3D grounding: semantic keypoints
are selected over farthest-point-sampling candidates~\cite{qi2017pointnetpp},
and part masks come from native 3D part decomposition with
P\textsuperscript{3}-SAM and X-Part~\cite{ma2025p3sam,yan2025xpart}.
These outputs ground handles, drawer fronts, garment corners, openings,
hooks, support edges, and other interaction-relevant regions.

Room-scale scenes receive the same hierarchy at scene granularity:
language descriptions of layout, furniture zones, and dense context;
plus multi-height occupancy maps, floor plans, and wall-structure maps
built from simulated LiDAR and ray-cast scans. Objects inside rooms are
annotated object-centrically and transformed back to the global scene
frame through shared instance identifiers, so object anchors remain
aligned with room-level reachable space.

This stage is static and asset-side. Runtime occupancy scanning belongs
to Section~\ref{sec:sensors} and the local planner of
Section~\ref{sec:control}; runtime language is deterministic and
fact-rendered in \S\ref{sec:annotation-language}; and per-frame detection
or projection labels are produced by the runtime capture stack, not by
these offline anchors.

\subsection{Asset Annotation II: Physics-Validated Action Candidate Banks}
\label{sec:annotation-bank}

The physics stage turns proposed anchors into many validated ways to act.
The unit of executable annotation is not a single label but a \emph{bank}:
\begin{equation*}
\text{visual anchor} \rightarrow \text{functional affordance}
\rightarrow \text{compatible skills} \rightarrow \text{candidate bank}.
\end{equation*}
For each anchor $h$ and compatible skill $s$, MagicSim stores
\begin{equation}
B_{h,s}=\left\{a^{(i)}_{h,s}\right\}_{i=1}^{N_{h,s}},\qquad
 a^{(i)}_{h,s}=\bigl(s,o(h),h,\varphi(h),x^{(i)},\theta^{(i)},
 \tau^{(i)},v^{(i)},d^{(i)}\bigr).
\label{eq:bank}
\end{equation}
Here $s$ is the skill, $o(h)$ the associated instance, and
$\varphi(h)$ the functional affordance. The remaining fields specify the
concrete target, skill parameters, optional trajectory, validation
metadata, and diversity descriptor (Table~\ref{tab:annotation-bank-fields}).

\begin{table}[htbp]
\centering
\small
\begin{tabular}{lll}
\toprule
Field & Meaning & Examples \\
\midrule
$x$ & concrete target & contact point, handle point, socket, base pose \\
$\theta$ & skill parameters & 6D grasp pose, hand joints, insertion / articulation axis \\
$\tau$ & optional trajectory & approach--contact--retreat, drawer pull, garment fold, nav path \\
$v$ & validation metadata & collision, IK, task success, contact stability, path feasibility \\
$d$ & diversity descriptor & approach side, contact mode, perturbation seed, trajectory family \\
\bottomrule
\end{tabular}
\caption{Core per-candidate fields of the action annotation schema in
Eq.~\eqref{eq:bank}.}
\label{tab:annotation-bank-fields}
\end{table}

Functional affordance is what separates this schema from a pile of stable
grasps. A mug handle carries grasp-for-use, while the mug body may only
carry stabilizing grasp. A cabinet handle carries grasp-for-opening, so
its grasp candidates are conditioned on the drawer or door motion they
must enable. A grasp that is physically feasible but blocks the next task
step succeeds locally and fails as long-horizon supervision.

Candidates are generated and filtered by the companion pipeline: target
localization and sampling; trajectory template generation; optimization
under task, contact, collision, kinematic, and smoothness objectives;
parallel physics validation with floating end-effectors or full robots as
needed; and physics-aware augmentation by local perturbation and
symmetry-aware expansion~\cite{annotateanything2026}. The resulting
catalog is a broad, dozens-scale set of skill-conditioned annotation
products, summarized by family in
Table~\ref{tab:annotation-skill-families}.

\begin{table}[htbp]
\centering
\small
\begin{tabular}{ll}
\toprule
Family & Representative annotation products \\
\midrule
Grasp & parallel-jaw grasp, functional grasp, lift grasp \\
DexGrasp & dexterous grasp, stable hold, dexterous handle grasp \\
BiGrasp / BiDexGrasp & bimanual grasp, dual-hand dexterous grasp, coordinated hold \\
Articulation & open, close, push / pull part, rotate joint, press \\
Insertion / Hanging & peg-in-hole, connector insertion, hang-on-hook, hang-on-edge \\
Deformable & garment pick, fold, spread, fling, stretch, retrieval-style motion \\
Navigation / Approach & navigation target, object-centric approach, interaction-ready base pose \\
\bottomrule
\end{tabular}
\caption{Representative action-annotation families. The full catalog is
not count-synchronized with the executable AtomicSkill library.}
\label{tab:annotation-skill-families}
\end{table}

Across 17{,}005 assets, AnnotateAnything produces on the order of
$10^8$ physics-validated action annotations. On the audited evaluation
suite, an attempted asset--skill pair generates 2{,}315 raw candidates on
average; 1{,}384 survive geometric constraints, 842 survive IK and
trajectory feasibility, 615 pass physics validation, and 538 are retained
in the final bank, corresponding to a 26.6\% physics-validation pass rate
and a 23.2\% retained-candidate rate~\cite{annotateanything2026}.
Physics annotation is not physics parameters: mass, friction, joint
limits, and solver settings remain runtime state owned by the sim-ready
asset and Section~\ref{sec:engine}.

\subsection{The Annotation-to-Execution Interface and the Living Bank}
\label{sec:annotation-interface}

The candidate bank is the seam between AnnotateAnything and MagicSim. It
is not just a dataset table; it is the runtime API through which skills
turn asset priors into behavior. A skill retrieves multiple candidates,
packs them into a goal-set IK or planning request, and lets the planner
select a feasible low-cost solution under the current randomized
robot--object configuration (Section~\ref{sec:skills};
\cite{curobo2023}). The bank supplies options; the planner
supplies the decision.

The prior and the record are duals. A candidate in Eq.~\eqref{eq:bank}
says what \emph{can} be done: this handle can be grasped, this drawer can
be pulled, this base pose can approach the target. The native
\texttt{affordance} annotator of \S\ref{sec:annotation-native} says what
\emph{was selected and executed}: the planner-induced ray-cast target
point on the manipulated mesh and the commanded grasp orientation for one
segment.

During collection (Section~\ref{sec:collection}), rollout outcomes are
attributed back to the candidates that produced them. Repeated failures
under randomized layouts, poses, cameras, or planner constraints are
pruned or down-weighted, making the bank a population under runtime
selection. The annotation catalog and the executable atomicskills are
related but not one-to-one: some annotation products are consumed through
existing primitives, and some executable skills require no asset
annotation. The bank is where the vocabularies meet.

\subsection{Runtime Capture Substrate and Omni Annotators}
\label{sec:annotation-capture}

Runtime annotation begins with cameras. \texttt{CameraManager} owns
camera prims and pose, including pose randomization, while
\texttt{CaptureManager} and \texttt{TiledCaptureManager} own render
products and annotators. The shared \texttt{camera:} YAML block is split
so creation, capture, and optional flying-camera planning read the same
per-camera sub-blocks. Pinhole cameras are configuration-driven;
Kinect- and RealSense-style cameras use mesh USDs and fixed mode tables;
link-mounted cameras ride the robot body. Full reset rebuilds render
products, partial reset is a no-op, and capture has no \texttt{reset\_to}
because it is a derived sink rather than reproducible simulator state.

Stock Omni Replicator annotators~\cite{nvidia2025isaacsim,mittal2023orbit}
run on this substrate: image channels (RGB, normals, depth, motion,
occlusion), segmentation, 2D/3D boxes, and geometry or metadata such as
camera parameters, point clouds, and skeleton data. Runtime semantic
segmentation is meaningful because the asset layer writes semantic labels
into USD semantics; instance segmentation becomes usable because stable
logical identities map instance ids back to object records.

\subsection{MagicSim-Native Runtime Annotators}
\label{sec:annotation-native}
\label{sec:native-annotators}

Omni annotators label what the renderer sees; MagicSim-native annotators
label what the skill meant. Three pure-CPU annotators are added by
MagicSim (Table~\ref{tab:annotation-native}). They are execution traces,
not asset metadata.

\begin{table}[htbp]
\centering
\small
\begin{tabular}{lll}
\toprule
Annotator & Granularity & Records \\
\midrule
\texttt{eef\_waypoint} & per frame & gripper image-plane trail, one or two arms \\
\texttt{obj\_bbox} & per AtomicSkill segment & frozen 8-corner projected box of the target object \\
\texttt{affordance} & per GlobalPlanner segment & ray-cast target point plus commanded grasp quaternion \\
\bottomrule
\end{tabular}
\caption{MagicSim-native runtime annotators.}
\label{tab:annotation-native}
\end{table}

The native annotators share a pinhole projection core with no Isaac
dependency. Per-environment origins lift local coordinates to world;
per-step memoization and per-segment fingerprint caches keep geometry off
the hot path; missing manager bindings produce empty payloads; and merged
JSON sidecars are flushed only for successful episodes. Projection is
see-through: no occlusion test is performed. The native
\texttt{affordance} annotator is distinct from the validated affordance prior; it records the grasp point and end-effector orientation actually realized during an executed trajectory, rather than an asset-level affordance prior.

\begin{figure}[htbp] \centering \includegraphics[ width=0.98\linewidth ]{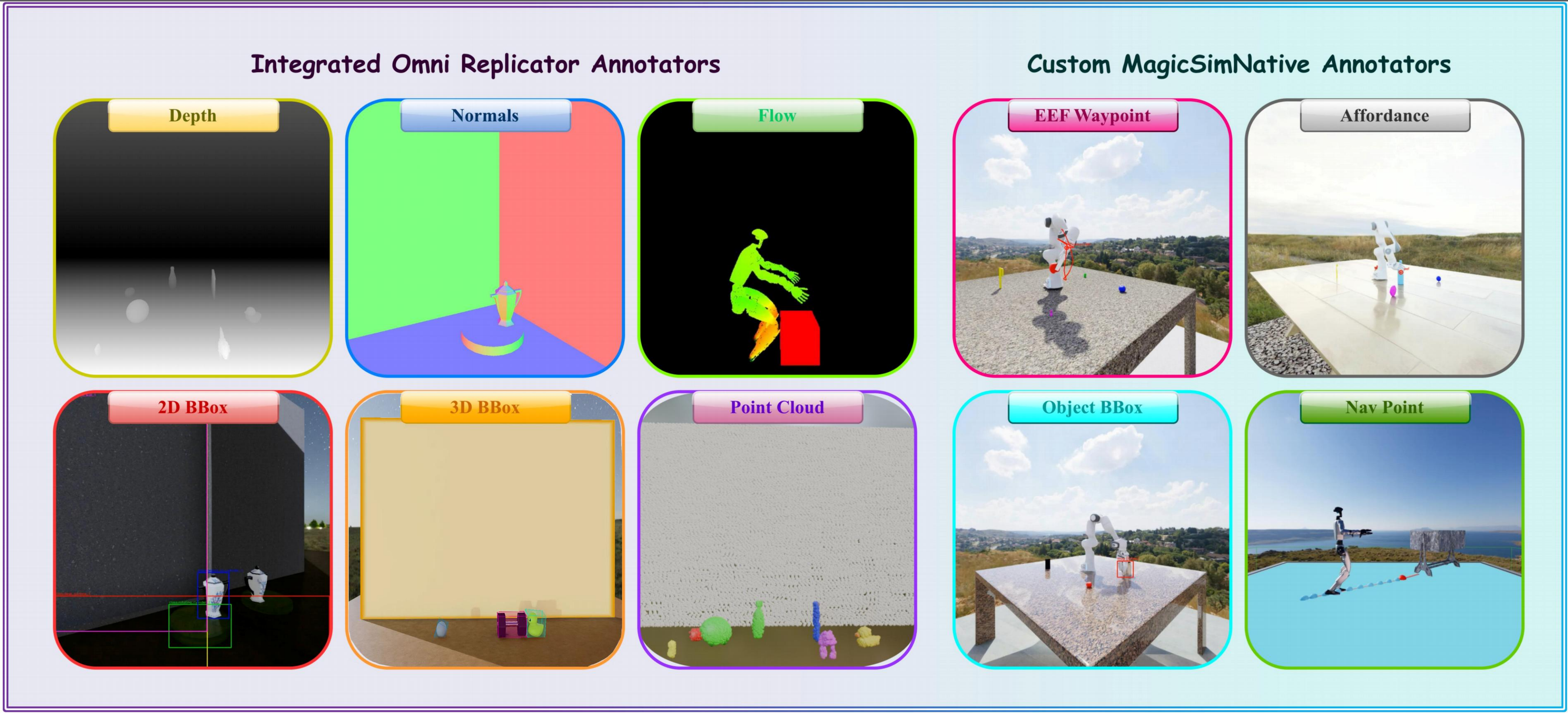} \caption[Runtime capture and annotation modalities in MagicSim.]{ \textbf{Representative runtime capture and annotation modalities in MagicSim.} The left group shows integrated Omni Replicator outputs, including depth, surface normals, optical flow, two- dimensional bounding boxes, three-dimensional bounding boxes, and point clouds. The right group shows MagicSim-native execution annotations, including end-effector waypoints, planner-selected affordance targets, target-object bounding boxes, and navigation points. The modalities are synchronized with robot actions, task states, skill phases, planner segments, camera streams, and language traces during automated collection. Native sidecars and renderer-derived outputs are retained only when the episode passes the trajectory success gate. } \label{fig:runtime-annotation-modalities} \end{figure}

Figure~\ref{fig:runtime-annotation-modalities} summarizes the representative visual and execution-grounded supervision channels available during automated data collection. The renderer-derived channels describe what is visible from a camera, whereas the MagicSim-native channels preserve targets and trajectories selected by the active skill and planning stack. 

\subsection{Runtime Language Annotation}
\label{sec:annotation-language}
\label{sec:language-annotation}

Runtime language is the semantic index of a rollout. It turns the same
execution that cameras and native annotators record geometrically into a
scene description, a task instruction, and a temporally aligned narration
that can be searched, filtered, and used as supervision. MagicSim stores
three levels: L1 describes the episode scene, including the active robot,
objects, appearance, spatial relations, lighting, effects, background,
and camera viewpoint; L2 states the task instruction after task slots have
been resolved; and L3 narrates the active AtomicSkill or GlobalPlanner
state at each step, deduplicated to change points so the saved language is
a sequence of events rather than repeated idle frames.

The language layer is fact-rendered rather than image-guessed. Managers
answer \emph{what is true}: \texttt{SceneManager} contributes object and
artifact facts such as identity, noun, color, material, pose, and bounding
box; \texttt{CameraManager} contributes viewpoint and intrinsics;
\texttt{RobotManager} contributes the active embodiment; and the command,
skill, planner, and collection managers contribute task type, target
object, hand selection, phase, and planner state. \texttt{LanguageManager}
answers \emph{how to say it}: templates, slot filling, vocabulary maps,
optional clauses, and hand-assembled relation sentences convert those
structured facts into English. L1 is built lazily once the episode has a
task, L2 when the task instruction is known, and L3 after manager updates
so skill and planner phases are current. A dense L3 mirror rides in the
\texttt{collect/} buffer; the compact L1/L2/L3 JSON files are dumped only
when the trajectory passes the success gate.

Offline asset-language annotations from \S\ref{sec:annotation-vl} enter
this runtime chain as stable object facts, not as another model call.
Semantic phrases populate object vocabulary, functional sentences expose
manipulable parts, and part-aware paragraphs provide compact interaction
summaries that scene descriptions can surface. Runtime language therefore
inherits the object knowledge produced by the asset compiler while
remaining deterministic during rollout: no VLM is queried on the hot path,
and every sentence is grounded in the same logical identities, skill
segments, and planner targets used by the geometric annotations.

\subsection{Capture and Annotation Efficiency}
\label{sec:annotation-hotpath}
\label{sec:annotation-schema} 

Efficiency is a design constraint of the annotation stack, not an
implementation afterthought. MagicSim runs many environments in lockstep;
a label that blocks every frame is more expensive than the supervision it
creates. The runtime therefore treats annotation as three cost centers---rendering,
geometry, and I/O---plus one lifecycle hazard, and pushes each one off
the critical path whenever possible.

\begin{table}[htbp]
\centering
\small
\begin{tabular}{p{0.18\linewidth}p{0.30\linewidth}p{0.42\linewidth}}
\toprule
Cost center & Failure mode & MagicSim design choice \\
\midrule
Rendering & one render product per env--camera pair & tiled \texttt{CameraView} per camera, batched over envs \\
Native geometry & repeated ray casts and projections & per-step memo plus per-segment fingerprint caches \\
Disk I/O & per-frame JSON writes & in-memory accumulation and success-time sidecar flush \\
Lifecycle & duplicated sinks after reset & idempotent rebuild; partial reset is a no-op \\
\bottomrule
\end{tabular}
\caption{Runtime annotation is engineered around the hot path. Rendering is
batched, geometry is cached, disk writes are deferred, and reset behavior
is idempotent.}
\label{tab:annotation-efficiency}
\end{table}

On the render side, tiled capture builds one batched \texttt{CameraView}
per camera over all environments, rather than one render product per
$(\mathrm{env},\mathrm{camera})$. Preallocated Warp buffers preserve the
same downstream output shape as the non-tiled path, so record code does
not need a second schema. MagicSim-native annotators deliberately bypass
\texttt{CameraView}; they continue through the projector path, so the same
YAML works with either capture backend.

On the geometry side, native annotators are segment-aware. End-effector
trails need per-frame projection, but object boxes and affordance targets
usually change only when the active skill or planner segment changes. The
projector therefore maintains a per-step memo for repeated queries within
a frame and a per-segment fingerprint cache for world points whose source
segment has not changed. Missing manager bindings produce empty payloads
instead of forcing synchronization, and see-through projection avoids
renderer-dependent visibility tests.

On the I/O side, annotation is success-gated. Native outputs, Omni
sidecars, and language records accumulate during rollout and are flushed
only when the collection layer accepts the trajectory. The collection
chapter owns the directory layout and success gate; this chapter only
specifies the annotation contribution to that layout: camera videos and
Omni/native sidecars under \texttt{camera/}, and L1/L2/L3 language under
\texttt{language/}. Failed or truncated episodes drop their buffers
instead of writing partial labels. Together, these choices make the
marginal per-frame cost of annotation dominated by pixels already being
rendered; the additional semantic, geometric, and language supervision
remains lightweight enough to run inside large-scale collection.

Together, the asset and runtime annotation systems close the supervision
loop: assets are compiled before execution, consumed by skills and
planners during execution, recorded as visual, geometric, action, and
language evidence after execution, and refined by the failures that the
runtime observes.

\FloatBarrier
%
%
%

\section{Task and Benchmark Layer}
\label{sec:tasks}
\label{sec:task-layer}
\label{sec:tasks-benchmarks}

Existing benchmarks often optimize task diversity and RL-readiness at
different layers. Some emphasize semantic and activity coverage, as in
BEHAVIOR-1K's human-centered everyday activities~\cite{li2024behavior1k};
others emphasize fast vectorized control and scalable learning interfaces,
as in ManiSkill3's GPU-parallel simulation and rendering~\cite{tao2025maniskill3}.
MagicSim's claim at the task layer is not simply that it contains more
tasks, but that task diversity is organized by a single contract: every
task, regardless of embodiment or interaction regime, instantiates the same
MDP-facing interface, configuration schema, and status ontology. A fuller
comparison with neighboring platforms is deferred to the related-work
discussion.

Section~\ref{sec:overview-one-mdp-three-drivers} stated the per-task form
of this idea: fix a task, and the same MDP can be driven by RL training,
scripted collection, or inference. This section takes the transpose: vary
the task, and the contract stays fixed. A tabletop insertion task, a
garment manipulation task, a humanoid loco-manipulation task, and an
avatar-conditioned HRI manipulation task are different MDPs, with different
observations, actions, and dynamics. What they share is the contract every
task fills. We first define this contract, then show how it composes
embodiments and interaction regimes into a diverse task surface, how the
same interface becomes RL-ready, and how the contract is frozen into a
benchmark protocol.


\subsection{One Task Contract}
\label{sec:task-contract}

A MagicSim task is not a standalone script but a parameterized MDP
instance. Each task specifies a reset distribution, a policy observation
view, an optional privileged observation view, a reward, a termination and
truncation rule, task parameters, and a status signal. The action space is
not declared by the task; it is inherited from the embodiment through
RobotManager's combined action space
(\S\ref{sec:control-robotmanager}). Table~\ref{tab:task-contract} shows
where each MDP element lives in the contract.

\begin{table}[htbp]
\centering
\small
\begin{tabularx}{\linewidth}{>{\bfseries}p{0.28\linewidth} X}
\toprule
MDP element & MagicSim realization \\
\midrule
Substrate / reset distribution &
\texttt{config.env} plus task-level randomization \\
Observation &
\texttt{get\_policy\_obs} and \texttt{get\_privilege\_obs} \\
Action space &
RobotManager's embodiment-level combined action space \\
Reward &
\texttt{get\_reward}, declared through \texttt{config.task} \\
Termination / truncation &
\texttt{get\_termination} plus batched episode buffers \\
Task parameters &
\texttt{config.task.command} \\
Evaluation signal &
\texttt{running}, \texttt{success}, \texttt{truncated},
and \texttt{failed} \\
\bottomrule
\end{tabularx}
\caption{The task contract. Different tasks instantiate different MDPs,
but expose the same contract to training, collection, and evaluation.}
\label{tab:task-contract}
\end{table}

The implementation point is \texttt{TaskBaseEnv}, which wraps the
Env-Core environment \texttt{SyncRobotEnv-V0}
(\S\ref{sec:runtime}) and reads a flat two-block configuration.
\texttt{config.env} defines the substrate consumed by the Core managers:
scene, robot, camera, animation, navigation, and simulation settings.
\texttt{config.task} defines the MDP-facing layer:
\texttt{\{observation, reward, termination[, command]\}}. Category base
environments provide defaults, while concrete tasks override five hooks:
\texttt{get\_obs\_space}, \texttt{get\_policy\_obs},
\texttt{get\_privilege\_obs}, \texttt{get\_reward}, and
\texttt{get\_termination}.

The status ontology is uniform even when success predicates are
task-specific. A task may define success by placing an object, inserting a
peg, folding a garment, reaching a camera viewpoint, or completing a
human-facing manipulation outcome with an avatar, but the episode still
reports the same status labels. In this section the label is the evaluation
signal; in the collection layer it also becomes the write gate for
trajectories (\S\ref{sec:collection}). The same contract also carries task
parameters through \texttt{config.task.command}, so training, evaluation,
and AutoCollect differ by command distributions rather than by separate
task definitions. How the substrate beneath this contract resets
deterministically, randomizes under disjoint seed streams, and snapshots
state is the subject of Section~\ref{sec:runtime}; this section treats
those properties as given.


\subsection[Diversity by Composition: Embodiment x Interaction]{Diversity by Composition: Embodiment $\times$ Interaction}
\label{sec:task-diversity}

Task diversity comes from recombining embodiments and interaction regimes
under this contract. MagicSim registers eight task families --- TableTop,
MobileManip, LocoManip, Dexterous, Garment, ContactRich, Camera, and HRI
--- spanning more than forty concrete tasks. These names organize
implementation and registration: each family is anchored by a base
environment and populated by per-task configuration files. Conceptually,
however, the benchmark is better read as an embodiment-by-interaction
matrix than as a flat list of categories.

\begin{figure}[!htb]
\centering
\includegraphics[width=0.98\linewidth]{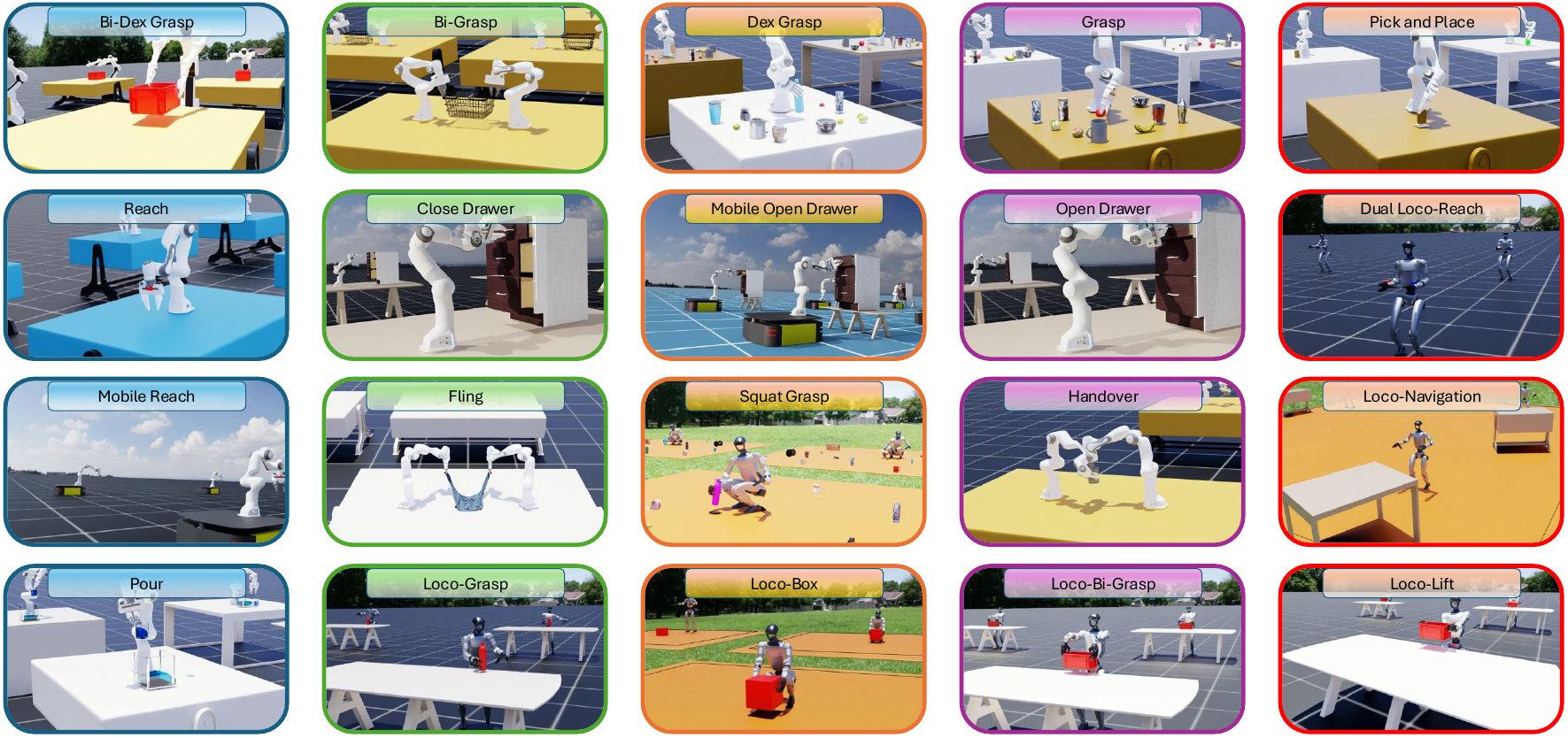}
\caption{Representative downstream tasks supported by MagicSim. The 4$\times$5 grid summarizes the task coverage exposed through the shared task and benchmark layer.}
\label{fig:downstream-tasks-grid}
\end{figure}

\begin{table}[htbp]
\centering
\small
\begin{tabularx}{\linewidth}{p{0.19\linewidth} p{0.23\linewidth} X X}
\toprule
Registered family & Embodiment / control family &
Interaction regimes & Representative task forms \\
\midrule
TableTop &
fixed-base single- and dual-arm &
rigid prehensile; articulated; contact-rich &
PickPlace, Handover, Insert \\
MobileManip &
wheeled base + arm(s) &
prehensile; articulated; navigation &
MobileGrasp, MobileOpenDrawer, Pack \\
LocoManip &
legged / humanoid + arm(s) &
loco-manipulation; articulated; navigation &
LocoGrasp, LocoOpenDoor, LocoRoomNav \\
Dexterous &
arm + dexterous hand &
dexterous prehensile; articulated &
DexGrasp, BiDexGrasp, DexOpenDrawer \\
Garment &
single- and dual-arm &
deformable manipulation &
Fling, GarmentFold, HangGarment \\
ContactRich &
fixed-base arm &
precision contact and assembly &
GearAssembly, InsertUSB, PegInHole \\
Camera &
controllable sensor carrier &
perception; viewpoint control &
GoTo \\
HRI &
robot manipulation in avatar-populated scenes &
avatar-conditioned contact-rich manipulation &
human-facing manipulation outcomes \\
\bottomrule
\end{tabularx}
\caption{Task families organize registration and substrate; interaction
regimes cut across them. HRI is a registered task family because its tasks
share an avatar-conditioned manipulation substrate; its semantics remain
manipulation-level and contact-rich.}
\label{tab:task-families}
\end{table}

The distinction between families and regimes is what keeps the taxonomy
coherent. A family fixes shared substrate: base environment, scene
assumptions, embodiment wiring, and default configuration. An interaction
regime describes the dynamics a task exercises. Thus a TableTop task may be
contact-rich, as in insertion, while ContactRich remains its own registered
family for precision assembly tasks that need a specialized substrate. The
registry is therefore not a one-dimensional task list; it is a structured
surface over controllable bodies and interaction types.

The controllable-body axis spans fixed-base arms, dual-arm systems,
dexterous hands, wheeled mobile manipulators, legged or humanoid
loco-manipulators, and camera-only sensor carriers. The interaction axis
spans rigid grasping, articulated mechanisms, non-prehensile pushing,
deformable garments, precision contact, locomotion and navigation,
viewpoint control, and avatar-conditioned contact-rich manipulation. A new
benchmark cell is a composition of these axes: the embodiment contributes
the action space, the task contributes observation, reward, termination,
and command semantics, and the category base supplies defaults. Task logic
also ports across families --- the same task form can be registered under
more than one family --- and a newly registered embodiment extends every
compatible cell without new task code.

HRI deserves its own row in the registry, but not a separate storyline in
the section. It is anchored by \texttt{HRIBaseEnv} because its substrate
contains an animated avatar and its success predicates may depend on
robot--object--avatar relations. This makes HRI adjacent to contact-rich
manipulation, but not redundant with the ContactRich family. ContactRich
tasks emphasize robot--object--fixture precision, such as insertion or
assembly against a board, socket, hole, or other rigid fixture. HRI tasks
emphasize human-facing manipulation outcomes conditioned on an avatar's
pose, motion, or interaction state. The avatar is part of the environment
state and is advanced by environment-side dynamics
(\S\ref{sec:avatar-control}); no entry of the robot action vector controls
it. HRI therefore changes the transition dynamics and success predicates of
a manipulation task without changing the agent-facing MDP contract. Compared
with avatar-centric embodied-AI settings such as Habitat~3.0
\cite{puig2023habitat}, the distinction is task level: MagicSim's HRI is
manipulation-level and contact-rich, rather than primarily navigation- or
household-rearrangement-oriented.


\subsection{RL-Ready by Construction}
\label{sec:task-rl}

The same contract makes the task layer consumable by vectorized RL
pipelines. Here, \emph{RL-ready} is an interface property rather than an
experimental result: a task is RL-ready when standard vectorized learning
pipelines can consume it without task-specific glue code. Concretely, this
means batched stepping, partial resets, stable observation dictionaries,
reward tensors, termination tensors, privileged-observation splits, and
reproducible reset streams.

These properties follow from the contract. \texttt{TaskBaseEnv} runs many
replicas inside one vectorized simulation instance and maintains
per-replica episode lengths, termination flags, and truncation flags. The
policy/privileged observation split makes asymmetric actor--critic training
a first-class interface rather than a per-task convention. Reward and
termination are declared through \texttt{config.task}, so task variants are
configuration edits rather than code forks. Reproducibility is inherited
from the runtime through seeded randomization, disjoint streams, and
\texttt{reset\_to} replay (\S\ref{sec:runtime}). Finally,
\texttt{IsaacRLEnv} subclasses IsaacLab's \texttt{DirectRLEnv}, allowing
IsaacLab-compatible learning workflows to attach to the same surface.

In the current system, low-level policy RL is exercised through an
\texttt{rsl\_rl} integration~\cite{rudin2022learning}. Closed-loop RL for
higher-level planners --- for example through post-training infrastructures
such as \texttt{verl}~\cite{sheng2024hybridflow} or
\texttt{RLinf}~\cite{yu2025rlinf} --- remains a planned use of the
agent-facing interfaces (\S\ref{sec:agent-capability}) rather than a
completed capability. The Camera family satisfies the same task contract
but is exercised through the camera-trajectory collection stack
(\S\ref{sec:collection-camera}) rather than low-level policy RL, so we make
no training claim for it here.


\subsection{Benchmark Protocol}
\label{sec:task-protocol}

The benchmark protocol freezes the contract into comparable evaluation
settings. For each task, a protocol instance fixes the reset distribution,
evaluation seed split, observation modality, action space, episode horizon,
success criterion, privileged-state allowance, command distribution, and
reported metrics.

The command block is the protocol's difficulty knob. Under RL, commands are
sampled from the task's training or evaluation distribution. Under
AutoCollect, \texttt{AsyncRobotEnv} injects
\texttt{config.collect.command} into \texttt{config.task.command}, so the
task reads the same parameters regardless of driver. Changing target-object
ranges, initial states, scene distributions, avatar states, interaction
poses, or goal tolerances therefore changes benchmark difficulty without
creating a new task implementation. Task-level randomization is declared
here as distributions over reset and command space; the runtime supplies
the reproducible sampling machinery through per-environment seeds and
disjoint streams (\S\ref{sec:runtime}), so the separation between training
and evaluation streams holds by construction rather than by convention.

The success criterion is task-specific, but its output is standardized.
\texttt{PickPlace}, for example, emits \texttt{success:} only when its
place-success termination fires, with the tolerance sized to the place pad
--- the tolerance is part of the benchmark definition, not an
implementation detail. HRI tasks follow the same rule: their predicates may
involve robot--object--avatar relations or contact outcomes, but they still
emit the same standardized status labels. Privileged observations may be
used by critics or teachers during training, but benchmark reports must
separate policy inputs from privileged channels, matching the asymmetric
training interface in \S\ref{sec:task-rl}.

Metrics derive from the same status ontology used during execution:
success rate, truncation rate, and failure rate over the evaluation
distribution, optionally with normalized episode length among successful
episodes. The status string is therefore not log text; it is the common
source of evaluation metrics in this section and the trajectory write gate
in the collection layer (\S\ref{sec:collection}). This section defines the
protocol and evaluation interface only; experiments and per-task scores are
reported separately.
\FloatBarrier
%

\section{MagicSim AtomicSkill System}
\label{sec:atomicskill}
\label{sec:skills}
\label{sec:atomicskills}

The task layer (\S\ref{sec:task-layer}) defines what success means; the
AtomicSkill layer defines the verified actions through which that success is
executed, recorded, and exposed to higher-level decision makers. AtomicSkill
is MagicSim's physics-facing action layer. AutoCollect uses it to turn
sampled command sequences into verified trajectories and aligned supervision;
hierarchical policy evaluation uses the same vocabulary as an action
interface. Individual skills may be realized by motion generation, frozen
policies, model-predictive control, or scripts, but they expose the same
lifecycle states, typed outcomes, success checks, and recording boundaries.

The design choice is simple: fix the layer that touches physics. Above the
AtomicSkill interface, task intent may be scripted, sampled, or selected by a
policy. Below it, actions are grounded in live scene state, executed by
pre-validated backends, recovered through typed failure paths, and checked by
success gates. This makes task intent executable, batchable where planning is
used, recoverable, and selectable by high-level policies.

\subsection{Why Atomic Skills: Fixing the Layer That Touches Physics}
\label{sec:skill-design-space}

Scalable robot-data generation must answer three questions: where expert
behavior comes from, where generativity is allowed, and what keeps execution
success high enough for collection. Teleoperation answers with human
operators; it gives faithful demonstrations but scales poorly with data
volume and embodiment complexity, especially for bimanual, dexterous, mobile,
and whole-body systems~\cite{zhao2023aloha}. Demonstration-transform methods
answer with source demonstrations: MimicGen, DexMimicGen, SkillMimicGen, and
SoftMimicGen amplify recorded behaviors through object-, skill-, or
deformable-object transforms
~\cite{mandlekar2023mimicgen,jiang2024dexmimicgen,garrett2024skillmimicgen,moghani2026softmimicgen}.
They inherit human strategies, but their generalization is conditioned on the
support of the source demonstrations.

LLM-based systems answer with generated execution logic. RoboTwin~2.0,
GenSim, GenSim2, and RoboGen generate task code or per-task solvers
~\cite{chen2025robotwin2,wang2023gensim,hua2024gensim2,wang2023robogen};
HumanoidGen defines atomic operations but uses LLM-generated constraint
chains to sequence them~\cite{jing2025humanoidgen}. These systems expand task
coverage, but the generated artifact sits inside the execution boundary: code
or constraints must be semantically correct before touching physics. Thus
program generation, repair, and execution success become entangled.

MagicSim takes the complementary position. Generativity is allowed above the
interface, but the executable layer is fixed. atomicskills encode the
physical decisions that determine success---grasp goal construction, descent
gates, stable-hold checks, contact predicates, base-placement constraints, or
learned-policy success gates---once, then reuse them across tasks and scenes.
The trade is explicit coverage: a fixed vocabulary cannot invent arbitrary new
strategies by prompting alone. In return, execution success is engineered in a
verified physics-facing layer rather than re-established for every generated
program.

\subsection{One Vocabulary, Two Interfaces: Actions Upward, Execution Downward}
\label{sec:skill-two-faces}

The high-level interface is not a state machine. It is an action vocabulary
over the shared MDP state. At the command level, an action specifies which
command to run, on which object, with which robot, hand, mode, and target
parameters. At the skill level, an action is a single AtomicSkill invocation.
A scripted command source, hierarchical policy, or VLM/LLM high level planner
selects one of these actions.

The skill executor owns those lifecycle states. It expands commands into
skills, steps each skill, issues backend requests, handles retries and
timeouts, propagates typed statuses, and emits terminal states consumed by
the success gate. Thus an AtomicSkill has two faces: upward it is an action
symbol; downward it is a traceable execution unit with phase state, target
state, typed outcomes, and recording boundaries.

\begin{table}[htbp]
\centering
\small
\begin{tabularx}{\linewidth}{>{\bfseries}p{0.25\linewidth} X p{0.18\linewidth}}
\toprule
Component & Role & Lifecycle FSM? \\
\midrule
Command source / high-level policy &
Selects command- or skill-level actions over the shared MDP state. &
No \\
Skill executor &
Expands commands into skills; drives backend execution; handles retry,
timeout, truncation, status propagation, and logging. &
Yes \\
AtomicSkill &
Maps selected action intent to backend requests and typed execution status. &
Action upward; trace downward \\
\bottomrule
\end{tabularx}
\caption{One vocabulary with two interfaces. Lifecycle state belongs to the
skill executor, not to the policy-facing action API.}
\label{tab:skill-two-faces}
\end{table}

\subsection{AtomicSkill as an Online Goal Compiler}
\label{sec:skill-library}

Most atomicskills are planner-grounded. They store no demonstration or
reference trajectory; they compile task intent and live scene state into
planner goals, success predicates, and recovery attempts. Learned or
MPC-backed skills (\S\ref{sec:skill-backends}) share the same external
interface but replace this compiler with a fixed policy, state-base RL policy or controller. The
planner-grounded case makes the contrast to demonstration methods
clearest: a transform adapts recorded behavior, while an AtomicSkill computes
current-scene goals directly.

The planner-grounded library separates intent from realization. Skills own
target construction, phase gates, contact checks, object-state predicates,
and terminal status strings. Motion realization is restricted to a small set
of GlobalPlanner primitives, including \texttt{MoveL}, \texttt{ServoL},
\texttt{MobileMoveL}, \texttt{MobileServoL}, \texttt{RetractMoveL},
\texttt{NavTo}, \texttt{ParallelGripper}, and \texttt{DexHand}. Roughly
twenty-two skills over eight primitives realize on the order of thirty
command types across the task families and embodiments of
\S\ref{sec:task-layer}.
The claim is not that any skill is universal, but that covered interaction
patterns are validated at scene time and reused across poses, layouts,
instances, and embodiments.

\begin{figure}[!htb]
\centering
\includegraphics[width=0.98\linewidth]{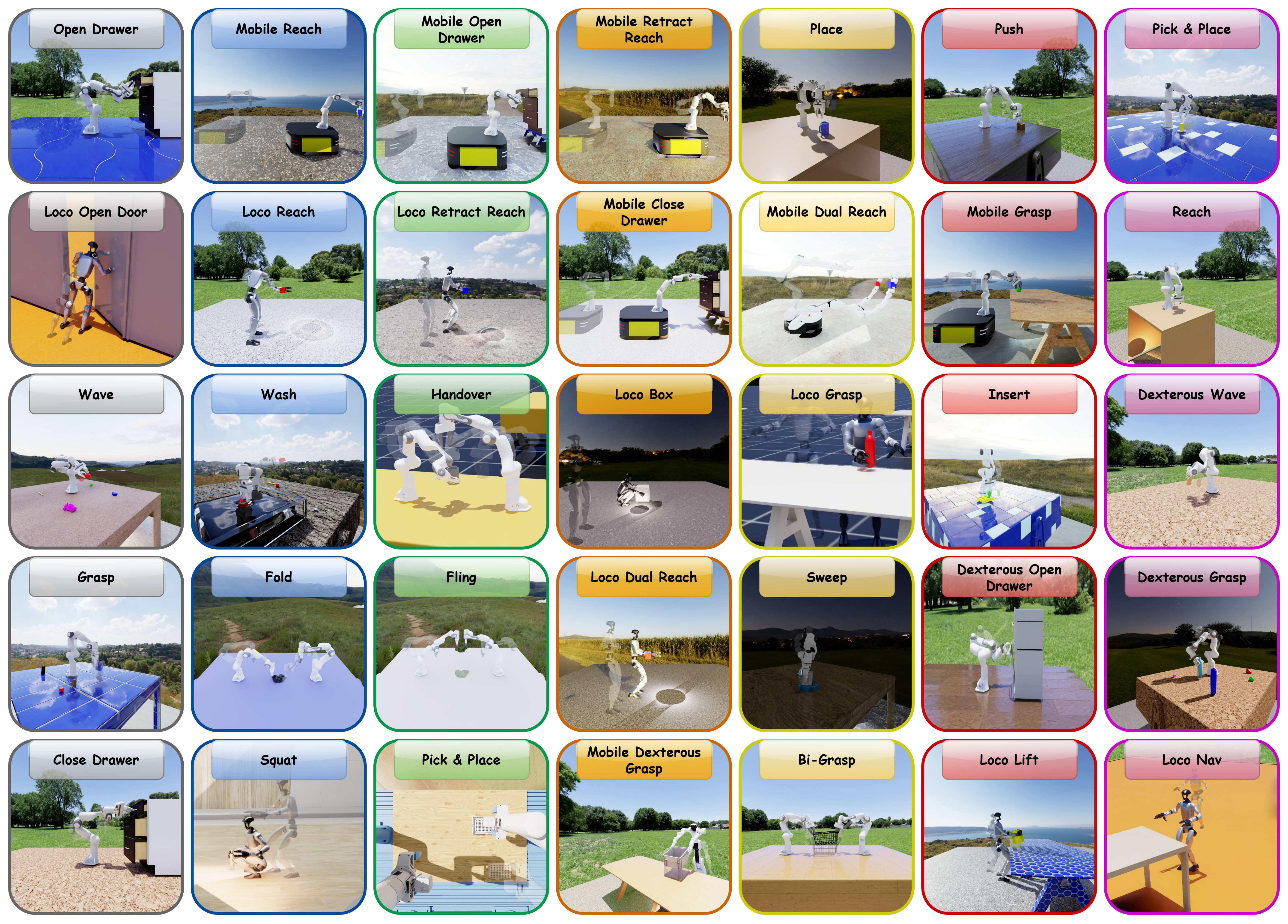}
\caption[Representative AtomicSkill-enabled behaviors.]{
\textbf{Representative AtomicSkill-enabled behaviors across embodiments.}
The repertoire spans reaching, grasping, placing, pushing, inserting,
articulated-object interaction, cloth manipulation, sweeping, washing,
handover, navigation, and bimanual manipulation across tabletop arms, mobile
manipulators, humanoids, and dexterous hands. These behaviors share the same
verified skill interface even when their realization backends differ.}
\label{fig:atomic-skill-summary}
\end{figure}

\subsection{Skill Backends and the Batch Boundary}
\label{sec:skill-backends}
\label{sec:skill-planner-boundary}

Fixing the AtomicSkill interface does not fix how every skill is realized.
The backend is chosen for the physics. Geometric manipulation uses motion
generation because it is fast, reliable, and easy to check: target poses,
collision constraints, active hands, and goalsets are constructed from the
live scene, then verified before and after execution. These planner-grounded
skills are also the throughput path. They emit normalized requests of the
form \texttt{((robot\_id, hand\_id, planner\_mode), target)}, which are
grouped by solver type, robot configuration, and planner mode and solved
asynchronously by the planning farm (\S\ref{sec:async-planning}). This is the
only backend for which we claim cross-environment batched motion solves.

Other regimes need different backends. High-DOF dexterous behaviors, such as
in-hand reorientation, are not realistically hand-coded or solved as simple
geometric plans, so MagicSim trains a series of state-based RL policies like
~\cite{chen2021inhand}. These policies add behaviors and micro-strategy
diversity that scripted trajectories cannot express, while still reporting
through the same skill lifecycle and success gate. Contact-rich continuous
skills, such as nonprehensile pushing, instead need tight closed-loop
correction; an MPC backend supplies small feedback actions as contact evolves
~\cite{bui2025pushanything}. MPC runs per environment rather than through the
motion-generation solve farm, because its value is feedback, not batching.
Simple deterministic behaviors may remain scripts.

The unification is the contribution. Motion generation, frozen policies, MPC,
and scripts are fixed, pre-validated artifacts at collection time. Each
receives a skill invocation, advances under executor-owned lifecycle control,
reports typed outcomes, is checked by the success gate, and writes the same
command--skill--backend hierarchy. A single episode can chain a planned
grasp, learned in-hand rotation, and MPC push in one persistent world and one
trajectory format. The backend changes how a selected skill is realized; it
does not change how the skill is selected, verified, segmented, or recorded.
RL-as-backend is therefore distinct from RL-as-driver: the former realizes one
skill below the interface, while the latter selects commands or skills above
it (\S\ref{sec:skill-action-interface}).

\subsection{Typed Failure, Multi-Command Episodes, and Data Boundaries}
\label{sec:skill-failure}
\label{sec:skill-long-horizon}

AtomicSkill does not remove failure; it makes failure local, typed, and
useful.  With a fixed skill vocabulary, semantic correctness is paid
once per skill or backend. Collection-time failures are mostly geometric or
physical: no IK, collision, unstable grasp, missed contact, timeout,
controller failure, or adverse object state. These failures trigger a fixed
ladder---planner retry, alternative goalset, grasp/contact resampling, skill
retry. \textbf{This provides precious failure-recovery data which is harder to collect in the real world}.

An episode is a persistent-world sequence of commands over the same
AtomicSkill vocabulary. Later commands consume the physical consequences of
earlier commands; long-horizon behavior is obtained by sequencing commands,
not by adding a separate long-horizon controller. Storage is segmented for
reliability: each successful command can flush its own trajectory record,
while episode, command, skill, and backend identifiers recover the hierarchy.
This preserves verified segments even if a later command fails.

The same hierarchy is also supervision. Command boundaries are subtask
boundaries; skill boundaries are execution boundaries; backend or planner
segments are motion-affordance boundaries. Native annotators attach
end-effector waypoints, target-object boxes, affordances, and narration at the
same boundaries (\S\ref{sec:native-annotators},
\S\ref{sec:language-annotation}). Thus long-horizon data is composable at
collection time and decomposable at training time. The G1 example in
\S\ref{sec:introduction} follows exactly this format: locomotion and
manipulation commands share one scene, while each verified command is stored
with recoverable command, skill, and backend indices.

\subsection{AtomicSkill as a High-Level Control Interface}
\label{sec:skill-action-interface}

The vocabulary that drives collection is also a high-level action space.
Replacing the scripted command source with a learned high level LLM policy changes who
selects the command or skill; it does not change grounding, control,
recovery, verification, recording, or annotation. A command-level policy can
select task-scale actions such as pick, place, open, navigate, or execute a
multi-command program; a skill-level policy can select reusable options for
hierarchical imitation, option learning, or reinforcement learning.

\FloatBarrier
\section{Asynchronous Data Collection and Serving}
\label{sec:collection}
\label{sec:collection-camera}
\label{sec:camera-capture}

The previous sections define the task contract, AtomicSkill executor, planner
services, annotations, and replay. This section only describes the
collection-side scheduler and the serving boundary. The invariant is simple:
physics is batched, but collection state is per environment. Every sub-environment
crosses the same \texttt{sim\_step} barrier, while commands, skill phases,
planner futures, recovery attempts, language updates, episode boundaries, and
writes advance independently. The collector therefore has one shared clock, but
not one shared timeline.

\subsection{The Collector Loop}
\label{sec:collection-loop}

AutoCollect turns a task MDP into a scripted demonstration generator by adding
three layers above the task environment: command sampling, skill/planner
execution, and recording. Each tick follows the same descent-and-ascent pattern:
\[
\begin{aligned}
&\texttt{AutoCollectManager.step}
\rightarrow
\texttt{AtomicSkillManager.step}
\rightarrow
\texttt{GlobalPlannerManager.step}
\\
&\rightarrow
\texttt{env.step(robot\_action)}
\rightarrow
\texttt{GlobalPlannerManager.update}
\\
&\rightarrow
\texttt{AtomicSkillManager.update}
\rightarrow
\texttt{AutoCollectManager.update}.
\end{aligned}
\]
The downward pass produces the robot action for the current tick; the upward pass
reads the post-physics state and advances the per-environment state machines.

The collector does not reinterpret skill semantics. AtomicSkill owns typed
failure, recovery, retries, and phase transitions; the task layer owns the final
success predicate. AutoCollect only maintains slots: it samples each
environment's task from the weighted \texttt{task\_string} distribution, tracks
which environments are still running, failed, or pending reset, clears terminal
slots, and lets the next tick resample them. Because one environment can reset
while its neighbors keep running, skills validate substeps against live scene
state rather than cached poses. Language and recording run after the manager
updates, so the per-step narration and collect stream describe the state that
actually held after the tick.

\subsection{Free-Running Episodes on a Lockstep Tick}
\label{sec:collection-async}

Desynchronization of giving high-level commands comes from three sources: motion horizons differ across
targets and layouts; planner futures return at different times; and episodes
contain different command sequences, retries, failures, and resampled tasks. The
solve farm itself is owned by \S\ref{sec:curobo-async}; the collection loop only
uses its caller-side contract: submit a request, keep stepping the batch, and
harvest the future when it is ready. Thus an environment waiting on a planner can
hold pose or continue a previous segment without preventing other environments
from finishing skills, entering recovery, or writing successful trajectories.

The efficiency argument is structural. A barriered collector pays the maximum
phase length in the batch, $\max_i h_i$, at every phase boundary. A free-running
collector pays each environment's own horizon and synchronizes only at the
physics tick. In one sentence, one long phase should not become a batch-wide stall.

\subsection{Serving and Driver Attachment}
\label{sec:collection-serving}
\label{sec:serving}
\label{sec:collection-three-drivers}

Serving applies the same scheduling rule at process scale. A single simulator
process still has one safe stepping thread, so \texttt{GymService} marshals
concurrent HTTP requests onto that thread through request/response queues.
Across processes, \texttt{AsyncClient} fans calls out to multiple upstream
servers, each with its own \texttt{(port, env\_num)}, and exposes the pool as one
larger blocking batched environment. The outside sees concurrency; each simulator
inside the pool still advances on one clock.

This makes serving a transport layer rather than a new environment definition.
The same \texttt{reset}, \texttt{reset\_to}, \texttt{step}, observation, and
\texttt{info} contract can host different drivers. An RL runner drives
\texttt{TaskBaseEnv} directly through the RobotManager action space. AutoCollect
adds the scripted command--skill--planner scheduler and recorder above that MDP.
An inference or replay runner attaches as an external driver whose actions still
enter through the RobotManager action contract, not through the collection
layer's skills or planners. A \texttt{verl}-facing wrapper attaches at this same
boundary; closed-loop high-level post-training through \texttt{verl} or
\texttt{RLinf} remains an integration point rather than a completed result. Full
step \texttt{info}, including the dense language mirror, can be forwarded to
served clients when the deployment accepts the bandwidth cost.
\FloatBarrier

\section{Supported Capabilities}
\label{sec:capabilities}
\label{sec:supported-capabilities}

MagicSim's supported capabilities are three ways of reading the same
runtime rather than three separate subsystems. The same episode can be
used as a benchmark rollout, a success-gated trajectory, or an
agent-facing interaction loop. Section~\ref{sec:overview-one-mdp-three-drivers}
stated this as the one-MDP--three-drivers claim; the preceding sections
then defined the mechanisms. This section only assembles those mechanisms
from the user's point of view: what product the user gets, which entry
point they run, and where the current status boundary lies. The recurring
signal is the task status ontology: in benchmarking it is the metric
source, in collection it is the write gate, and in agent interaction it
becomes structured feedback.

\begin{table}[htbp]
\centering
\small
\begin{tabularx}{\linewidth}{p{0.18\linewidth} X X X}
\toprule
View & User-facing product & Main entry point & Status boundary \\
\midrule
Benchmark &
per-task evaluation records over the task registry, including robot,
avatar-conditioned HRI, and camera-only / flying-camera tasks &
\texttt{env\_string} + task configuration + benchmark protocol over
\texttt{TaskBaseEnv} &
Current for the gym-compatible MDP surface and low-level RL interface;
camera tasks are trajectory / active-perception benchmarks rather than
manipulation-policy training evidence \\
Data collection &
successful multimodal trajectories with actions, collect traces, camera
streams, native annotations, and language &
\texttt{task\_string} weights + scene/task distributions +
\texttt{AutoCollectEnv} / camera-collection mirror stack &
Current for robot and camera collection; served at scale through the
environment-pool path \\
Agent / VLM interaction &
action--observation loops with visual, language, scene, status, and
scenario feedback &
\texttt{TaskBaseEnv} locally or GymService / AsyncClient remotely, with
additional grounded and world-authoring surfaces &
Current for MDP control, serving, camera/avatar scenario surfaces, and
in-process feedback; high-level command injection, VLM layout scope, and
closed-loop planner RL are bounded below \\
\bottomrule
\end{tabularx}
\caption{Supported capability views. Each row is a product view over the
same runtime. Mechanism details remain in the task, collection, serving,
control, annotation, and layout sections referenced below.}
\label{tab:capability-summary}
\end{table}

The long-horizon G1 episode from the introduction is the simplest way to
see the reuse. As a benchmark, it is a rollout with task status and
per-task metrics. As data, it is a
\texttt{\{trajectory\_id\}/} directory with synchronized action, collect,
environment, camera, annotation, and language streams. As an agent
interaction, it is the same world exposed as an observation-and-feedback
loop through the gym or served interface. Nothing about the episode has
to change; only the driver and the consumer change.


\subsection{Benchmark Capability}
\label{sec:capability-benchmark}

MagicSim supports benchmark use by exposing the task registry as
gym-compatible MDPs with shared observation, action, reward,
termination, success, and status semantics. The main point in this
section is not another list of task mechanics, because the task contract,
benchmark protocol, and task-family table are owned by
\S\ref{sec:task-contract}--\S\ref{sec:task-protocol}. The capability is
breadth under one contract: the registry spans fixed-base manipulation,
mobile manipulation, humanoid and legged loco-manipulation, dexterous
hands, garments, contact-rich assembly, avatar-conditioned HRI, and the
Camera family, where the controllable body is a camera / flying-camera
carrier rather than a robot. A benchmark user therefore chooses an
\texttt{env\_string}, task configuration, protocol, and seed split, then
reports per-task success, truncation, failure, and episode-length
metrics. Low-level policy RL is exercised through the task-layer RL
interface (\S\ref{sec:task-rl}); camera-family tasks share the same MDP
contract but should be described as camera-trajectory or active-perception
benchmarks, not as evidence for manipulation-policy training.


\subsection{Data Collection Capability}
\label{sec:capability-collection}

MagicSim supports data collection by turning task specifications and
scene distributions into successful, annotated embodied trajectories. A
user provides \texttt{task\_string} weights, scene and task distributions,
camera / annotation settings, and an \texttt{output\_dir};
\texttt{AutoCollectEnv} drives the task through the Command--Skill--Planner
hierarchy, the asynchronous planning services keep the batched loop from
blocking, and the record layer writes only episodes whose task state
passes the \texttt{success} gate (\S\ref{sec:skills},
\S\ref{sec:collection}, \S\ref{sec:collection-async}). The saved product
is a trajectory directory with robot or camera actions, collect state,
observations, camera streams, native annotators, and L1/L2/L3 language
(\S\ref{sec:annotation-language}--\S\ref{sec:annotation-schema}). The
camera-trajectory mirror stack reuses the same Record, Capture, and
Language path (\S\ref{sec:collection-camera}), so collection is not tied
to manipulation: the same capability also produces flying-camera / active
perception demonstrations. This is the most mature of the three
capabilities, including the success gate, trajectory layout, annotation
flush, language dump, and served environment-pool path
(\S\ref{sec:collection-serving}).


\subsection{Agent / VLM Interaction Capability}
\label{sec:agent-capability}

MagicSim's agent-facing capability exposes the same runtime as an
interactive substrate. The current loop has three exercised pieces. First,
\texttt{TaskBaseEnv} provides the gym-level MDP boundary: observations and
status return from the task, while actions enter through RobotManager's
combined action space (\S\ref{sec:control-robotmanager}). Second,
GymService and AsyncClient expose the same reset, \texttt{reset\_to},
step, observation, and feedback contract through the serving layer
(\S\ref{sec:collection-serving}). Third, the observation side includes
camera streams, native runtime annotations, three-level language, scene
facts, and task status; in process these fields are available through the
task and collection \texttt{info}, and in served deployments the full-info
payload can be forwarded when the client needs structured feedback rather
than a minimal observation--reward channel.

Around that loop, MagicSim exposes grounded environment surfaces rather
than magic actions. The Command--Skill--Planner hierarchy is implemented
and exercised inside the collection stack, so high-level commands are
known to lower into skill state machines, planner requests, and ultimately
robot or camera actions stepped by the simulator (\S\ref{sec:collection}).
Exposing that hierarchy as an external served-agent command API remains
an integration point. Scenario control is already current on the
environment side: avatars can be commanded and read back as animated
world actors, while camera poses and flying-camera planners provide an
active-perception entry point without changing the robot MDP action
boundary (\S\ref{sec:avatar-control}, \S\ref{sec:flying-camera}). The
layout layer supplies the world-authoring side of the interface through
its VLM layout contract; its exact scope and implementation status should
remain aligned with the layout and limitations sections
(\S\ref{sec:layout-manager}, \S\ref{sec:limitations}).

The remaining agent drivers are deliberately not claimed as completed
results. InferenceRunner is a planned policy / replay driver over the same
\texttt{TaskBaseEnv}, with actions flowing through RobotManager rather
than through the collection skill stack. External command injection is the
upgrade path from the current internal hierarchy to an agent-facing
command surface. Closed-loop reinforcement learning of high-level planners
is likewise a planned use of the same interfaces, not an experimental
claim in this section (\S\ref{sec:task-rl}). The downstream research
questions enabled by these surfaces are deferred to
\S\ref{sec:domains-vlm}; this section only fixes the interfaces and their
status.
\FloatBarrier
%

\section{Downstream Research Domains}
\label{sec:domains}
\label{sec:downstream}

MagicSim downstream use is organized around one executable substrate. Robotics
reads an episode as an MDP rollout and a success-gated demonstration; physics
reads it as a replayable physical interaction under controlled variation;
embodied vision--language research reads it as a privileged trace that can also
be evaluated interactively. The sections below therefore expose three readings of
one system rather than three separate simulators.


\subsection{Robotics}
\label{sec:domains-robotics}

Robotics is the native reading of MagicSim: the task layer exposes a
gym-compatible MDP, and the collection layer drives the same tasks through the
Command--Skill--Planner hierarchy (\S\ref{sec:task-rl}; \S\ref{sec:collection}).
RL uses the task contract directly; imitation and robot foundation models use
success-gated trajectories; domain randomization shapes both for transfer
(\S\ref{sec:dr-seed-streams}). The shared surface is the task-family by
embodiment matrix of \S\ref{sec:task-diversity}.


\subsubsection{Policy Learning Paradigms}
\label{sec:domains-policy}

A saved trajectory can be read as behavior, dynamics data, or
representation-level supervision. The difference is what the learner consumes,
not which simulator generated the episode.

\paragraph{End-to-end policies.}
Vision--language--action and world--action models need aligned observation,
language, and action streams at scale~\cite{Intelligence2026pi07AS,generalist2025gen0,Nvidia2025GR00TNA,Ye2026WorldAM}.
MagicSim trajectories pair camera streams, robot actions, language narration, and
skill/planner state, giving alignment at decision granularity
(\S\ref{sec:collection}). Parallel, success-gated generation then supports
scaling, diversity, and quality-ablation studies.

\paragraph{World and dynamics models.}
World-model research reads the same record as an action-conditioned rollout with
pixels, actions, and privileged state~\cite{Ye2025LearningTF,yuan2026fastwamworldactionmodels}.
The recorded action supplies inverse-dynamics supervision. Deterministic replay
also gives a same-initial-state, same-action-sequence protocol: restore a
\texttt{reset\_to} snapshot, execute the recorded actions, and compare model and
simulator rollouts step by step (\S\ref{sec:runtime}; \S\ref{sec:task-contract}).

\paragraph{Representation-centric policies.}
Representation-centric methods use point clouds, affordances, waypoints, boxes,
or object trajectories as control interfaces. MagicSim exports these through
depth, normals, segmentation, waypoint, object-box, and affordance annotators;
affordance is grounded in the planner-selected grasp target realized in a
successful execution (\S\ref{sec:annotation}). Each demonstration therefore pairs
2D observation, 3D or representation-level supervision, and robot action\cite{ning2023where2explorefewshotaffordancelearning, wu2022vatmartlearningvisualaction, shen2025biassemblelearningcollaborativeaffordance, chen2026pa3ff,Li2024BroadcastingSR}.


\subsubsection{System-Level Competencies}
\label{sec:domains-competencies}

We use \emph{competencies} for episode-level behavior: planning over time,
recovering from failure, and transferring across embodiments.

\paragraph{Long-horizon and hierarchical planning.}
Task and motion planning couples discrete task structure to continuous
motion~\cite{garrett2021tamp}. MagicSim exposes this coupling directly: commands
decompose into atomicskills, skills into planner primitives, and primitives into
per-tick robot actions (\S\ref{sec:skills}). Asynchronous planner services keep
batched environments stepping while IK and motion-generation solves are in flight
(\S\ref{sec:control-mid}; \S\ref{sec:curobo-async}).

\paragraph{Failure recovery and replanning.}
Failures are labeled at the tier where they occur and propagate through the same
status ontology used for metrics (\S\ref{sec:task-protocol}). Snapshot replay
restores failing initial conditions; successful trajectories can contain
retry-to-success segments; and, with the optional failed-episode persistence flag,
terminal failures can be saved as replayable corpora instead of cleared by the
default success gate (\S\ref{sec:runtime}; \S\ref{sec:collection}).

\paragraph{Cross-embodiment generalization.}
Roughly thirty-three embodiments across seven control categories share one task
contract: tasks define observations, rewards, and termination, while robots
define action spaces (\S\ref{sec:control-robotmanager}). Adding a new embodiment
extends every compatible task cell without rewriting task code, making held-out
embodiment evaluation a configuration-level experiment
(\S\ref{sec:task-diversity}).


\subsubsection{Application Domains}
\label{sec:domains-applications}

Home robotics is the exercised setting; safety research reads episodes through
contact and failure; autonomous laboratories are the capstone integration.

\paragraph{Home assistant robotics.}
MagicSim combines indoor rooms, articulated furniture, clutter, mobile and
humanoid embodiments, manipulation tasks, and language-conditioned commands
(\S\ref{sec:task-diversity}). Human-facing cases use environment-side avatars
whose dynamics are part of the transition, not the robot action space
(\S\ref{sec:avatar-control}). This places handover and human-aware placement near
Habitat~3.0 and BEHAVIOR-1K~\cite{puig2023habitat,li2024behavior1k}, while
remaining contact-rich and manipulation-centered.

\paragraph{Robot safety research.}
MagicSim is not a certification framework; it is a replayable substrate for
studying safety-relevant behavior. Collision-aware planning, tactile/contact
signals, avatar proximity, and snapshot-level failure replay map onto HRC safety
axes such as sensing, planning, replanning, and contact
behavior~\cite{li2024hrcsafety} (\S\ref{sec:control-mid}; \S\ref{sec:sensors};
\S\ref{sec:avatar-control}; \S\ref{sec:runtime}). These scenes also support
semantic-safety evaluation for VLM-controlled robots~\cite{sermanet2025asimov}.

\begin{figure}[htbp]
\centering
\includegraphics[width=0.98\linewidth]{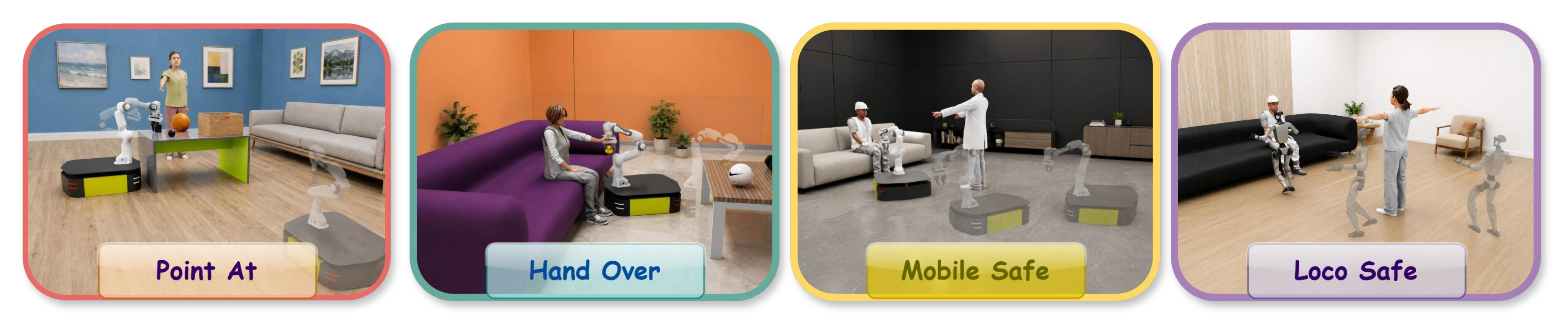}
\caption[Human--robot interaction and safety scenarios in MagicSim.]{
\textbf{Representative HRI and safety scenarios.} Examples include pointing,
handover, mobile operation near people, and humanoid locomotion in populated
environments. Avatars are environment-side actors; the robot is the controlled
embodiment.}
\label{fig:hri-safety}
\end{figure}

\paragraph{Autonomous laboratories.}
Autonomous laboratories are the strongest integrative robotics domain for
MagicSim. A self-driving laboratory needs a robot-facing world where containers,
liquids, granular materials, tools, fixtures, and articulated devices interact
under an executable protocol~\cite{tom2024sdl}. MagicSim addresses this
execution-and-evaluation layer: scenes stage liquid transfer, granular handling,
precision insertion, stirring, shaking, pouring, and equipment operation; the
protocol is an executable skill sequence with verification, retry, and failure
labels; and high-level commands are grounded in skills rather than state edits
(\S\ref{sec:engine}; \S\ref{sec:skills}; \S\ref{sec:collection};
\S\ref{sec:agent-capability}).\cite{pan2025advevo_marl, pan2025fairreason}

\begin{figure}[htbp]
\centering
\includegraphics[width=0.98\linewidth]{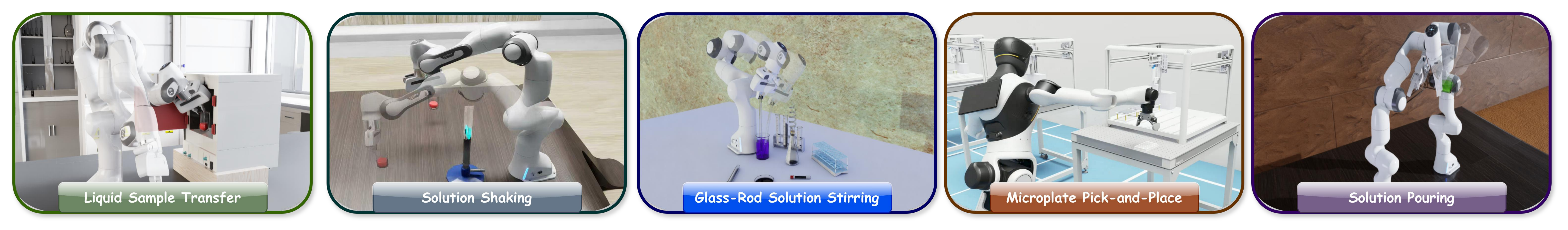}
\caption[Representative autonomous-laboratory operations in MagicSim.]{
\textbf{Representative autonomous-laboratory operations.} Liquid transfer,
shaking, stirring, pipette manipulation, and pouring run through the same
Command--Skill--Planner--Robot stack and are recorded with actions,
observations, annotations, language, and tier-level status.}
\label{fig:autonomous-laboratory}
\end{figure}

The claim is narrower than a full autonomous scientist but stronger than a visual
demo: MagicSim supplies the robotic protocol-execution substrate such loops
require. A protocol is run, recorded, replayed, and ablated; deterministic replay
restores the same recorded state while perturbing one factor such as pose, mass,
friction, appearance, timing, or object choice (\S\ref{sec:runtime};
\S\ref{sec:dr-seed-streams}). Auto-lab is therefore the capstone application: it
combines long-horizon manipulation, measurable coupled interaction, and
language-grounded closed-loop execution.


\subsection{Physics and Physical Reasoning}
\label{sec:domains-physics}

Physics reads an episode from the world side: what interaction occurred, what
caused it, and how the cause changes under controlled variation. MagicSim joins
executable interaction, measurable state, and deterministic replay
(\S\ref{sec:engine}; \S\ref{sec:runtime}).

\paragraph{Large-scale coupled physical interaction data.}
The value is not a solver list, but co-instantiation inside one executable
episode. A pour couples robot, container, and fluid; garments drape against
furniture; rope, granular media, soft bodies, and articulated devices can share a
scene (\S\ref{sec:engine}). Contact and tactile channels record force, pressure,
and deformation, while serving turns batched episodes into corpora under seeded
variation (\S\ref{sec:sensors}; \S\ref{sec:collection}; \S\ref{sec:serving}).

\begin{figure}[!htb]
\centering
\includegraphics[width=0.98\linewidth]{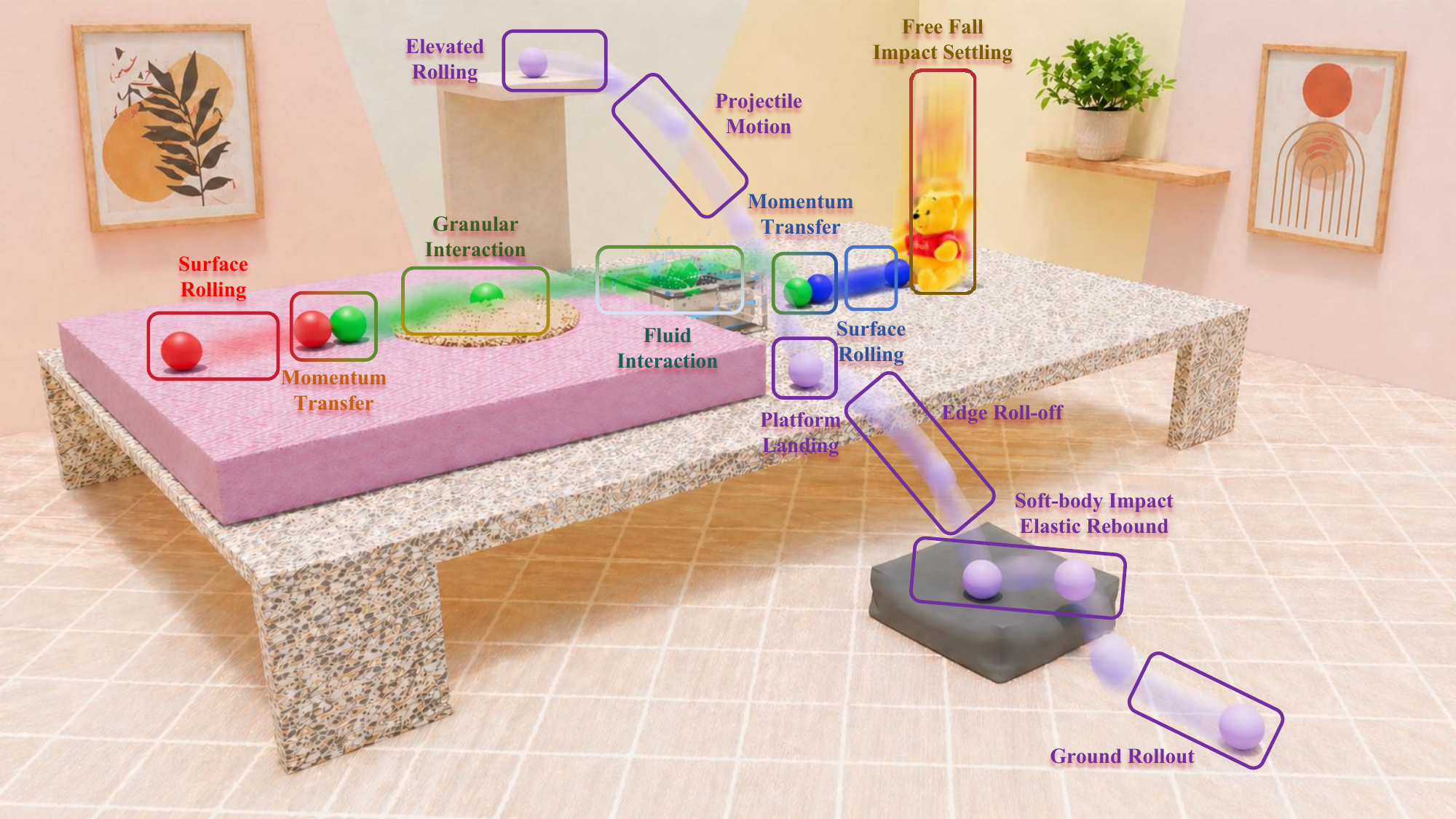}
\caption[Composite physics events in a shared MagicSim environment.]{
\textbf{Composite physics events.} A single scene contains rolling, collision,
granular and fluid interaction, free fall, impact, projectile motion, soft-body
contact, rebound, and dissipative rollout.}
\label{fig:composite-physics-events}
\end{figure}

\paragraph{Counterfactual reasoning and controlled variation.}
\texttt{reset\_to} plus a single-parameter intervention yields paired
counterfactuals: the same episode with one friction, mass, material, or initial
pose changed while other streams remain fixed (\S\ref{sec:dr-seed-streams}).
Read in reverse, the same variation becomes system-identification data when saved
state or metadata exposes the DR parameters. Domain randomization thus supports
both causal reasoning and physical-parameter inference.

\begin{figure*}[!htb] 
\centering 
\includegraphics[width=\textwidth]{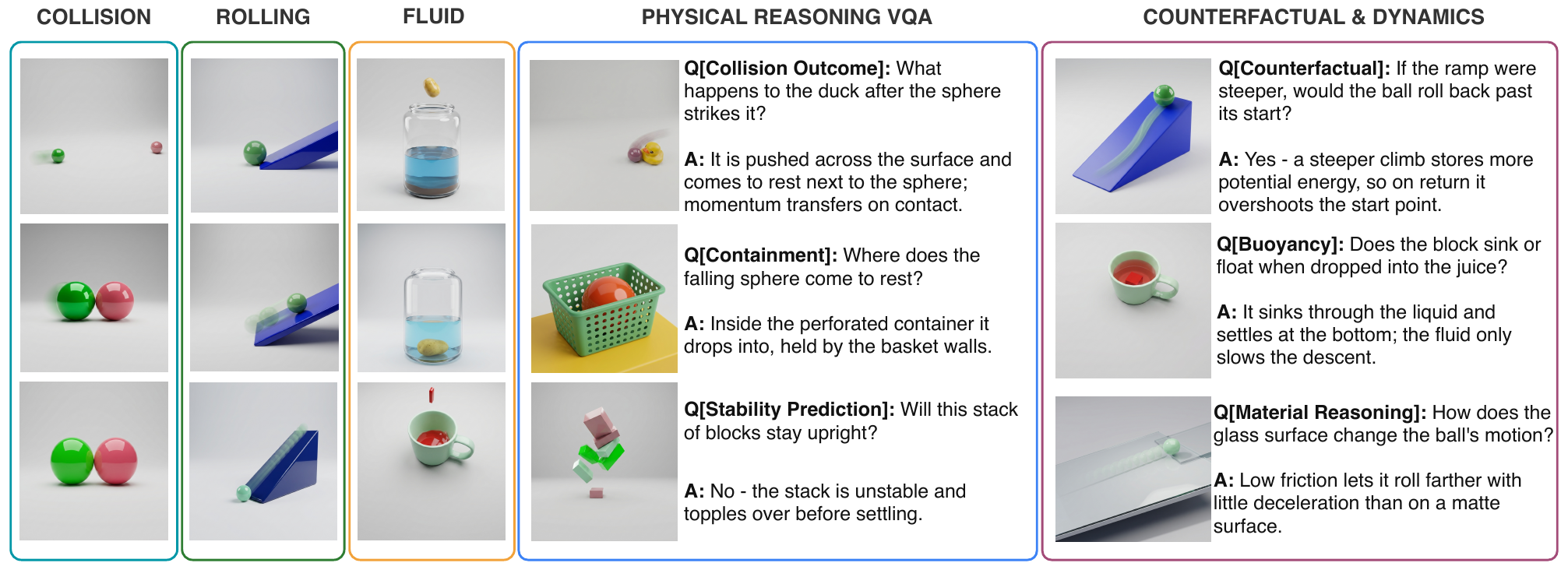} 
\caption{\textbf{Physics and physical reasoning supported by MagicSim.}
  Coupled interaction episodes---rigid collision, rolling on ramps and
  surfaces, and fluid---are recorded as deterministic, replayable rollouts
  with synchronized RGB observations and tier-level physical state. The same
  state converts into robot- and scene-centric VQA covering collision
  outcome, containment, and stability prediction, and into counterfactual and
  dynamics questions over ramp height, buoyancy, and surface friction, each
  grounded in single-parameter variation via \texttt{reset\_to}. These outputs
  also repackage as physically grounded reasoning data for dynamics
  prediction, physical-parameter (system) identification, and 4D-consistent
  generation.}
  \label{fig:physics_reasoning}
\end{figure*}

\paragraph{4D dynamic scene reconstruction and physically grounded generation.}
Each episode is an annotated 4D scene with time-varying geometry, depth, normals,
segmentation, motion vectors, and camera parameters over cloth, particles,
articulations, and contact-rich objects (\S\ref{sec:annotation}). This supplies
supervision for dynamic reconstruction and physics-grounded generation, including
Phys4D-style simulation-supervised training and 4D consistency
evaluation~\cite{lu2026phys4d}. The runtime pins each scene to seed, state, and
replay policy rather than a one-off render (\S\ref{sec:runtime}).


\subsection{VLM and Embodied Agent Learning}
\label{sec:domains-vlm}

\begin{figure*}[htbp] 
\centering 
\includegraphics[width=\textwidth]{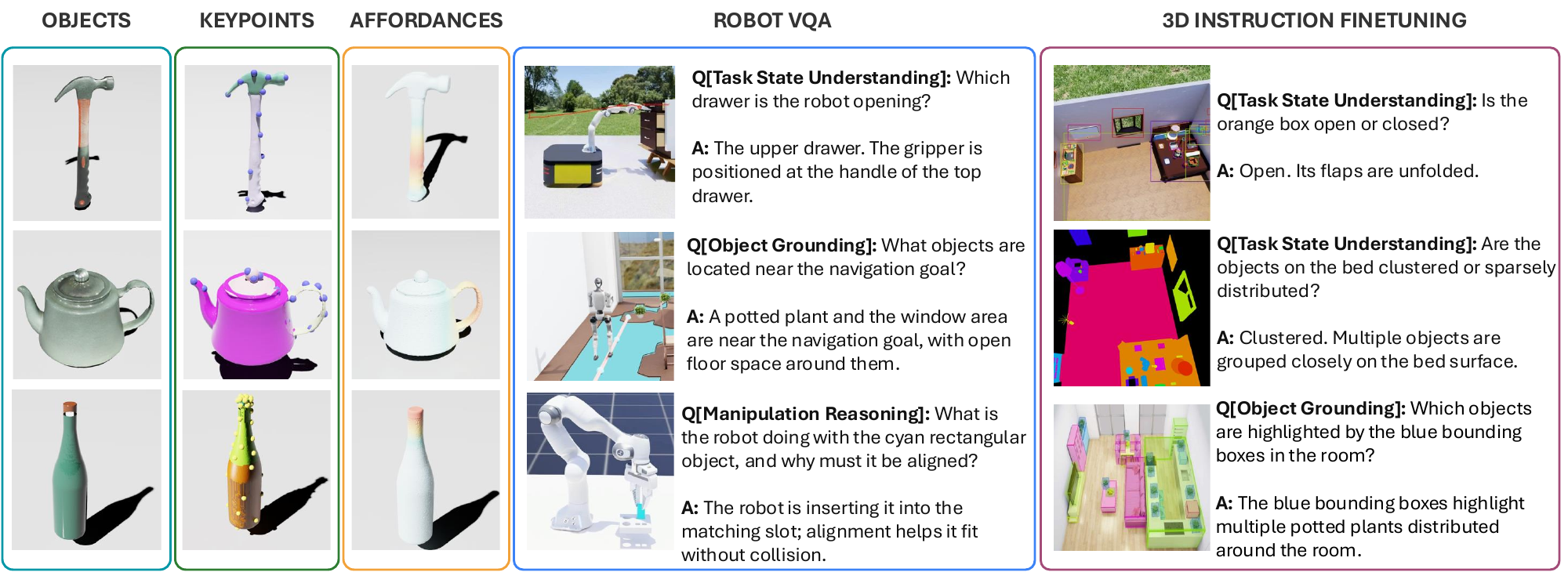} 
\caption{ Downstream tasks supported by MagicSim. Object, keypoint, and affordance annotations provide reusable grounding signals for perception and interaction. Runtime scenes, robot states, task states, and language traces convert the same supervision into robot-centric VQA examples covering object grounding, manipulation reasoning, navigation reasoning, and task-state understanding. These outputs can also be repackaged as 3D instruction fine-tuning data for language-grounded scene and action reasoning. } 
\label{fig:downstream-annotation-tasks} 
\end{figure*}

Embodied VLM research reads privileged state as supervision. Because MagicSim
knows poses, contacts, boxes, cameras, phases, skills, planner outcomes, and
language facts, embodied ability dimensions can be generated programmatically;
because the same world is interactive, those labels can become closed-loop
evaluations. Some post-training consumers remain planned, as noted in
\S\ref{sec:limitations}. Figure~\ref{fig:downstream-annotation-tasks} summarizes this supervision view: object, keypoint, affordance, action, and language traces can be repackaged as perception targets, robot-centric VQA, and 3D instruction-tuning data.

\paragraph{Spatial and multi-view understanding.}
Spatial relations, distances, counts, occlusion, and cross-view correspondence
are functions of privileged state. MagicSim can therefore synthesize QA pairs
across aligned egocentric and exocentric views from multi-camera and tiled
capture (\S\ref{sec:camera-capture}) \cite{wang2025mindcube, zhang2026spacenumrevisitingspatialnumerical}.

\paragraph{Affordance and embodied grounding.}
Affordance points, object boxes, end-effector waypoints, grasp poses, and room or
support-surface areas are direct products of the annotation schema
(\S\ref{sec:annotation}). The key claim is execution grounding: the affordance
corresponds to the planner-selected grasp target realized in a successful
trajectory \cite{Wu2024UniGarmentManipAU,li2023imagemanipimagebasedroboticmanipulation}.

\paragraph{Task progress and skill-structure understanding.}
Each saved frame carries command, skill, planner, phase, and status context.
Progress estimation, skill segmentation, and success/failure prediction therefore
inherit dense labels from the same state machines that generated the
demonstration (\S\ref{sec:collection}; \S\ref{sec:task-protocol})\cite{zhang2026progresslmprogressreasoningvisionlanguage}.

\paragraph{High-level planning and embodied reasoning.}
Each record also contains the instruction, realized skill sequence, planner
results, and outcome. This supports plan generation and critique---predict a
skill decomposition, predict the induced trajectory, or judge whether a sequence
can still succeed---with execution as the arbiter (\S\ref{sec:collection};
\S\ref{sec:agent-capability}).

\paragraph{Active spatial exploration and interactive evaluation.}
The same labels can serve static supervision and online evaluation. The
camera-only embodiment makes viewpoint selection an action; occupancy and scene
state support information-gain-style objectives; and the agent interface runs
commands across many parallel worlds (\S\ref{sec:flying-camera};
\S\ref{sec:collection-camera}; \S\ref{sec:agent-capability}). Post-training via
\texttt{verl}~\cite{sheng2024hybridflow} or \texttt{RLinf}~\cite{yu2025rlinf}
remains planned consumption, but the runtime already lets an ability benchmark
become an agent benchmark (\S\ref{sec:limitations}).
\FloatBarrier
\section{Discussion, Limitations, and Future Work}
\label{sec:limitations}
\label{sec:discussion}

MagicSim is already operational across the major components required by an
executable embodied-interaction stack: heterogeneous world construction,
deterministic batched execution, embodiment control, planning, sensing,
annotation, task MDPs, atomicskills, success-gated collection, serving, and
policy-facing interfaces. The remaining open questions are therefore not missing
top-level modules, but limits that appear when these components are run together
in heterogeneous, contact-rich, long-horizon episodes. We organize the discussion
around three axes that can always be pushed further: more realistic simulation,
stronger learned skills, and more complex longer-horizon tasks.

\subsection{Simulation Fidelity}
\label{sec:limitations-fidelity}

MagicSim supports broad co-execution of rigid bodies, articulated objects,
deformables, garments, fluids, granular media, rope-like objects, avatars,
robots, and sensors, but this should not be read as uniformly metric-accurate
physical prediction. Cross-solver contact can still produce interpenetration or
solver-dependent outcomes, especially for thin shells, composite or
multi-material objects, dense contact, deformables, fluids, and granular media.
Position-based and particle-based methods are useful because they are stable and
executable at task scale, but they remain approximations rather than
material-identification-grade models. In practice, collision geometry, contact
offsets, friction, damping, stiffness, particle spacing, and solver iterations
can still require asset-specific tuning, and large fine-particle scenes remain
limited by memory, compute cost, and stability. Physical realism is therefore a
real source of headroom in the current system.

\subsection{Policy Performance and Skill Coverage}
\label{sec:limitations-policy}

MagicSim includes scripted atomicskills, planner-grounded primitives, learned
controllers, model-predictive backends, and policy-facing task interfaces, so the
collection-to-training-to-deployment loop exists. The limitation is that current
behavior still falls short on harder regimes. Contact-rich manipulation,
dexterous manipulation, deformable-object interaction, humanoid
loco-manipulation, and long-horizon multi-command episodes expose failures in
success rate, robustness, and generalization. The fixed skill vocabulary is
inspectable and reusable, but it only covers behaviors that have been engineered,
trained, or explicitly provided as backends. Geometric planning is strong for
reachable poses, goalsets, collision-aware trajectories, navigation targets, and
gripper commands, but weaker when success depends on rolling, sliding,
compliance, deformation history, force regulation, or long chains of partially
observed physical effects. The default success-gated corpus is also biased
toward recoverable successes, so saved demonstrations should not be interpreted
as an unbiased sample of all attempted interactions.

\subsection{Future Work}
\label{sec:future-work}

\subsubsection{More Realistic Simulation}
\label{sec:fw-simulation}

A first direction is to improve physical fidelity, especially contact and
material behavior at solver boundaries. This includes reducing penetration,
improving thin-shell and composite-object behavior, expanding multi-material
parameterization, and making deformation, cutting, pouring, and granular
interaction more faithful. Longer term, unified or differentiable simulation
layers could make heterogeneous episodes more stable, measurable, and
controllable under intervention.

\subsubsection{Stronger Learned Skills}
\label{sec:fw-learned-skills}

A second direction is stronger learned behavior both below and above the
AtomicSkill interface. Below the interface, learned and model-predictive
backends should improve contact-rich manipulation, dexterous control,
deformable manipulation, whole-body loco-manipulation, and recovery. Above the
interface, learned command or skill selectors can choose what to try next while
the runtime still grounds every world change through executable planners,
controllers, sensors, and task predicates.

\subsubsection{More Complex, Longer-Horizon Tasks}
\label{sec:fw-tasks}

A third direction is to make tasks longer, denser, and more compositional. HRI,
tool-use-style manipulation, deformable manipulation, mobile manipulation,
humanoid loco-manipulation, autonomous-laboratory operations, and camera-based
exploration are already part of the system; the next step is to combine these
ingredients into episodes where success depends on persistent world state,
recovery, cross-subtask dependencies, and coordinated perception, navigation,
manipulation, and human-facing behavior.

\subsection{Broader Positioning}
\label{sec:positioning}

MagicSim is best understood as executable data infrastructure for embodied
intelligence. It is not only a physics engine, benchmark, data generator, or
agent interface: its value is the shared episode substrate that lets one rollout
be used as an evaluation trial, a success-gated multimodal trajectory, a
replayable physical experiment, or an agent-facing interaction. The main future
direction is therefore not to add isolated components, but to keep pushing the
three axes that make this substrate more useful: more realistic simulation,
stronger learned skills, and more complex longer-horizon tasks.
\FloatBarrier

\clearpage
\appendix
\begin{center}
{\LARGE\bfseries Appendix}\par
\vspace{1em}
\end{center}
\addcontentsline{toc}{section}{Appendix}
\section{Manager Lifecycle Reference}
\label{app:manager-lifecycle}

This appendix documents the concrete lifecycle contract that realizes the
deterministic manager runtime of Section~\ref{sec:runtime}. It specifies the
Env-Core wrapper chain, the per-manager lifecycle methods, the fixed
construction and reset order, the seeding fan-out, the P1/P2/P3 object lifecycle
modes, and the snapshot/replay (\texttt{reset\_to}) capability of each manager.
All names refer to classes and methods under
\texttt{src/magicsim/Env/}.

\subsection{Env-Core Wrapper Chain}
\label{app:manager-wrapper-chain}

MagicSim assembles managers through a linear wrapper stack. Each wrapper adds
one group of managers and forwards the shared reset/step lifecycle to the
managers it owns, so a full environment is a composition of wrappers rather than
a monolithic class.

\begin{table}[htbp]
\centering
\small
\setlength{\tabcolsep}{4pt}
\begin{tabularx}{\linewidth}{>{\ttfamily}p{0.26\linewidth} X}
\toprule
Wrapper & Managers introduced at this level \\
\midrule
BaseEnv & Launches Isaac Sim host (\texttt{IsaacRLEnv} / \texttt{SimulationContext}); owns the launch manifest and seed normalization. \\
SyncBaseEnv & Adds world-core managers: \texttt{TerrainManager}, \texttt{LayoutManager}, \texttt{SceneManager}, \texttt{AvatarManager} (if \texttt{anim}), \texttt{NavManager} (if \texttt{nav}). \\
SyncCollectEnv & Adds the capture layer: \texttt{CameraManager}, \texttt{CaptureManager} / \texttt{TiledCaptureManager}. \\
SyncRobotEnv & Adds \texttt{RobotManager} and \texttt{PlannerManager}; exposed to the task layer as \texttt{SyncRobotEnv}. \\
SyncCameraEnv & Camera-only sibling of \texttt{SyncRobotEnv}; adds \texttt{CameraPlannerManager} for flying-camera motion. \\
\bottomrule
\end{tabularx}
\caption{The Env-Core wrapper chain. \texttt{SyncRobotEnv} and
\texttt{SyncCameraEnv} are the two terminal branches;
\texttt{TaskBaseEnv} (Section~\ref{sec:task-contract}) wraps
\texttt{SyncRobotEnv} to expose the MDP boundary.}
\label{tab:app-wrapper-chain}
\end{table}

\subsection{Common Manager Lifecycle}
\label{app:manager-lifecycle-methods}

Every manager follows the same lifecycle shape, which is what lets the wrapper
chain drive the world through one ordered sequence rather than accumulating
subsystem-specific reset paths.

\begin{itemize}
\item \texttt{\_\_init\_\_(num\_envs, config, device, seeds\_per\_env, \dots)}:
  construct from configuration and per-environment seeds; allocate per-env state
  containers but touch no live stage.
\item \texttt{update\_env\_seeds(seeds)}: (re)derive the manager's per-env random
  streams from the current global seed fan-out (\S\ref{app:manager-seeding}).
\item \texttt{initialize(sim)}: bind to the live \texttt{SimulationContext},
  create stage prims, and register physics/render views. Called once during
  \texttt{\_setup\_scene}.
\item \texttt{post\_init()} / \texttt{post\_setup()}: finalize bindings that
  require the cloned, played stage (e.g., camera xforms, terrain prims under env
  roots).
\item \texttt{reset(soft=False)}: reset the full batch.
\item \texttt{reset\_idx(env\_ids, soft)}: reset a selected subset; \texttt{soft}
  distinguishes re-randomization in place from a hard rebuild.
\item \texttt{step(action, env\_ids)} / \texttt{update(info)}: advance
  action-consuming managers (robot, planner, camera, capture) within the tick.
\item \texttt{get\_state()} / \texttt{set\_state(state, env\_ids)} /
  \texttt{reset\_to(state, env\_ids)}: snapshot and restore manager-owned initial
  conditions (\S\ref{app:manager-snapshot}).
\end{itemize}

Table~\ref{tab:app-manager-methods} summarizes which managers own which
lifecycle stages and the state each manager holds.

\begin{table*}[htbp]
\centering
\footnotesize
\setlength{\tabcolsep}{4pt}
\renewcommand{\arraystretch}{1.12}
\begin{tabularx}{\textwidth}{>{\ttfamily}p{0.16\textwidth} p{0.30\textwidth} c c c c X}
\toprule
Manager & State owned & init & reset\_idx & step & reset\_to & Random stream \\
\midrule
TerrainManager & ground/room carriers, mutable appearance & \Y & \Y & \N & \N (locked geometry) & \texttt{terrain} (stage-wide) \\
LayoutManager & per-env pose plans over logical objects & \Y & \Y & \N & \Pt (via \texttt{set\_state}) & \texttt{layout} \\
SceneManager & physical/artifact object instances, background & \Y & \Y & \N & \Y & (per-env raw seed) \\
AvatarManager & animated human instances, animation state & \Y & \Y & on\_update & \Y & \texttt{animation} \\
NavManager & navmesh volume, occupancy/navmesh submanagers & \Y & \N & \N & \N & --- \\
CameraManager & camera prims and poses & \Y & \Y & \Y & \Y & \texttt{camera} \\
CaptureManager & render products, annotators (derived sink) & \Y & \Y & \Y & \N & --- \\
RobotManager & articulations, per-robot action/obs managers & \Y & \Y & \Y & \Y & \texttt{robot} \\
PlannerManager & per-robot planner/solver services & \Y & \Y & \Y & \N & --- \\
CameraPlannerManager & per-camera flying-camera planners & \Y & \Y & \Y & \N & --- \\
\bottomrule
\end{tabularx}
\caption{Per-manager lifecycle coverage and owned state.
\Y~=~implemented, \Pt~=~partial/derived, \N~=~not applicable.
Capture is a derived sink with no reproducible simulator state, so it has no
\texttt{reset\_to}; planner and nav managers hold no randomized state and inherit
determinism from the launch seed and their inputs.}
\label{tab:app-manager-methods}
\end{table*}

\subsection{Construction and Reset Order}
\label{app:manager-order}

Because the batched simulator is not globally stopped or rebuilt when a single
environment resets, the order in which managers are invoked is part of the
determinism contract. Two orderings matter.

\paragraph{Construction (\texttt{\_setup\_scene}).}
Managers are initialized bottom-up along the wrapper chain, and within
\texttt{SyncBaseEnv} the order is dictated by USD reference resolution:
(1)~\texttt{TerrainManager} (pre-cloner ground/generator work),
(2)~\texttt{LayoutManager} (plan the first layout before objects pull poses),
(3)~\texttt{SceneManager} (spawn pre-load objects, pulling poses via
\texttt{get\_layout}),
(4)~\texttt{AvatarManager},
(5)~\texttt{NavManager} (create the navmesh volume before cloning),
then \texttt{CameraManager}, \texttt{CaptureManager}, \texttt{RobotManager}, and
\texttt{PlannerManager}. A post-setup pass runs after IsaacLab clones the
sub-environments (\texttt{TerrainManager.post\_setup}, camera
\texttt{post\_init}, and nav volume-bounds configuration against the room bbox).

\paragraph{Reset.}
Reset preserves the same layout-first invariant:
\texttt{LayoutManager.reset\_idx} runs before \texttt{SceneManager.reset\_idx}
because scene objects call \texttt{get\_layout} while (re)binding to prims;
\texttt{TerrainManager} is always reset with \texttt{soft=True} because its
geometry is locked; camera, capture, robot, and planner managers follow. Partial
reset dispatches each requested \texttt{env\_id} into either a hard or a soft
reset based on a per-environment soft-reset counter (or a
\texttt{force\_hard\_reset} option), so environments that have exceeded the
soft-reset budget are rebuilt while their neighbors keep stepping.

\paragraph{Step ordering.}
The step contract is intentionally simple: actions are applied
(\texttt{PlannerManager.step} then \texttt{RobotManager.step}, or camera-planner
then camera for \texttt{SyncCameraEnv}), the batched simulator advances one
physics tick, and post-physics \texttt{update} passes advance the per-env state
machines and read back observations.

\subsection{Seeding Fan-Out}
\label{app:manager-seeding}

A single input seed expands into disjoint, manager-local streams so that
unrelated random choices never couple (Section~\ref{sec:dr-seed-streams}). The
input seed is normalized by \texttt{BaseEnv} into a per-environment seed vector:
\texttt{None} yields a non-deterministic run, an integer $N$ yields
$[N, N{+}1, \dots, N{+}\mathrm{num\_envs}{-}1]$, and an explicit list is used
directly. On every reset, \texttt{\_update\_seed\_managers} pushes this vector to
each manager, which derives its own stream via
\texttt{stream\_seed(seed\_i, \textnormal{name})}.

\begin{table}[htbp]
\centering
\small
\begin{tabularx}{\linewidth}{>{\ttfamily}p{0.20\linewidth} >{\ttfamily}p{0.22\linewidth} X}
\toprule
Manager & Stream name & Scope \\
\midrule
LayoutManager & layout & per-env \\
SceneManager & (raw per-env seed) & per-env; background uses \texttt{seed[0]} \\
TerrainManager & terrain & stage-wide (\texttt{seed[0]}) \\
AvatarManager & animation & per-env \\
CameraManager & camera & per-env \\
RobotManager & robot & per-env \\
\bottomrule
\end{tabularx}
\caption{Manager-local random streams. Adding a new draw inside one stream cannot
perturb another; resetting one environment does not advance another.}
\label{tab:app-seed-streams}
\end{table}

\subsection{Object Lifecycle Modes (P1/P2/P3)}
\label{app:manager-lifecycle-modes}

Once \texttt{SimulationContext.reset()} locks the USD stage, whether an object can
be dynamically imported or deleted is a backend capability rather than a runtime
preference. \texttt{SceneManager} therefore collapses each object's two backend
capabilities (dynamic import, dynamic delete) into three modes, and the hard-reset
scheduler dispatches on the mode rather than on the object type
(Section~\ref{sec:lifecycle-modes}).

\begin{table}[htbp]
\centering
\small
\begin{tabularx}{\linewidth}{>{\bfseries}p{0.10\linewidth} p{0.26\linewidth} X}
\toprule
Mode & Backend capability & Hard-reset behavior \\
\midrule
P1 & import + delete & Destroy the current prim, create a replacement, optionally choose a different asset (``volatile''). \\
P2 & import, no delete & Retire the old prim (disable gravity/collision, hide, move aside) and create the replacement at a fresh \texttt{prim\_path} (``persistent with zombies''). \\
P3 & neither after lock & Build once before the stage lock; only re-randomize state and pose in place (``pre-load''). \\
\bottomrule
\end{tabularx}
\caption{Object lifecycle modes. Logical identity
\texttt{(env\_id, cat\_name, inst\_id)} is stable across all modes; only the
physical \texttt{prim\_path} may be reallocated.}
\label{tab:app-lifecycle-modes}
\end{table}

\subsection{Snapshot and Replay}
\label{app:manager-snapshot}

State restoration through \texttt{reset\_to} is part of the lifecycle contract, not
a separate log replay: each manager restores its own initial conditions. The
capability inherits the lifecycle-mode limits of
Table~\ref{tab:app-lifecycle-modes}. A P1 or P2 snapshot can be restored with a
different selected asset because the prim can be rebuilt or reallocated, whereas
a P3 object can only restore pose and randomized state onto the existing asset.
\texttt{LayoutManager} restores a plan by arming a one-shot replay policy that the
next \texttt{reset\_idx} consumes; \texttt{SceneManager}, \texttt{AvatarManager},
\texttt{CameraManager}, and \texttt{RobotManager} implement full
\texttt{get\_state}/\texttt{set\_state}/\texttt{reset\_to}; \texttt{TerrainManager}
exposes state but keeps geometry locked; and \texttt{CaptureManager},
\texttt{PlannerManager}, and \texttt{NavManager} hold no reproducible simulator
state and therefore expose no snapshot.

\section{Scene YAML Schema}
\label{app:scene-yaml}

MagicSim follows the ``configuration over code'' principle of
Section~\ref{sec:introduction}: scenes, layouts, assets, robots, cameras,
sensors, randomization, and task parameters are declared through structured
YAML rather than simulator code. This appendix specifies that schema. Env-Core
configuration files live under \texttt{src/magicsim/Env/Conf/}; task
configuration files live under \texttt{src/magicsim/Task/*/Conf/}. The schema is
the same for both; a task file simply nests the Env-Core blocks under an
\texttt{env:} key (\S\ref{app:yaml-toplevel}). The runtime resolution of ranges,
asset selection, and the \texttt{ratio} gate is implemented in
\texttt{Env/Scene/config\_resolve.py}, and the object schema is documented in
\texttt{Env/Scene/objects.md}.

\subsection{Top-Level Structure}
\label{app:yaml-toplevel}

An Env-Core config declares a subset of the following top-level blocks.

\begin{table}[htbp]
\centering
\small
\begin{tabularx}{\linewidth}{>{\ttfamily}p{0.14\linewidth} X}
\toprule
Block & Role \\
\midrule
sim & Simulation engine and batch settings: device, fabric, decimation, seed, soft-reset budget, \texttt{scene.num\_envs}, \texttt{env\_spacing}, PhysX/render options. \\
scene & Background, lighting, and the object families of Section~\ref{sec:engine}. \\
robot & Embodiments, action channels, and planner blocks (\S\ref{app:yaml-robot}). \\
camera & Cameras, capture backend, and annotators (\S\ref{app:yaml-camera}). \\
anim & Animation-driven avatars (\S\ref{app:yaml-anim}). \\
nav & Navmesh baking parameters (agent height, radius, step, slope). \\
\bottomrule
\end{tabularx}
\caption{Env-Core top-level blocks. A task configuration file wraps these under
\texttt{env:} and adds a sibling \texttt{task:} block; the AutoCollect driver
adds a \texttt{collect:} block (\S\ref{app:yaml-task}).}
\label{tab:app-yaml-toplevel}
\end{table}

A task file therefore has the shape
\begin{verbatim}
env:            # Env-Core substrate (sim/scene/robot/camera/...)
  sim: {...}
  scene: {...}
  robot: {...}
task:           # MDP-facing layer
  observation: {...}
  reward: {...}
  termination: {timeout_steps: ...}
  command: {...} # optional task parameters
collect:        # only under AutoCollect
  command: {...}
\end{verbatim}
\texttt{TaskBaseEnv} reads \texttt{config.env} for the Core managers and
\texttt{config.task} for observation, reward, termination, and command
(Section~\ref{sec:task-contract}); under AutoCollect,
\texttt{config.collect.command} is injected into \texttt{config.task.command}
so the task reads the same parameters regardless of driver.

\subsection{The Randomization Shape Rule}
\label{app:yaml-shape-rule}

Every numeric field in the schema is resolved by shape, so the same key can be a
fixed value or a sampling range without a separate flag
(Section~\ref{sec:dr-seed-streams}):
a scalar (\texttt{0.5}) is a constant; a one-dimensional pair (\texttt{[a, b]})
is sampled uniformly; and a pair of vectors (\texttt{[[x0,y0,z0],[x1,y1,z1]]})
is sampled component-wise. The \texttt{ratio} field is not a physical
coefficient but a policy gate: it decides whether a perturbation, distractor,
material change, or physics variation is active for a given environment or reset,
applying a multiplicative jitter of the form
$\mathrm{value}\pm\mathrm{value}\times(\mathrm{ratio}-1)$.

\subsection{sim Block}
\label{app:yaml-sim}

\begin{table}[htbp]
\centering
\small
\begin{tabularx}{\linewidth}{>{\ttfamily}p{0.28\linewidth} X}
\toprule
Key & Meaning \\
\midrule
device & \texttt{cuda:0} or \texttt{cpu}. \\
use\_fabric & Enable the PhysX Fabric backend. \\
decimation & Simulation sub-steps per environment step. \\
seed & Launch seed: integer or per-env list (\S\ref{app:manager-seeding}). \\
soft\_resets & Soft-reset budget before a hard reset is forced. \\
scene.num\_envs & Number of batched sub-environments. \\
scene.env\_spacing & Distance between sub-environment origins. \\
physx.* & GPU solver limits, e.g. \texttt{gpu\_collision\_stack\_size}, \texttt{gpu\_max\_particle\_contacts}. \\
render.* & Optional render options (translucency, reflections, DLSS). \\
\bottomrule
\end{tabularx}
\caption{\texttt{sim} block fields.}
\label{tab:app-yaml-sim}
\end{table}

\subsection{scene Block}
\label{app:yaml-scene}

The \texttt{scene} block declares the world's appearance and its interactive
content. It contains \texttt{background} (dome-light intensity range and a list
of HDR IBL textures), \texttt{light} (one or more named light sources, each with
a \texttt{types} list and per-type intensity/size ranges plus a shared
\texttt{common} block for pose, color temperature, and color scale), and
\texttt{objects}.

\paragraph{Unified object schema.}
Each entry under \texttt{objects} is a logical category $c$ realized as one of
the simulation families of Section~\ref{sec:engine}. A category declares its
asset source(s) and instance count, an optional \texttt{common} block of
category defaults, and one block per instance that may override those defaults.

\begin{table}[htbp]
\centering
\small
\begin{tabularx}{\linewidth}{>{\ttfamily}p{0.24\linewidth} X}
\toprule
Key & Meaning \\
\midrule
num\_per\_env & Instances of this category per environment. \\
random & \texttt{true}: sample the asset from \texttt{usd}; \texttt{false}: pick deterministically by \texttt{(env\_id, inst\_id)}. \\
usd & List of asset sources: USD files, folders (expanded to member USDs), or primitive names (\texttt{Cube}, \texttt{Sphere}, \dots). \\
semantic\_label & Semantic tag written into USD semantics; falls back to filename or primitive name. \\
common.pos / ori / scale & Category-default pose, orientation (Euler degrees), and scale; scalar or range. \\
common.primitive\_params & Size/radius/height/resolution for primitive assets. \\
common.visual & \texttt{color} (priority~1) or \texttt{visual\_material} (priority~2, MDL/USD material list). \\
common.physics & \texttt{type} (family) plus family-specific parameters (Table~\ref{tab:app-yaml-physics-families}) and the \texttt{ratio} gate. \\
\texttt{<inst>\_k} & Per-instance overrides (scalar pose/ori/scale, visual, physics). \\
\bottomrule
\end{tabularx}
\caption{Unified object schema. Resolution priority is instance $>$ common $>$
class default; material priority is \texttt{color} $>$ \texttt{visual\_material}
$>$ none.}
\label{tab:app-yaml-object}
\end{table}

\paragraph{Per-family physics parameters.}
The \texttt{physics.type} field selects the simulation family and the set of
solver-specific parameters exposed under \texttt{physics}
(Section~\ref{sec:engine-supported-families}).

\begin{table}[htbp]
\centering
\small
\begin{tabularx}{\linewidth}{>{\ttfamily}p{0.20\linewidth} X}
\toprule
\texttt{type} & Representative parameters \\
\midrule
rigid & mass or density, collision, contact/rest offset, contact-slop, linear/angular velocity, physics\_material (static/dynamic friction, restitution, combine modes). \\
geometry & static collision geometry; contact-sensor and filter options. \\
articulation & joint state, joint limits, drives, link collision, rigid material. \\
rope & rope\_length, num\_links, link\_radius, cone-angle limit, stiffness, damping, friction. \\
deformable & FEM tetrahedral resolution, Young's modulus, Poisson ratio, damping, contact/rest offset, self-collision. \\
fem\_cloth & density, thickness, bending stiffness, position-iteration count, velocity damping, Poisson ratio. \\
garment & particle\_system (iterations, contact/rest offset, self-collision), particle\_material (friction, damping, drag, lift, cohesion), garment\_config (particle mass, stretch/bend/shear stiffness, spring damping). \\
inflatable & particle\_system + particle\_material + inflatable\_config (pressure, spring stiffness). \\
fluid & particle\_system (contact/rest offset, fluid-rest offset, smoothing, anisotropy, isosurface), particle\_material (viscosity, cohesion, surface tension), optional \texttt{container} binding. \\
sand / rice & granular particle radius or spacing, density, friction, cohesion-like terms, solver iterations. \\
\bottomrule
\end{tabularx}
\caption{Family-specific \texttt{physics} parameters. Appearance and physics
randomization reuse the shape rule of \S\ref{app:yaml-shape-rule}.}
\label{tab:app-yaml-physics-families}
\end{table}

\subsection{robot Block}
\label{app:yaml-robot}

Each entry under \texttt{robot} declares one embodiment instance (see
Appendix~\ref{app:robot-list} for the registered embodiments). The key fields are
\texttt{name} (the registered embodiment id, e.g. \texttt{franka}, \texttt{g1}),
\texttt{type} (control family, e.g. \texttt{manipulator}, \texttt{mobilemanip},
\texttt{humanoid}), \texttt{common} (base \texttt{pos}/\texttt{ori} and
\texttt{initial\_joint\_pos}), and \texttt{action}, which fills the
\texttt{base}/\texttt{arm}/\texttt{eef} channels of
Section~\ref{sec:control-robotmanager} with action terms such as
\texttt{joint\_pos}, \texttt{ik\_abs}, \texttt{ik\_rel}, \texttt{ik\_pink},
\texttt{binary} (gripper), or a base drive
(\texttt{holonomic\_action}, \texttt{differential\_drive}). A \texttt{reset}
policy (\texttt{default} or \texttt{randomize\_joint} with mean/std) controls
initial-state randomization. Robots that use motion planning declare a
\texttt{planner\_type: Curobo} and a \texttt{planner} block with \texttt{ik} and
\texttt{motiongen} sub-blocks (instance counts, batch sizes, seed counts,
position/rotation thresholds, \texttt{relative\_to\_world\_frame}), an optional
\texttt{dual\_mode} with virtual \texttt{base\_joint\_names}, and a \texttt{body}
sub-block that parameterizes the base P-controller and dynamic-window planner
(\S\ref{sec:control-mid}).

\subsection{camera Block}
\label{app:yaml-camera}

The \texttt{camera} block carries global capture settings (\texttt{enable\_tiled}
to select \texttt{TiledCaptureManager} vs \texttt{CaptureManager},
\texttt{colorize\_depth}, \texttt{output\_dir}) and one sub-block per camera. Each
camera sub-block splits into \texttt{camera} (creation) and \texttt{annotator}
(capture). Pinhole cameras are configuration-driven
(\texttt{resolution}, \texttt{frequency}, \texttt{focal\_length},
\texttt{focus\_distance}, \texttt{clipping\_range}); Kinect- and RealSense-style
cameras set \texttt{mesh} and a fixed \texttt{mode}; link-mounted cameras set
\texttt{mount\_link} to ride a robot body. Pose (\texttt{pos}/\texttt{ori})
follows the shape rule, so viewpoint randomization is expressed as a range. The
\texttt{annotator} sub-block enables Omni Replicator channels (\texttt{rgb},
\texttt{distance\_to\_image\_plane}, \texttt{bounding\_box\_3d}, etc.), each with
a \texttt{device} field (Section~\ref{sec:annotation-capture}).

\subsection{anim Block}
\label{app:yaml-anim}

The \texttt{anim} block declares animation-driven avatars
(Section~\ref{sec:engine-supported-families}). Under \texttt{anim.avatar}, each
character family sets \texttt{usd} (character asset folder), \texttt{num\_per\_env},
\texttt{random}, \texttt{semantic\_label}, and a \texttt{common} block for
\texttt{pos}/\texttt{ori}/\texttt{scale}. Avatars are environment-side actors,
not robot channels, and are advanced by environment dynamics rather than the
action vector (Section~\ref{sec:task-diversity}).

\subsection{task and collect Blocks}
\label{app:yaml-task}

The \texttt{task} block defines the MDP-facing layer:
\texttt{observation} and \texttt{reward} specifications (often empty when a task
uses defaults), \texttt{termination} (at least \texttt{timeout\_steps}), and an
optional \texttt{command} block of task parameters. Under AutoCollect, the
\texttt{collect} block carries the command distribution
(\texttt{config.collect.command}) that is sampled per environment and injected
into the task's command slot, so training, evaluation, and collection differ by
command distribution rather than by separate task definitions
(Section~\ref{sec:task-protocol}). The command grammar consumed by the skill
layer is specified in Appendix~\ref{app:skill-list}.

\section{Robot Embodiment List}
\label{app:robot-list}

This appendix enumerates the robot embodiments registered with
\texttt{RobotManager} (Section~\ref{sec:control-robotmanager}). Embodiments are
registered in the \texttt{ROBOT\_DICT} table under
\texttt{src/magicsim/Env/Robot/Cfg/}, organized into seven embodiment categories
(the subdirectories \texttt{Manipulator}, \texttt{DualManipulator},
\texttt{Dexterous}, \texttt{Mobile}, \texttt{MobileManip}, \texttt{Humanoid},
\texttt{Quadruped}). The current registry contains \textbf{34 entries over 33
distinct configuration classes} (the humanoid key \texttt{g1\_mobile} is an alias
of \texttt{g1}), consistent with the ``roughly thirty-three registered robots''
of Section~\ref{sec:control-robotmanager}. New embodiments enter through an
import pipeline and a single \texttt{ROBOT\_DICT} entry without touching any layer
above the channel interface.

Regardless of morphology, every embodiment is reduced at configuration time to
at most three action channels---\texttt{base}, \texttt{arm}, \texttt{eef}---each
filled by an action term from the shared library or left absent where it does not
apply (Section~\ref{sec:control-low}). Planners, skills, and the three drivers
address channels rather than robots, so a heterogeneous team is configured no
differently from a single arm.

\begin{table*}[htbp]
\centering
\footnotesize
\setlength{\tabcolsep}{4pt}
\renewcommand{\arraystretch}{1.1}
\begin{tabularx}{\textwidth}{>{\ttfamily}p{0.17\textwidth} p{0.30\textwidth} c c c X}
\toprule
Registry key & Embodiment & base & arm & eef & End-effector / notes \\
\midrule
\multicolumn{6}{l}{\textbf{Single-arm manipulators} (\texttt{Manipulator})}\\
\midrule
franka          & Franka Emika Panda                 & \N & \Y & \Y & parallel gripper \\
frankarobotiq   & Franka + Robotiq 2F                & \N & \Y & \Y & parallel gripper \\
franka\_umi     & Franka (UMI gripper variant)       & \N & \Y & \Y & parallel gripper \\
franka\_tactile & Franka with tactile sensing        & \N & \Y & \Y & parallel gripper + taxel sensing \\
xarm7           & UFACTORY xArm~7                     & \N & \Y & \Y & parallel gripper \\
ur10            & Universal Robots UR10               & \N & \Y & \Y & parallel gripper \\
ur10e           & Universal Robots UR10e + Robotiq    & \N & \Y & \Y & parallel gripper \\
ur5e            & Universal Robots UR5e + Robotiq     & \N & \Y & \Y & parallel gripper \\
piper\_x        & AgileX Piper                        & \N & \Y & \Y & parallel gripper \\
arx\_x5         & ARX X5                              & \N & \Y & \Y & parallel gripper \\
so101           & SO-ARM100 (SO101)                   & \N & \Y & \Y & parallel gripper \\
openarm         & OpenArm (unimanual)                 & \N & \Y & \Y & parallel gripper \\
\midrule
\multicolumn{6}{l}{\textbf{Dual-arm manipulators} (\texttt{DualManipulator})}\\
\midrule
dual\_franka    & Bimanual Franka                     & \N & \Y & \Y & two parallel grippers \\
dual\_piper     & Bimanual Piper                      & \N & \Y & \Y & two parallel grippers \\
dual\_arx\_x5   & Bimanual ARX X5                     & \N & \Y & \Y & two parallel grippers \\
dual\_so101     & Bimanual SO-ARM100                  & \N & \Y & \Y & two parallel grippers \\
dual\_openarm   & Bimanual OpenArm                    & \N & \Y & \Y & two parallel grippers \\
xtrainer        & Xtrainer dual-arm platform          & \N & \Y & \Y & two end-effectors, cuRobo IK \\
\midrule
\multicolumn{6}{l}{\textbf{Dexterous-hand manipulators} (\texttt{Dexterous})}\\
\midrule
franka\_xhand   & Franka + dexterous hand             & \N & \Y & \Y & multi-finger dexterous hand \\
sharpa\_wave\_floating & Floating SharpaWave hand     & \N & \N & \Y & dexterous hand, floating base \\
\midrule
\multicolumn{6}{l}{\textbf{Mobile bases} (\texttt{Mobile})}\\
\midrule
novacarter      & NVIDIA Nova Carter                  & \Y & \N & \N & differential/Ackermann drive \\
leatherback     & Leatherback platform                & \Y & \N & \N & holonomic/differential drive \\
\midrule
\multicolumn{6}{l}{\textbf{Mobile manipulators} (\texttt{MobileManip})}\\
\midrule
ridgebackfranka & Ridgeback + Franka                  & \Y & \Y & \Y & omnidirectional base + parallel gripper \\
ridgebacksawyer & Ridgeback + Sawyer                  & \Y & \Y & \Y & omnidirectional base + gripper \\
vega1p\_sharpa  & Vega-1P + SharpaWave                & \Y & \Y & \Y & mobile base + dexterous hand \\
genie1          & Genie-1 mobile dual-arm             & \Y & \Y & \Y & mobile base + two grippers \\
mobile\_x7s     & Mobile X7s dual-arm                 & \Y & \Y & \Y & holonomic base + two grippers \\
lift2           & Lift2 mobile manipulator            & \Y & \Y & \Y & mobile base + parallel gripper \\
\midrule
\multicolumn{6}{l}{\textbf{Humanoids} (\texttt{Humanoid})}\\
\midrule
g1              & Unitree G1 (whole-body controller)  & \Y & \Y & \Y & learned WBC base; two arms + grippers \\
g1\_mobile      & G1 (alias of \texttt{g1})           & \Y & \Y & \Y & alias entry \\
g1\_sonic       & G1 with SONIC upper-body pipeline   & \Y & \Y & \Y & Pink-IK reference realized by learned policy \\
g1\_dex1        & G1 with Dex1 grippers               & \Y & \Y & \Y & two arms + dexterous grippers \\
g1\_fixed\_hand & G1 with rigid hands                 & \Y & \Y & \N & two arms + fixed end-effectors \\
\midrule
\multicolumn{6}{l}{\textbf{Quadrupeds} (\texttt{Quadruped})}\\
\midrule
go2             & Unitree Go2                         & \Y & \N & \N & learned locomotion policy \\
\bottomrule
\end{tabularx}
\caption{Registered robot embodiments in \texttt{ROBOT\_DICT}, grouped by the
seven embodiment categories. \Y/\N~indicate whether the embodiment fills the
\texttt{base}/\texttt{arm}/\texttt{eef} channel. Humanoid and quadruped base
channels are realized by learned whole-body / locomotion controllers rather than
a wheel model, so the stack above the channel cannot tell legs from wheels
(Section~\ref{sec:control-low}).}
\label{tab:app-robot-list}
\end{table*}

\paragraph{Action terms per channel.}
The channel interface is filled from the shared action-term library of
Section~\ref{sec:control-low}. Arm channels are driven by direct joint mapping
(\texttt{joint\_pos}), Jacobian differential IK, cuRobo batched IK, or Pink IK;
end-effector channels by binary or continuous gripper terms, or by dexterous-hand
joint terms; and base channels by differential, Ackermann, or holonomic drive
models for wheeled bases, or by learned controllers for the quadruped
(\texttt{go2} locomotion policy) and the G1 humanoid (whole-body controllers such
as \texttt{homie\_v2}, the SONIC upper-body policy, and an Agile lower-body
variant).

\paragraph{Non-robot embodiments.}
Two movers are not robots (Section~\ref{sec:flying-camera}). Animation-driven
\emph{avatars} are human entities controlled by skeletal animation rather than
the robot controller stack, providing human-scene context and interaction targets
(Section~\ref{sec:engine-supported-families}). The \emph{flying camera} is a
camera-only embodiment planned through the same step interface as
\texttt{PlannerManager}, supporting eased-linear GoTo motion and gimbal
orbit/track modes for free-viewpoint and multi-view capture. Both are exposed
through the same runtime contract as robot embodiments but do not occupy the
robot action space.

\section{Task List}
\label{app:task-list}

This appendix enumerates the tasks registered under the task and benchmark layer
of Section~\ref{sec:tasks-benchmarks}. Tasks live under
\texttt{src/magicsim/Task/}, organized by task family; each family has an
\texttt{Env/} directory of Gym-registered environment classes and a
\texttt{Conf/} directory of per-task YAML files. Every concrete task instantiates
the same task contract (Section~\ref{sec:task-contract}): it overrides the MDP
hooks \texttt{get\_obs\_space}, \texttt{get\_policy\_obs},
\texttt{get\_privilege\_obs}, \texttt{get\_reward}, and \texttt{get\_termination},
inherits its action space from the embodiment through \texttt{RobotManager}, and
reports the uniform status ontology
(\texttt{running}/\texttt{success}/\texttt{truncated}/\texttt{failed}).

MagicSim registers eight task families---TableTop, MobileManip, LocoManip,
Dexterous, Garment, ContactRich, Camera, and HRI---spanning more than forty
concrete tasks (Section~\ref{sec:task-diversity}). Families organize registration
and shared substrate, while interaction regimes (rigid prehensile, articulated,
contact-rich, deformable, locomotion/navigation, dexterous, viewpoint control,
avatar-conditioned) cut across them. Table~\ref{tab:app-task-list} lists the
registered tasks per family; the count excludes the family base environments
(\texttt{TableTopEnv}, \texttt{ContactRichEnv}), which supply defaults rather than
concrete tasks.

\begin{table*}[htbp]
\centering
\footnotesize
\setlength{\tabcolsep}{4pt}
\renewcommand{\arraystretch}{1.1}
\begin{tabularx}{\textwidth}{>{\ttfamily}p{0.20\textwidth} p{0.24\textwidth} X}
\toprule
Registered task & Interaction regime & Success predicate (informal) \\
\midrule
\multicolumn{3}{l}{\textbf{TableTop} --- fixed-base single/dual-arm; base env \texttt{TableTopEnv}}\\
\midrule
ReachEnv        & point reaching            & end-effector reaches the target pose within tolerance \\
DualReachEnv    & bimanual reaching         & both end-effectors reach their targets \\
GraspEnv        & rigid prehensile grasp    & object lifted and held near the gripper \\
BiGraspEnv      & bimanual grasp            & single object jointly grasped by two arms \\
PickPlaceEnv    & pick-and-place            & object placed on the target support within tolerance \\
PickDropEnv     & pick-and-drop             & object released into the target region \\
PourEnv         & fluid/pour manipulation   & container tilted to transfer contents to the target \\
XtrainerPickPlaceEnv & dual-arm pick-place  & object placed on a support dock (Xtrainer) \\
HandoverEnv     & object handover           & object passed between right and left grippers \\
WaveEnv         & gesture                   & end-effector executes the waving pattern \\
PushEnv         & non-prehensile push       & object pushed to the target location \\
InsertEnv       & contact-rich insertion    & peg inserted with contact feedback \\
OpenDrawerEnv   & articulated (open)         & drawer/door opened to the target joint angle \\
CloseDrawerEnv  & articulated (close)        & drawer/door closed to zero angle \\
RandomReachEnv / RandomPushEnv & randomized variants & randomized reach / push registrations \\
\midrule
\multicolumn{3}{l}{\textbf{MobileManip} --- wheeled base + arm(s)}\\
\midrule
MobileReachEnv     & mobile reaching        & end-effector reaches target via base+arm coordination \\
MobileGraspEnv     & mobile grasp           & object lifted after navigation \\
MobileDualReachEnv & mobile bimanual reach  & both arms reach targets via base motion \\
MobileDexGraspEnv  & mobile dexterous grasp & dexterous hand grasps after navigation \\
MobileOpenDrawerEnv  & mobile articulation (open)  & drawer opened via base+arm \\
MobileCloseDrawerEnv & mobile articulation (close) & drawer closed via base+arm \\
\midrule
\multicolumn{3}{l}{\textbf{LocoManip} --- legged/humanoid + arm(s)}\\
\midrule
LocoReachEnv    & loco reaching             & humanoid reaches target via whole-body motion \\
LocoGraspEnv    & loco grasp                & humanoid walks to and grasps the object \\
LocoBiGraspEnv  & loco bimanual grasp       & humanoid grasps object with two arms \\
LocoLiftEnv     & loco lift                 & humanoid grasps and lifts the object \\
LocoBoxEnv      & loco push                 & humanoid pushes a box to the target \\
SquatGraspEnv   & squat + grasp             & humanoid squats to grasp a low object \\
LocoOpenDoorEnv & loco articulation         & humanoid opens a door by its handle \\
LocoNavEnv      & navigation                & humanoid walks to the target location \\
LocoRoomNavEnv  & room navigation           & humanoid navigates a multi-room scene to the goal \\
\midrule
\multicolumn{3}{l}{\textbf{Dexterous} --- arm + dexterous hand}\\
\midrule
DexReachEnv     & dexterous reaching        & dexterous hand reaches the target \\
DexGraspEnv     & dexterous grasp           & object grasped by finger manipulation \\
BiDexGraspEnv   & bimanual dexterous grasp  & two dexterous hands cooperatively grasp \\
DexPickPlaceEnv & dexterous pick-place      & object picked and placed by a dexterous hand \\
DexOpenDrawerEnv & dexterous articulation   & drawer opened by finger manipulation \\
SharpaScrewdriverEnv & tool use (screwdriver) & SharpaWave hand drives a screw \\
SharpaNutEnv    & tool use (nut)            & SharpaWave hand manipulates a nut on a bolt \\
\midrule
\multicolumn{3}{l}{\textbf{Garment} --- single/dual-arm deformable manipulation}\\
\midrule
GarmentFoldEnv  & deformable fold           & garment folded to target shape (tops/pants configs) \\
FlingEnv        & deformable fling          & garment unfolded past a coverage threshold \\
HangGarmentEnv  & deformable hang           & garment hung and stable on a hook \\
\midrule
\multicolumn{3}{l}{\textbf{ContactRich} --- fixed-base arm; base env \texttt{ContactRichEnv} + task configs}\\
\midrule
GearAssembly    & precision assembly        & gears aligned and seated (\texttt{task\_gear\_assembly.yaml}) \\
InsertUSB       & connector insertion       & USB connector fully inserted (\texttt{task\_insert\_usb.yaml}) \\
PegInHole       & peg-in-hole               & peg inserted with force control (\texttt{task\_peg\_in\_hole.yaml}) \\
\midrule
\multicolumn{3}{l}{\textbf{Camera} --- controllable sensor carrier; base env \texttt{TaskCameraBaseEnv}}\\
\midrule
GoToEnv         & viewpoint control         & camera reaches the target pose \\
\midrule
\multicolumn{3}{l}{\textbf{HRI} --- robot manipulation in avatar-populated scenes; base env \texttt{HRIBaseEnv}}\\
\midrule
(avatar-conditioned) & avatar-conditioned contact-rich manipulation & human-facing manipulation outcome conditioned on avatar pose/motion \\
\bottomrule
\end{tabularx}
\caption{Registered tasks by family. Family names organize registration and
shared substrate; interaction regimes cut across families
(Section~\ref{sec:task-diversity}). ContactRich uses a single base environment
parameterized by task configuration files rather than one class per task, and
GarmentFold ships tops and pants configurations over one class. HRI is anchored
by \texttt{HRIBaseEnv} whose substrate contains an animated avatar advanced by
environment-side dynamics; its success predicates may depend on
robot--object--avatar relations without changing the agent-facing MDP contract.}
\label{tab:app-task-list}
\end{table*}

\paragraph{Diversity by composition.}
The registry is best read as an embodiment-by-interaction matrix rather than a
flat list (Section~\ref{sec:task-diversity}). A new benchmark cell is a
composition of the two axes: the embodiment
(Appendix~\ref{app:robot-list}) contributes the action space, while the task
contributes observation, reward, termination, and command semantics. Task logic
ports across families---the same task form can be registered under more than one
family---and a newly registered embodiment extends every compatible cell without
new task code. Task difficulty is a property of the command distribution and
reset distribution declared in the benchmark protocol
(Section~\ref{sec:task-protocol}), not of separate task implementations.

\section{AtomicSkill and GlobalPlanner Primitive List}
\label{app:skill-list}

This appendix enumerates the executable action vocabulary of
Section~\ref{sec:atomicskills}: the AtomicSkills, the GlobalPlanner primitives
they lower into, the command types that select them, the typed status codes, and
the retry ladder. These are implemented under
\texttt{src/magicsim/Collect/}: skills in \texttt{AtomicSkill/}, primitives in
\texttt{GlobalPlanner/}, and command types in \texttt{Command/}. An AtomicSkill
has two faces (Section~\ref{sec:skill-two-faces}): upward it is an action symbol
selected by a scripted command source or a high-level policy; downward it is a
traceable execution unit whose lifecycle states are owned by the skill executor
(\texttt{AtomicSkillManager}), not by the policy-facing action API.

\subsection{AtomicSkills}
\label{app:skill-atomicskills}

The executor registers \textbf{24 AtomicSkills}
(\texttt{AtomicSkillManager.create\_atomic\_skill}). Most are planner-grounded
online goal compilers (Section~\ref{sec:skill-library}): they store no reference
trajectory but compile task intent and live scene state into planner goals,
success predicates, and recovery attempts. Others are realized by learned
state-based RL policies (high-DOF dexterous behaviors), by a model-predictive
backend (contact-rich continuous correction), or by deterministic scripts, while
exposing the same lifecycle, typed outcomes, and recording boundaries
(Section~\ref{sec:skill-backends}).

\begin{table*}[htbp]
\centering
\footnotesize
\setlength{\tabcolsep}{4pt}
\renewcommand{\arraystretch}{1.1}
\begin{tabularx}{\textwidth}{>{\ttfamily}p{0.15\textwidth} X p{0.30\textwidth}}
\toprule
AtomicSkill & Function & Primary primitives \\
\midrule
Reach        & move an end-effector to a target pose              & MoveL \\
RetractReach & reach with the arm held in a retracted configuration & RetractMoveL \\
MobileReach  & reach combining base navigation and arm motion    & MobileMoveL, NavTo \\
Grasp        & single-arm parallel-gripper grasp (pre/grasp/close/retreat) & MoveL, ParallelGripper \\
BiGrasp      & bimanual grasp with paired reachable candidates   & MoveL, ParallelGripper \\
DexGrasp     & dexterous-hand grasp                              & MoveL, DexHand \\
BiDexGrasp   & bimanual dexterous grasp                          & MoveL, DexHand \\
Place        & place a held object onto a validated support       & MoveL, ParallelGripper \\
Drop         & release a held object without a placement target   & MoveL, ParallelGripper \\
Lift         & vertical lift of a grasped object                 & MoveL \\
Push         & non-prehensile push of an object                  & MoveL (MPC-corrected) \\
Insert       & peg/connector insertion with lateral servo         & MoveL, ServoL \\
Pour         & tilt-and-pour from a held container                & MoveL, ServoL \\
Handover     & pass an object between the two arms                & MoveL, ParallelGripper \\
Wave         & bimanual waving / gesture                          & MoveL \\
OpenDrawer   & open an articulated part along a validated trajectory & MoveL, ServoL \\
CloseDrawer  & close an articulated part                          & MoveL, ServoL \\
DexOpenDrawer & open an articulated part with a dexterous hand    & MoveL, ServoL, DexHand \\
Dehatch      & release/open a latched hatch mechanism             & MoveL \\
Fold         & bimanual cloth folding                             & MoveL \\
Fling        & bimanual garment fling to unfold                   & MoveL, ParallelGripper \\
NavTo        & compute and drive a mobile base to a goal pose     & NavTo \\
LocoOpenDoor & locomotion-plus-arm door opening                  & MobileMoveL, ServoL \\
LocoBox      & locomotion-plus-manipulation box transport        & MobileMoveL, ParallelGripper \\
\bottomrule
\end{tabularx}
\caption{The 24 registered AtomicSkills and the GlobalPlanner primitives they
compile into. Motion realization is restricted to the eight primitives of
Table~\ref{tab:app-primitives}; skills own target construction, phase gates,
contact and object-state predicates, and terminal status.}
\label{tab:app-atomicskills}
\end{table*}

\subsection{GlobalPlanner Primitives}
\label{app:skill-primitives}

Skills realize motion through \textbf{eight GlobalPlanner primitives}
(\texttt{GlobalPlannerManager.create\_global\_planner}). Each primitive parses a
normalized request header \texttt{((robot\_id, hand\_id, planner\_mode), target)}
and emits a per-tick action over the \texttt{base}/\texttt{arm}/\texttt{eef}
channels. Arm-side primitives submit to the asynchronous solve-farm
(\S\ref{app:skill-services}); gripper and hand primitives issue direct
end-effector commands.

\begin{table}[htbp]
\centering
\small
\setlength{\tabcolsep}{4pt}
\begin{tabularx}{\linewidth}{>{\ttfamily}p{0.20\linewidth} X}
\toprule
Primitive & Target and behavior \\
\midrule
MoveL         & target end-effector pose(s); mode-gated snap / motion-generation / linear interpolation to the goal. \\
ServoL        & a Cartesian trajectory tracked in closed loop with pre-solved IK. \\
MobileMoveL   & arm target with a mobile base (locked or free through virtual planar joints). \\
MobileServoL  & mobile-base trajectory tracking with arm servo. \\
RetractMoveL  & mobile-base move with the arm held in a retracted configuration. \\
NavTo         & a base goal pose; navmesh path following. \\
ParallelGripper & a scalar gripper command (open/close). \\
DexHand       & a dexterous-hand joint-target vector. \\
\bottomrule
\end{tabularx}
\caption{The eight GlobalPlanner primitives. Inactive arms in a multi-frame
request are marked with NaN rows, which the solver flips to a disabled tool-frame
criterion for that solve only (Section~\ref{sec:curobo-capabilities}).}
\label{tab:app-primitives}
\end{table}

\subsection{Command Types}
\label{app:skill-commands}

The AutoCollect driver samples \textbf{37 command types}
(\texttt{Collect/Command/}) and expands each into one or more AtomicSkills. A
command specifies which command to run, on which object, with which robot, hand,
mode, and target parameters (Section~\ref{sec:skill-two-faces}). Compound
commands chain skills (e.g., \texttt{PickPlace} = Grasp then Place); mobile and
loco variants prepend navigation.

\begin{table*}[htbp]
\centering
\footnotesize
\setlength{\tabcolsep}{4pt}
\renewcommand{\arraystretch}{1.08}
\begin{tabularx}{\textwidth}{p{0.24\textwidth} X}
\toprule
Group & Command types \\
\midrule
Reaching & \texttt{Reach}, \texttt{DualReach}, \texttt{MobileReach}, \texttt{MobileDualReach}, \texttt{LocoReach}, \texttt{DualLocoReach}, \texttt{LocoRetractReach}, \texttt{DualLocoRetractReach} \\
Grasping & \texttt{Grasp}, \texttt{BiGrasp}, \texttt{XtrainerGrasp}, \texttt{MobileGrasp}, \texttt{DexGrasp}, \texttt{BiDexGrasp}, \texttt{LocoDexGrasp}, \texttt{LocoDexBiGrasp}, \texttt{SquatDexGrasp} \\
Pick-and-place & \texttt{PickPlace}, \texttt{PickDrop}, \texttt{Pour}, \texttt{XtrainerPickPlace}, \texttt{DexPickPlace}, \texttt{LocoLift}, \texttt{LocoBox} \\
Articulated objects & \texttt{OpenDrawer}, \texttt{CloseDrawer}, \texttt{DexOpenDrawer}, \texttt{MobileOpenDrawer}, \texttt{MobileCloseDrawer}, \texttt{LocoOpenDoor} \\
Contact / other manip. & \texttt{Insert}, \texttt{Push}, \texttt{Wave}, \texttt{Handover} \\
Deformable & \texttt{Fling}, \texttt{Fold} \\
Navigation & \texttt{NavTo} \\
\bottomrule
\end{tabularx}
\caption{The 37 command types grouped by intent. Roughly twenty-two to
twenty-four skills over eight primitives realize on the order of thirty command
types across the task families and embodiments of Section~\ref{sec:task-layer};
mobile and loco variants reuse the same underlying skills with navigation
prepended.}
\label{tab:app-commands}
\end{table*}

\subsection{Typed Status and Outcomes}
\label{app:skill-status}

Every AtomicSkill carries a lifecycle state, one of \texttt{ready},
\texttt{running}, \texttt{failed}, or \texttt{truncated}
(\texttt{AtomicSkill.current\_state}). On each update the executor propagates a
typed \texttt{truncated} code that distinguishes why a skill stopped, separating
environment-side termination from planner and skill failures.

\begin{table}[htbp]
\centering
\small
\begin{tabularx}{\linewidth}{c X}
\toprule
Code & Meaning \\
\midrule
0 & still running (or re-initializing for a retry) \\
1 & environment terminated first --- task success \\
2 & environment truncated first (e.g., horizon reached) \\
3 & GlobalPlanner failed to plan (IK / motion-generation failure) \\
4 & AtomicSkill failed to plan (no viable target / predicate failure) \\
\bottomrule
\end{tabularx}
\caption{Typed \texttt{truncated} codes emitted by the skill layer. Code~1 is the
write gate for collection; codes~3--4 are collection-time geometric or physical
failures that feed the recovery ladder (\S\ref{app:skill-retry}). The command
layer additionally bounds per-command retries with a \texttt{max\_attempt}
budget.}
\label{tab:app-status}
\end{table}

\subsection{Retry and Recovery Ladder}
\label{app:skill-retry}

AtomicSkill does not remove failure; it makes failure local, typed, and useful
(Section~\ref{sec:skill-failure}). Collection-time failures are mostly geometric
or physical (no IK, collision, unstable grasp, missed contact, timeout,
controller failure, adverse object state) and trigger a fixed ladder:

\begin{enumerate}
\item \textbf{IK goalset selection.} Grasp-like skills pack many candidate poses
  from the annotation bank (Section~\ref{sec:annotation-bank}) into one goalset
  request; the solver selects a reachable low-cost candidate, or reports failure
  if none solve.
\item \textbf{Alternative goalset / part.} If a functional grasp or a chosen part
  fails, the skill falls back to other candidates or parts in the bank.
\item \textbf{Motion-generation fallback.} When a direct pose target is
  infeasible or far, \texttt{MoveL}/\texttt{MobileMoveL} escalate from snap to
  collision-aware motion generation.
\item \textbf{Closed-loop correction.} Contact-rich skills (Insert, Push) apply
  servo or MPC correction as contact evolves rather than replanning from scratch.
\item \textbf{Skill retry.} On a false-positive finish (skill done but the
  environment did not confirm success), the command layer re-initializes and
  re-executes up to a \texttt{max\_attempt} budget.
\end{enumerate}

These typed, recoverable failures are themselves valuable supervision:
failure-recovery data is harder to collect in the real world
(Section~\ref{sec:skill-failure}). Storage is segmented so each successful command
flushes its own trajectory record with recoverable episode, command, skill, and
backend identifiers, preserving verified segments even if a later command fails.

\subsection{Planning Services}
\label{app:skill-services}

Arm-side primitives resolve through four in-process services that share the
asynchronous microbatch solve-farm of Section~\ref{sec:async-solve-farm}:
\texttt{IKServer} and \texttt{DualIKServer} for inverse kinematics (including
goalset candidate selection), and \texttt{MotionGenServer} and
\texttt{DualMotionGenServer} for collision-aware trajectory generation; the
\texttt{Dual} variants handle mobile manipulators under both locked-base and
free-base formulations. A submission returns a future immediately and resolves to
\texttt{(success, goalset\_index, env\_ids)} or \texttt{(actions, success,
env\_ids)}; the caller keeps stepping the batch and harvests the result when
ready (Section~\ref{sec:collection-async}). Ragged per-environment requests are
deduplicated, padded, grouped by goalset width, and fused into fixed-shape
batched solves, so per-agent planning is aggregated rather than issued as many
isolated calls.


\section{Extended Related Work}
\label{app:extended_related}

\subsection{Robotics and physics simulators}
Broader simulator literature spans GPU-native stacks, control-oriented engines, articulated-scene simulators, and manipulation-focused frameworks. Isaac Gym and Isaac Lab prioritize throughput and multimodal robot learning; MuJoCo remains a standard engine for dynamics-centric control research; SAPIEN and ManiSkill emphasize articulated objects and contact-rich manipulation; and robosuite offers a modular MuJoCo-based task framework \cite{makoviychuk2021isaacgym,isaacLab2025,todorov2012mujoco,xiang2020sapien,maniskill2_2023,robosuite2020}. MagicSim should be compared to this layer as middleware above the simulator, not as a replacement physics engine.

\subsection{Embodied AI platforms and interactive environments}
Embodied platforms differ in whether they emphasize navigation, household interaction, program execution, or rearrangement. Habitat and Habitat 2.0 focus on embodied navigation and rearrangement; AI2-THOR and ProcTHOR center interactive household environments and large-scale procedural generation; VirtualHome represents activities as executable programs; iGibson and the BEHAVIOR family push toward realistic household scenes with richer semantics; and the ThreeDWorld Transport Challenge explicitly couples perception with task-and-motion planning \cite{habitat2019,habitat2_2021,ai2thor2017,procTHOR2022,virtualhome2018,igibson2020,behavior2021,li2024behavior1k,tdwTransport2021}. MagicSim overlaps with this family in interactivity, but the paper should stress reusable simulation/data runtime structure rather than a new platform-only benchmark.

\subsection{Robot manipulation benchmarks}
Manipulation benchmarks vary in embodiment diversity, task semantics, demonstration availability, and evaluation protocol. RLBench and Meta-World are widely used for multi-task manipulation; robosuite provides modular tasks and reproducible interfaces; ManiSkill3 scales GPU-parallel simulation for broad embodied robotics; CALVIN and LIBERO emphasize long-horizon or transfer-oriented manipulation; RoboCasa focuses on realistic everyday household tasks; and ManiSkill-HAB connects low-level manipulation to home rearrangement \cite{rlbench2019,metaWorld2019,robosuite2020,tao2025maniskill3,calvin2021,libero2023,robocasa2024,maniskillHab2024}. In the supplement, MagicSim can be positioned as system-centric infrastructure that can host benchmark tasks, rather than as only another manipulation benchmark.

\subsection{Synthetic data and demonstration generation}
Large robot datasets and data-generation systems now form a distinct line of work. Open X-Embodiment standardizes cross-robot data; DROID and BridgeData V2 show the value of broad real-robot corpora; MimicGen expands demonstrations from a small human seed set; RoboMimic provides a common offline-learning substrate; BC-Z studies zero-shot generalization from broad imitation data; VIMA uses simulation-scale expert trajectories; and RoboCasa shows how realistic household simulation can feed large synthetic datasets \cite{openX2023,droid2024,bridgedata2023,mandlekar2023mimicgen,robomimic2021,bcz2022,vima2022,robocasa2024}. MagicSim should be framed here not primarily as a dataset paper, but as a generator of replayable, success-gated, multimodal trajectories.

\subsection{Motion planning, TAMP, and skill-based control}
Planning literature provides the algorithmic and software basis for executing structured robot behavior. OMPL and MoveIt offer general motion-planning infrastructure; hierarchical task-and-motion planning formalizes the coupling of symbolic and continuous reasoning; recent surveys organize optimization-based TAMP; and cuRobo demonstrates GPU-accelerated collision-aware motion generation \cite{ompl2012,moveit2014,kaelbling2011,tampSurvey2024,curobo2023}. The supplementary comparison should make clear that MagicSim's contribution is not a new planning algorithm per se, but a runtime architecture in which planner queries, skill execution, and simulator state share one system interface.

\subsection{Multimodal annotation and synthetic perception data}
Synthetic-data systems often focus on rendering and dense computer-vision annotations rather than robot-executable task traces. BlenderProc and Kubric are general-purpose pipelines for scalable synthetic image generation; GRADE uses Isaac Sim to create repeatable, richly annotated dynamic scenes; and BEHAVIOR-1K demonstrates how large embodied environments can expose rich semantic and physical structure \cite{blenderproc2019,kubric2022,grade2023,li2024behavior1k}. MagicSim's relation to this literature is that it extends beyond generic RGB-D, segmentation, or geometry labels toward robot-learning-specific annotations attached to executable trajectories and task state.

\subsection{Language, VLM, VLA, and embodied agents}
Embodied language and foundation-policy work spans instruction-following benchmarks, LLM-planned robotics, embodied multimodal models, and open-source VLAs. ALFRED and TEACh study instruction following and dialogue in simulated homes; SayCan and Inner Monologue ground language in skill feedback; PaLM-E and RT-2 connect large multimodal models to embodied control; OpenVLA and Octo provide open policy baselines; and MineDojo and Voyager frame long-horizon agentic interaction in open-ended worlds \cite{alfred2019,teach2021,saycan2022,innerMonologue2022,palme2023,rt2_2023,openvla2024,octo2024,minedojo2022,voyager2023}. MagicSim should be positioned as simulator-side infrastructure for such agents, not as a replacement for their policy architectures.

\subsection{Reproducibility, determinism, and simulation state replay}
Many simulation systems support seeding, procedural generation, or reproducible task instantiation, but the granularity of reproducibility differs substantially. Habitat 2.0, ProcTHOR, and the BEHAVIOR line support controlled generation of embodied tasks and layouts; GRADE explicitly studies repeatable playback of recorded experiments inside simulation; and robosuite emphasizes reproducible learning environments and unified interfaces \cite{habitat2_2021,procTHOR2022,behavior2021,li2024behavior1k,grade2023,robosuite2020}. This subsection is the right place to sharpen MagicSim's distinction between simple seed control and manager-level snapshot/reset/replay semantics.

\subsection{Serving and distributed simulation}
Parallel and distributed simulation systems focus on throughput and learner--simulator coupling. Isaac Gym and Isaac Lab emphasize batched GPU simulation; ManiSkill3 similarly targets large-scale GPU-parallel robotics; Habitat and Galactic demonstrate fast embodied training at scale; and RLlib Flow formalizes distributed RL as a dataflow problem \cite{makoviychuk2021isaacgym,isaacLab2025,habitat2019,tao2025maniskill3,galactic2023,rllib2020}. MagicSim can be positioned here as moving from vectorized simulation toward simulator-as-a-service, where the same runtime can back local training, scripted collection, replay, and remote clients.

\bibliographystyle{unsrtnat}
\bibliography{paper}

@inproceedings{todorov2012mujoco,
  title = {{MuJoCo}: A physics engine for model-based control},
  author = {Todorov, Emanuel and Erez, Tom and Tassa, Yuval},
  booktitle = {IEEE/RSJ International Conference on Intelligent Robots and Systems},
  year = {2012},
  
}

@misc{ning2023where2explorefewshotaffordancelearning,
      title={Where2Explore: Few-shot Affordance Learning for Unseen Novel Categories of Articulated Objects}, 
      author={Chuanruo Ning and Ruihai Wu and Haoran Lu and Kaichun Mo and Hao Dong},
      year={2023},
      eprint={2309.07473},
      archivePrefix={arXiv},
      primaryClass={cs.RO},
      url={https://arxiv.org/abs/2309.07473}, 
}

@misc{li2023imagemanipimagebasedroboticmanipulation,
      title={ImageManip: Image-based Robotic Manipulation with Affordance-guided Next View Selection}, 
      author={Xiaoqi Li and Yanzi Wang and Yan Shen and Ponomarenko Iaroslav and Haoran Lu and Qianxu Wang and Boshi An and Jiaming Liu and Hao Dong},
      year={2023},
      eprint={2310.09069},
      archivePrefix={arXiv},
      primaryClass={cs.RO},
      url={https://arxiv.org/abs/2310.09069}, 
}

@article{makoviychuk2021isaacgym,
  title = {{Isaac Gym}: High Performance {GPU}-Based Physics Simulation for Robot Learning},
  author = {Makoviychuk, Viktor and Wawrzyniak, Lukasz and Guo, Yunrong and Lu, Michelle and Storey, Kier and Macklin, Miles and Hoeller, David and Rudin, Nikita and Allshire, Arthur and Handa, Ankur and State, Gavriel},
  journal = {arXiv preprint arXiv:2108.10470},
  year = {2021}
}

@inproceedings{xiang2020sapien,
  title = {{SAPIEN}: A SimulAted Part-based Interactive ENvironment},
  author = {Xiang, Fanbo and Qin, Yuzhe and Mo, Kaichun and Xia, Yikuan and Zhu, Hao and Liu, Fangchen and Liu, Minghua and Jiang, Hanxiao and Yuan, Yifu and Wang, He and Yi, Li and Chang, Angel X. and Guibas, Leonidas J. and Su, Hao},
  booktitle = {IEEE/CVF Conference on Computer Vision and Pattern Recognition},
  year = {2020}
}

@misc{shen2025biassemblelearningcollaborativeaffordance,
      title={BiAssemble: Learning Collaborative Affordance for Bimanual Geometric Assembly}, 
      author={Yan Shen and Ruihai Wu and Yubin Ke and Xinyuan Song and Zeyi Li and Xiaoqi Li and Hongwei Fan and Haoran Lu and Hao dong},
      year={2025},
      eprint={2506.06221},
      archivePrefix={arXiv},
      primaryClass={cs.RO},
      url={https://arxiv.org/abs/2506.06221}, 
}

@misc{Intelligence2026pi07AS,
  author = {{Physical Intelligence}},
  title = {Pi-0.7: A Steerable Generalist Robotic Foundation Model with Emergent Capabilities},
  year = {2026},

  howpublished = {arXiv preprint},
  note = {CorpusID: 287607456}
}

@misc{generalist2025gen0,
  author = {{Generalist AI Team}},
  title = {GEN-0: Embodied Foundation Models That Scale with Physical Interaction},
  year = {2025},
  howpublished = {Generalist AI Blog},
  note = {November 4, 2025}
}

@misc{Nvidia2025GR00TNA,
  author = {{NVIDIA}},
  title = {GR00T N1: An Open Foundation Model for Generalist Humanoid Robots},
  year = {2025},
  howpublished = {arXiv:2503.14734}
}

@misc{Ye2026WorldAM,
  author = {Ye, Seonghyeon and Ge, Yunhao and Zheng, Kaiyuan and Jang, Joel},
  title = {World Action Models are Zero-shot Policies},
  year = {2026},
  howpublished = {arXiv:2602.15922}
}

@article{Intelligence202505AV,
  title={$\pi$0.5: a Vision-Language-Action Model with Open-World Generalization},
  author={Physical Intelligence and Kevin Black and Noah Brown and James Darpinian and Karan Dhabalia and Danny Driess and Adnan Esmail and Michael Equi and Chelsea Finn and Niccolo Fusai and Manuel Y. Galliker and Dibya Ghosh and Lachy Groom and Karol Hausman and Brian Ichter and Szymon Jakubczak and Tim Jones and Liyiming Ke and Devin LeBlanc and Sergey Levine and Adrian Li-Bell and Mohith Mothukuri and Suraj Nair and Karl Pertsch and Allen Z. Ren and Lucy Xiaoyang Shi and Laura Smith and Jost Tobias Springenberg and Kyle Stachowicz and James Tanner and Quan Vuong and Homer Rich Walke and Anna Walling and Haohuan Wang and Lili Yu and Ury Zhilinsky},
  journal={ArXiv},
  year={2025},
  volume={abs/2504.16054},
  url={https://api.semanticscholar.org/CorpusID:277993634}
}

@article{Li2024BroadcastingSR,
  title={Broadcasting Support Relations Recursively from Local Dynamics for Object Retrieval in Clutters},
  author={Yitong Li and Ruihai Wu and Haoran Lu and Chuanruo Ning and Yan Shen and Guanqi Zhan and Hao Dong},
  journal={ArXiv},
  year={2024},
  volume={abs/2406.02283},
  url={https://api.semanticscholar.org/CorpusID:270226492}
}

@article{Wu2024UniGarmentManipAU,
  title={UniGarmentManip: A Unified Framework for Category-Level Garment Manipulation via Dense Visual Correspondence},
  author={Ruihai Wu and Haoran Lu and Yiyan Wang and Yubo Wang and Hao Dong},
  journal={2024 IEEE/CVF Conference on Computer Vision and Pattern Recognition (CVPR)},
  year={2024},
  pages={16340-16350},
  url={https://api.semanticscholar.org/CorpusID:269757227}
}

@article{Shen2025BiAssembleLC,
  title={BiAssemble: Learning Collaborative Affordance for Bimanual Geometric Assembly},
  author={Yan Shen and Ruihai Wu and Yubin Ke and Xinyuan Song and Zeyi Li and Xiaoqi Li and Hongwei Fan and Haoran Lu and Hao Dong},
  journal={ArXiv},
  year={2025},
  volume={abs/2506.06221},
  url={https://api.semanticscholar.org/CorpusID:279244917}
}

@INPROCEEDINGS{11128651,
  author={Wu, Ruihai and Chen, Haozhe and Zhang, Mingtong and Lu, Haoran and Li, Yitong and Li, Yunzhu},
  booktitle={2025 IEEE International Conference on Robotics and Automation (ICRA)}, 
  title={Neural Dynamics Augmented Diffusion Policy}, 
  year={2025},
  volume={},
  number={},
  pages={13234-13241},
  doi={10.1109/ICRA55743.2025.11128651}}

@misc{chen2026learningpartawaredense3d,
      title={Learning Part-Aware Dense 3D Feature Field for Generalizable Articulated Object Manipulation}, 
      author={Yue Chen and Muqing Jiang and Kaifeng Zheng and Jiaqi Liang and Chenrui Tie and Haoran Lu and Ruihai Wu and Hao Dong},
      year={2026},
      eprint={2602.14193},
      archivePrefix={arXiv},
      primaryClass={cs.RO},
      url={https://arxiv.org/abs/2602.14193}, 
}

@misc{yuan2026fastwamworldactionmodels,
      title={Fast-WAM: Do World Action Models Need Test-time Future Imagination?}, 
      author={Tianyuan Yuan and Zibin Dong and Yicheng Liu and Hang Zhao},
      year={2026},
      eprint={2603.16666},
      archivePrefix={arXiv},
      primaryClass={cs.CV},
      url={https://arxiv.org/abs/2603.16666}, 
}

@article{Ye2025LearningTF,
  title={Learning to Feel the Future: DreamTacVLA for Contact-Rich Manipulation},
  author={Guo Ye and Zexi Zhang and Xu Zhao and Shang Wu and Haoran Lu and Shihan Lu and Han Liu},
  journal={ArXiv},
  year={2025},
  volume={abs/2512.23864},
  url={https://api.semanticscholar.org/CorpusID:284350273}
}

@article{Wang2025VAGENRW,
  title={VAGEN: Reinforcing World Model Reasoning for Multi-Turn VLM Agents},
  author={Kangrui Wang and Pingyue Zhang and Zihan Wang and Yaning Gao and Linjie Li and Qineng Wang and Hanyang Chen and Chi Wan and Yiping Lu and Zhengyuan Yang and Lijuan Wang and Ranjay Krishna and Jiajun Wu and Fei-Fei Li and Yejin Choi and Manling Li},
  journal={ArXiv},
  year={2025},
  volume={abs/2510.16907},
  url={https://api.semanticscholar.org/CorpusID:282210682}
}

@article{Wang2025RAGENUS,
  title={RAGEN: Understanding Self-Evolution in LLM Agents via Multi-Turn Reinforcement Learning},
  author={Zihan Wang and Kangrui Wang and Qineng Wang and Pingyue Zhang and Linjie Li and Zhengyuan Yang and Kefan Yu and Minh Nhat Nguyen and Licheng Liu and Eli Gottlieb and Monica S. Lam and Yiping Lu and Kyunghyun Cho and Jiajun Wu and Fei-Fei Li and Lijuan Wang and Yejin Choi and Manling Li},
  journal={ArXiv},
  year={2025},
  volume={abs/2504.20073},
  url={https://api.semanticscholar.org/CorpusID:278170861}
}

@article{Fung2025EmbodiedAA,
  title={Embodied AI Agents: Modeling the World},
  author={Pascale Fung and Yoram Bachrach and Asli Celikyilmaz and Kamalika Chaudhuri and Delong Chen and Willy Chung and Emmanuel Dupoux and Herv{\'e} J{\'e}gou and Alessandro Lazaric and Arjun Majumdar and Andrea Madotto and Franziska Meier and Florian Metze and Th{\'e}o Moutakanni and Juan Pino and Basile Terver and Joseph Tighe and Jitendra Malik},
  journal={ArXiv},
  year={2025},
  volume={abs/2506.22355},
  url={https://api.semanticscholar.org/CorpusID:280010887}
}

@article{isaacLab2025,
  title={Isaac Lab: A GPU-Accelerated Simulation Framework for Multi-Modal Robot Learning},
  author={Mittal, Mayank and Roth, Pascal and Tigue, James and Richard, Antoine and Zhang, Octi and Du, Peter and others},
  journal={arXiv preprint arXiv:2511.04831},
  year={2025}
}

@article{maniskill2_2023,
  title={ManiSkill2: A Unified Benchmark for Generalizable Manipulation Skills},
  author={Gu, Jiayuan and Xiang, Fanbo and Li, Xuanlin and Ling, Zhan and Liu, Xiqiang and Mu, Tongzhou and Tang, Yihe and Tao, Stone and others},
  journal={arXiv preprint arXiv:2302.04659},
  year={2023}
}

@article{habitat2019,
  title={Habitat: A Platform for Embodied AI Research},
  author={Savva, Manolis and Kadian, Abhishek and Maksymets, Oleksandr and Zhao, Yili and Wijmans, Erik and Jain, Bhavana and Straub, Julian and Liu, Jia and Koltun, Vladlen and Malik, Jitendra and Parikh, Devi and Batra, Dhruv},
  journal={arXiv preprint arXiv:1904.01201},
  year={2019}
}

@article{habitat2_2021,
  title={Habitat 2.0: Training Home Assistants to Rearrange their Habitat},
  author={Szot, Andrew and Clegg, Alex and Undersander, Eric and Wijmans, Erik and Zhao, Yili and Turner, John and Maestre, Noah and Mukadam, Mustafa and Chaplot, Devendra Singh and Maksymets, Oleksandr and others},
  journal={arXiv preprint arXiv:2106.14405},
  year={2021}
}

@article{igibson2020,
  title={iGibson 1.0: A Simulation Environment for Interactive Tasks in Large Realistic Scenes},
  author={Shen, Bokui and Xia, Fei and Li, Chengshu and Mart{\'\i}n-Mart{\'\i}n, Roberto and Fan, Linxi and Wang, Guanzhi and P{\'e}rez-D'Arpino, Claudia and Buch, Shyamal and Srivastava, Sanjana and Tchapmi, Lyne P. and Tchapmi, Micael E. and Vainio, Kent and Wong, Josiah and Fei-Fei, Li and Savarese, Silvio},
  journal={arXiv preprint arXiv:2012.02924},
  year={2020}
}

@article{behavior2021,
  title={{BEHAVIOR}: Benchmark for Everyday Household Activities in Virtual, Interactive, and Ecological Environments},
  author={Srivastava, Sanjana and Li, Chengshu and Lingelbach, Michael and Mart{\'\i}n-Mart{\'\i}n, Roberto and Xia, Fei and Vainio, Kent and Lian, Zheng and Gokmen, Cem and Buch, Shyamal and Liu, C. Karen and Savarese, Silvio and Gweon, Hyowon and Wu, Jiajun and Fei-Fei, Li},
  journal={arXiv preprint arXiv:2108.03332},
  year={2021}
}

@article{ai2thor2017,
  title={{AI2-THOR}: An Interactive 3D Environment for Visual AI},
  author={Kolve, Eric and Mottaghi, Roozbeh and Han, Winson and VanderBilt, Eli and Weihs, Luca and Herrasti, Alvaro and Deitke, Matt and Ehsani, Kiana and Gordon, Daniel and Zhu, Yuke and Kembhavi, Aniruddha and Gupta, Abhinav and Farhadi, Ali},
  journal={arXiv preprint arXiv:1712.05474},
  year={2017}
}

@article{procTHOR2022,
  title={ProcTHOR: Large-Scale Embodied AI Using Procedural Generation},
  author={Deitke, Matt and VanderBilt, Eli and Herrasti, Alvaro and Weihs, Luca and Salvador, Jordi and Ehsani, Kiana and Han, Winson and Kolve, Eric and Farhadi, Ali and Kembhavi, Aniruddha and Mottaghi, Roozbeh},
  journal={arXiv preprint arXiv:2206.06994},
  year={2022}
}

@article{virtualhome2018,
  title={VirtualHome: Simulating Household Activities via Programs},
  author={Puig, Xavier and Ra, Kevin and Boben, Marko and Li, Jiaman and Wang, Tingwu and Fidler, Sanja and Torralba, Antonio},
  journal={arXiv preprint arXiv:1806.07011},
  year={2018}
}

@article{tdwTransport2021,
  title={The ThreeDWorld Transport Challenge: A Visually Guided Task-and-Motion Planning Benchmark for Physically Realistic Embodied AI},
  author={Gan, Chuang and Zhou, Siyuan and Schwartz, Jeremy and Alter, Seth and Bhandwaldar, Abhishek and Gutfreund, Dan and Yamins, Daniel L. K. and DiCarlo, James J. and McDermott, Josh and Torralba, Antonio and Tenenbaum, Joshua B.},
  journal={arXiv preprint arXiv:2103.14025},
  year={2021}
}

@article{rlbench2019,
  title={{RLBench}: The Robot Learning Benchmark \& Learning Environment},
  author={James, Stephen and Ma, Zicong and Rovick Arrojo, David and Davison, Andrew J.},
  journal={arXiv preprint arXiv:1909.12271},
  year={2019}
}

@article{calvin2021,
  title={{CALVIN}: A Benchmark for Language-Conditioned Policy Learning for Long-Horizon Robot Manipulation Tasks},
  author={Mees, Oier and Hermann, Lukas and Rosete-Beas, Erick and Burgard, Wolfram},
  journal={arXiv preprint arXiv:2112.03227},
  year={2021}
}

@article{robocasa2024,
  title={{RoboCasa}: Large-Scale Simulation of Everyday Tasks for Generalist Robots},
  author={Nasiriany, Soroush and Maddukuri, Abhiram and Zhang, Lance and Parikh, Adeet and Lo, Aaron and Joshi, Abhishek and Mandlekar, Ajay and Zhu, Yuke},
  journal={arXiv preprint arXiv:2406.02523},
  year={2024}
}

@article{metaWorld2019,
  title={{Meta-World}: A Benchmark and Evaluation for Multi-Task and Meta Reinforcement Learning},
  author={Yu, Tianhe and Quillen, Deirdre and He, Zhanpeng and Julian, Ryan and Narayan, Avnish and Shively, Hayden and Bellathur, Adithya and Hausman, Karol and Finn, Chelsea and Levine, Sergey},
  journal={arXiv preprint arXiv:1910.10897},
  year={2019}
}

@article{robosuite2020,
  title={robosuite: A Modular Simulation Framework and Benchmark for Robot Learning},
  author={Zhu, Yuke and Wong, Josiah and Mandlekar, Ajay and Mart{\'\i}n-Mart{\'\i}n, Roberto and Joshi, Abhishek and Lin, Kevin and Maddukuri, Abhiram and Nasiriany, Soroush and Zhu, Yifeng},
  journal={arXiv preprint arXiv:2009.12293},
  year={2020}
}

@article{libero2023,
  title={{LIBERO}: Benchmarking Knowledge Transfer for Lifelong Robot Learning},
  author={Liu, Bo and Zhu, Yifeng and Gao, Chongkai and Feng, Yihao and Liu, Qiang and Zhu, Yuke and Stone, Peter},
  journal={arXiv preprint arXiv:2306.03310},
  year={2023}
}

@article{maniskillHab2024,
  title={{ManiSkill-HAB}: A Benchmark for Low-Level Manipulation in Home Rearrangement Tasks},
  author={Shukla, Arth and Tao, Stone and Su, Hao},
  journal={arXiv preprint arXiv:2412.13211},
  year={2024}
}

@article{openX2023,
  title={Open X-Embodiment: Robotic Learning Datasets and {RT-X} Models},
  author={{Open X-Embodiment Collaboration}},
  journal={arXiv preprint arXiv:2310.08864},
  year={2023}
}

@article{vima2022,
  title={VIMA: General Robot Manipulation with Multimodal Prompts},
  author={Jiang, Yunfan and Gupta, Agrim and Zhang, Zichen and Wang, Guanzhi and Dou, Yongqiang and Chen, Yanjun and Fei-Fei, Li and Anandkumar, Anima and Zhu, Yuke and Fan, Linxi},
  journal={arXiv preprint arXiv:2210.03094},
  year={2022}
}

@article{droid2024,
  title={{DROID}: A Large-Scale In-The-Wild Robot Manipulation Dataset},
  author={Khazatsky, Alexander and Pertsch, Karl and Nair, Suraj and Balakrishna, Ashwin and Dasari, Sudeep and Karamcheti, Siddharth and Nasiriany, Soroush and Srirama, Mohan Kumar and others},
  journal={arXiv preprint arXiv:2403.12945},
  year={2024}
}

@article{bridgedata2023,
  title={BridgeData V2: A Dataset for Robot Learning at Scale},
  author={Walke, Homer and Black, Kevin and Lee, Abraham and Kim, Moo Jin and Du, Max and Zheng, Chongyi and Zhao, Tony and Hansen-Estruch, Philippe and Vuong, Quan and He, Andre and Myers, Vivek and Fang, Kuan and Finn, Chelsea and Levine, Sergey},
  journal={arXiv preprint arXiv:2308.12952},
  year={2023}
}

@article{robomimic2021,
  title={What Matters in Learning from Offline Human Demonstrations for Robot Manipulation},
  author={Mandlekar, Ajay and Xu, Danfei and Wong, Josiah and Nasiriany, Soroush and Wang, Chen and Kulkarni, Rohun and Fei-Fei, Li and Savarese, Silvio and Zhu, Yuke and Mart{\'\i}n-Mart{\'\i}n, Roberto},
  journal={arXiv preprint arXiv:2108.03298},
  year={2021}
}

@article{bcz2022,
  title={{BC-Z}: Zero-Shot Task Generalization with Robotic Imitation Learning},
  author={Jang, Eric and Irpan, Alex and Khansari, Mohi and Kappler, Daniel and Ebert, Frederik and Lynch, Corey and Levine, Sergey and Finn, Chelsea},
  journal={arXiv preprint arXiv:2202.02005},
  year={2022}
}

@article{moveit2014,
  title={Reducing the Barrier to Entry of Complex Robotic Software: a {MoveIt!} Case Study},
  author={Coleman, David and Sucan, Ioan and Chitta, Sachin and Correll, Nikolaus},
  journal={arXiv preprint arXiv:1404.3785},
  year={2014}
}

@article{ompl2012,
  title={The Open Motion Planning Library},
  author={Sucan, Ioan A. and Moll, Mark and Kavraki, Lydia E.},
  journal={IEEE Robotics \& Automation Magazine},
  volume={19},
  number={4},
  pages={72--82},
  year={2012}
}

@inproceedings{kaelbling2011,
  title={Hierarchical Task and Motion Planning in the Now},
  author={Kaelbling, Leslie Pack and Lozano-P{\'e}rez, Tom{\'a}s},
  booktitle={2011 IEEE International Conference on Robotics and Automation},
  pages={1470--1477},
  year={2011}
}

@article{tampSurvey2024,
  title={A Survey of Optimization-based Task and Motion Planning: From Classical To Learning Approaches},
  author={Zhao, Zhigen and Cheng, Shuo and Ding, Yan and Zhou, Ziyi and Zhang, Shiqi and Xu, Danfei and Zhao, Ye},
  journal={arXiv preprint arXiv:2404.02817},
  year={2024}
}

@article{curobo2023,
  title={cuRobo: Parallelized Collision-Free Minimum-Jerk Robot Motion Generation},
  author={Sundaralingam, Balakumar and Hari, Siva Kumar Sastry and Fishman, Adam and Garrett, Caelan and Van Wyk, Karl and Blukis, Valts and Millane, Alexander and Oleynikova, Helen and Handa, Ankur and Ramos, Fabio and Ratliff, Nathan and Fox, Dieter},
  journal={arXiv preprint arXiv:2310.17274},
  year={2023}
}

@article{grade2023,
  title={{GRADE}: Generating Realistic And Dynamic Environments for Robotics Research with Isaac Sim},
  author={Bonetto, Elia and Xu, Chenghao and Ahmad, Aamir},
  journal={arXiv preprint arXiv:2303.04466},
  year={2023}
}

@article{blenderproc2019,
  title={BlenderProc},
  author={Denninger, Maximilian and Sundermeyer, Martin and Winkelbauer, Dominik and Zidan, Youssef and Olefir, Dmitry and Elbadrawy, Mohamad and Lodhi, Ahsan and Katam, Harinandan},
  journal={arXiv preprint arXiv:1911.01911},
  year={2019}
}

@article{kubric2022,
  title={Kubric: A Scalable Dataset Generator},
  author={Greff, Klaus and Belletti, Francois and Beyer, Lucas and Doersch, Carl and Du, Yilun and Duckworth, Daniel and Fleet, David J. and Gnanapragasam, Dan and Golemo, Florian and others},
  journal={arXiv preprint arXiv:2203.03570},
  year={2022}
}

@article{rt1_2022,
  title={{RT-1}: Robotics Transformer for Real-World Control at Scale},
  author={Brohan, Anthony and Brown, Noah and Carbajal, Justice and Chebotar, Yevgen and Dabis, Joseph and Finn, Chelsea and Gopalakrishnan, Keerthana and Hausman, Karol and Herzog, Alex and Hsu, Jasmine and others},
  journal={arXiv preprint arXiv:2212.06817},
  year={2022}
}

@article{rt2_2023,
  title={{RT-2}: Vision-Language-Action Models Transfer Web Knowledge to Robotic Control},
  author={Brohan, Anthony and Brown, Noah and Carbajal, Justice and Chebotar, Yevgen and Chen, Xi and Choromanski, Krzysztof and Ding, Tianli and Driess, Danny and others},
  journal={arXiv preprint arXiv:2307.15818},
  year={2023}
}

@article{palme2023,
  title={{PaLM-E}: An Embodied Multimodal Language Model},
  author={Driess, Danny and Xia, Fei and Sajjadi, Mehdi S. M. and Lynch, Corey and Chowdhery, Aakanksha and Ichter, Brian and Wahid, Ayzaan and Tompson, Jonathan and Vuong, Quan and Yu, Tianhe and others},
  journal={arXiv preprint arXiv:2303.03378},
  year={2023}
}

@article{saycan2022,
  title={Do As I Can, Not As I Say: Grounding Language in Robotic Affordances},
  author={Ahn, Michael and Brohan, Anthony and Brown, Noah and Chebotar, Yevgen and Cortes, Omar and David, Byron and Finn, Chelsea and Fu, Chuyuan and Gopalakrishnan, Keerthana and Hausman, Karol and others},
  journal={arXiv preprint arXiv:2204.01691},
  year={2022}
}

@article{innerMonologue2022,
  title={Inner Monologue: Embodied Reasoning through Planning with Language Models},
  author={Huang, Wenlong and Xia, Fei and Xiao, Ted and Chan, Harris and Liang, Jacky and Florence, Pete and Zeng, Andy and Tompson, Jonathan and Mordatch, Igor and Chebotar, Yevgen and others},
  journal={arXiv preprint arXiv:2207.05608},
  year={2022}
}

@article{alfred2019,
  title={{ALFRED}: A Benchmark for Interpreting Grounded Instructions for Everyday Tasks},
  author={Shridhar, Mohit and Thomason, Jesse and Gordon, Daniel and Bisk, Yonatan and Han, Winson and Mottaghi, Roozbeh and Zettlemoyer, Luke and Fox, Dieter},
  journal={arXiv preprint arXiv:1912.01734},
  year={2019}
}

@article{teach2021,
  title={{TEACh}: Task-driven Embodied Agents that Chat},
  author={Padmakumar, Aishwarya and Thomason, Jesse and Shrivastava, Ayush and Lange, Patrick and Narayan-Chen, Anjali and Gella, Spandana and Piramuthu, Robinson and Tur, Gokhan and Hakkani-Tur, Dilek},
  journal={arXiv preprint arXiv:2110.00534},
  year={2021}
}

@article{minedojo2022,
  title={MineDojo: Building Open-Ended Embodied Agents with Internet-Scale Knowledge},
  author={Fan, Linxi and Wang, Guanzhi and Jiang, Yunfan and Mandlekar, Ajay and Yang, Yuncong and Zhu, Haoyi and Tang, Andrew and Huang, De-An and Zhu, Yuke and Anandkumar, Anima},
  journal={arXiv preprint arXiv:2206.08853},
  year={2022}
}

@article{voyager2023,
  title={Voyager: An Open-Ended Embodied Agent with Large Language Models},
  author={Wang, Guanzhi and Xie, Yuqi and Jiang, Yunfan and Mandlekar, Ajay and Xiao, Chaowei and Zhu, Yuke and Fan, Linxi and Anandkumar, Anima},
  journal={arXiv preprint arXiv:2305.16291},
  year={2023}
}

@article{openvla2024,
  title={{OpenVLA}: An Open-Source Vision-Language-Action Model},
  author={Kim, Moo Jin and Pertsch, Karl and Karamcheti, Siddharth and Xiao, Ted and Balakrishna, Ashwin and Nair, Suraj and Rafailov, Rafael and Foster, Ethan and Lam, Grace and Sanketi, Pannag and others},
  journal={arXiv preprint arXiv:2406.09246},
  year={2024}
}

@article{octo2024,
  title={Octo: An Open-Source Generalist Robot Policy},
  author={Ghosh, Dibya and Walke, Homer and Pertsch, Karl and Black, Kevin and Mees, Oier and Dasari, Sudeep and Hejna, Joey and Kreiman, Tobias and Xu, Charles and Luo, Jianlan and others},
  journal={arXiv preprint arXiv:2405.12213},
  year={2024}
}

@article{galactic2023,
  title={Galactic: Scaling End-to-End Reinforcement Learning for Rearrangement at 100k Steps-Per-Second},
  author={Berges, Vincent-Pierre and Szot, Andrew and Chaplot, Devendra Singh and Gokaslan, Aaron and Mottaghi, Roozbeh and Batra, Dhruv and Undersander, Eric},
  journal={arXiv preprint arXiv:2306.07552},
  year={2023}
}

@article{rllib2020,
  title={{RLlib Flow}: Distributed Reinforcement Learning is a Dataflow Problem},
  author={Liang, Eric and Wu, Zhanghao and Luo, Michael and Mika, Sven and Gonzalez, Joseph E. and Stoica, Ion},
  journal={arXiv preprint arXiv:2011.12719},
  year={2020}
}

@article{mittal2023orbit,
  title   = {Orbit: A Unified Simulation Framework for Interactive Robot Learning Environments},
  author  = {Mittal, Mayank and Yu, Calvin and Yu, Qinxi and Liu, Jingzhou and Rudin, Nikita and Hoeller, David and Yuan, Jia Lin and Singh, Ritvik and Guo, Yunrong and Mazhar, Hammad and Mandlekar, Ajay and Babich, Buck and State, Gavriel and Hutter, Marco and Garg, Animesh},
  journal = {IEEE Robotics and Automation Letters},
  volume  = {8},
  number  = {6},
  pages   = {3740--3747},
  year    = {2023}
}

@inproceedings{tobin2017domain,
  title     = {Domain Randomization for Transferring Deep Neural Networks from Simulation to the Real World},
  author    = {Tobin, Josh and Fong, Rachel and Ray, Alex and Schneider, Jonas and Zaremba, Wojciech and Abbeel, Pieter},
  booktitle = {IEEE/RSJ International Conference on Intelligent Robots and Systems (IROS)},
  pages     = {23--30},
  year      = {2017}
}

@article{brockman2016openai,
  title   = {OpenAI Gym},
  author  = {Brockman, Greg and Cheung, Vicki and Pettersson, Ludwig and Schneider, Jonas and Schulman, John and Tang, Jie and Zaremba, Wojciech},
  journal = {arXiv preprint arXiv:1606.01540},
  year    = {2016}
}

@inproceedings{sheng2024hybridflow,
  title     = {HybridFlow: A Flexible and Efficient {RLHF} Framework},
  author    = {Sheng, Guangming and Zhang, Chi and Ye, Zilingfeng and Wu, Xibin and Zhang, Wang and Zhang, Ru and Peng, Yanghua and Lin, Haibin and Wu, Chuan},
  booktitle = {Proceedings of the Twentieth European Conference on Computer Systems (EuroSys)},
  year      = {2025},
  note      = {The verl library implements HybridFlow}
}

@article{yu2025rlinf,
  title   = {{RLinf}: Flexible and Efficient Large-scale Reinforcement Learning via Macro-to-Micro Flow Transformation},
  author  = {Yu, Chao and Wang, Yuanqing and Guo, Zhen and Lin, Hao and Xu, Si and Zang, Hongzhi and Zhang, Quanlu and Wu, Yongji and Zhu, Chunyang and Hu, Junhao and others},
  journal = {arXiv preprint arXiv:2509.15965},
  year    = {2025}
}

@article{fox1997dwa,
  title   = {The Dynamic Window Approach to Collision Avoidance},
  author  = {Fox, Dieter and Burgard, Wolfram and Thrun, Sebastian},
  journal = {IEEE Robotics \& Automation Magazine},
  volume  = {4},
  number  = {1},
  pages   = {23--33},
  year    = {1997}
}

@article{ben2025homie,
  title   = {{HOMIE}: Humanoid Loco-Manipulation with Isomorphic Exoskeleton Cockpit},
  author  = {Ben, Qingwei and Jia, Feiyu and Zeng, Jia and Dong, Junting and
             Lin, Dahua and Pang, Jiangmiao},
  journal = {arXiv preprint arXiv:2502.13013},
  year    = {2025}
}

@misc{zhao2026agilecomprehensiveworkflowhumanoid,
  title         = {AGILE: A Comprehensive Workflow for Humanoid Loco-Manipulation Learning},
  author        = {Huihua Zhao* and Rafael Cathomen* and Lionel Gulich and Wei Liu and
                   Efe Arda Ongan and Michael Lin and Shalin Jain and Soha Pouya and Yan Chang},
  year          = {2026},
  eprint        = {2603.20147},
  archivePrefix = {arXiv},
  primaryClass  = {cs.RO},
  url           = {https://arxiv.org/abs/2603.20147}
}

@misc{pink,
  title = {{Pink: Python inverse kinematics based on Pinocchio}},
  author = {Caron, Stéphane and De Mont-Marin, Yann and Budhiraja, Rohan and Bang, Seung Hyeon and Domrachev, Ivan and Nedelchev, Simeon and Du, Peter and Escande, Adrien and Vaillant, Joris and Wingo, Bruce and Patapati, Santosh and San José Pro, Daniel and Marticorena Vidal, Nicolas Guillermo},
  license = {Apache-2.0},
  url = {https://github.com/stephane-caron/pink},
  version = {4.2.0},
  year = {2026}
}

@inproceedings{zheng2024sglang,
  title     = {{SGLang}: Efficient Execution of Structured Language Model Programs},
  author    = {Zheng, Lianmin and Yin, Liangsheng and Xie, Zhiqiang and Sun, Chuyue and
               Huang, Jeff and Yu, Cody Hao and Cao, Shiyi and Kozyrakis, Christos and
               Stoica, Ion and Gonzalez, Joseph E. and Barrett, Clark and Sheng, Ying},
  booktitle = {Advances in Neural Information Processing Systems},
  year      = {2024}
}

@article{wang2025mindcube,
  title   = {{MindCube}: Spatial Mental Modeling from Limited Views},
  author  = {Wang, Qineng and Yin, Baiqiao and Zhang, Pingyue and Zhang, Jianshu and
             Wang, Kangrui and Wang, Zihan and Zhang, Jieyu and
             Chandrasegaran, Keshigeyan and Liu, Han and Krishna, Ranjay and
             Xie, Saining and Wu, Jiajun and Fei-Fei, Li and Li, Manling},
  journal = {arXiv preprint arXiv:2506.21458},
  year    = {2025}
}

@article{lu2026phys4d,
  title   = {{Phys4D}: Fine-Grained Physics-Consistent {4D} Modeling from Video Diffusion},
  author  = {Lu, Haoran and Wu, Shang and Zhang, Jianshu and Su, Maojiang and Ye, Guo and
             Xu, Chenwei and Lu, Lie and Maneriker, Pranav and Du, Fan and
             Li, Manling and Wang, Zhaoran and Liu, Han},
  journal = {arXiv preprint arXiv:2603.03485},
  year    = {2026}
}

@misc{scenesmith2026,
  title={SceneSmith: Agentic Generation of Simulation-Ready Indoor Scenes},
  author={Nicholas Pfaff and Thomas Cohn and Sergey Zakharov and Rick Cory and Russ Tedrake},
  year={2026},
  eprint={2602.09153},
  archivePrefix={arXiv},
  primaryClass={cs.RO},
  url={https://arxiv.org/abs/2602.09153}
}

@inproceedings{yang2024holodeck,
  title={Holodeck: Language Guided Generation of 3D Embodied AI Environments},
  author={Yang, Yue and Sun, Fan-Yun and Weihs, Luca and VanderBilt, Eli and Herrasti, Alvaro and Han, Winson and Wu, Jiajun and Haber, Nick and Krishna, Ranjay and Liu, Lingjie and Callison-Burch, Chris and Yatskar, Mark and Kembhavi, Aniruddha and Clark, Christopher},
  booktitle={Proceedings of the IEEE/CVF Conference on Computer Vision and Pattern Recognition (CVPR)},
  pages={16227--16237},
  year={2024}
}

@article{akinola2024tacsl,
  title   = {{TacSL}: A Library for Visuotactile Sensor Simulation and Learning},
  author  = {Akinola, Iretiayo and Xu, Jie and Carius, Jan and Fox, Dieter and Narang, Yashraj},
  journal = {arXiv preprint arXiv:2408.06506},
  year    = {2024}
}

@article{si2022taxim,
  title   = {{Taxim}: An Example-Based Simulation Model for {GelSight} Tactile Sensors},
  author  = {Si, Zilin and Yuan, Wenzhen},
  journal = {IEEE Robotics and Automation Letters},
  volume  = {7},
  number  = {2},
  pages   = {2361--2368},
  year    = {2022}
}

@article{huang2026flexitac,
  title   = {{FlexiTac}: A Low-Cost, Open-Source, Scalable Tactile Sensing Solution for Robotic Systems},
  author  = {Huang, Binghao and Li, Yunzhu},
  journal = {arXiv preprint arXiv:2604.28156},
  year    = {2026}
}

@article{su2026tacmap,
  title   = {{Tacmap}: Bridging the Tactile Sim-to-Real Gap via Geometry-Consistent Penetration Depth Map},
  author  = {Su, Lei and Peng, Zhijie and Ren, Renyuan and Mao, Shengping and Du, Juan and Zhang, Kaifeng and Zhu, Xuezhou},
  journal = {arXiv preprint arXiv:2602.21625},
  year    = {2026}
}

@article{yuan2017gelsight,
  title   = {{GelSight}: High-Resolution Robot Tactile Sensors for Estimating Geometry and Force},
  author  = {Yuan, Wenzhen and Dong, Siyuan and Adelson, Edward H.},
  journal = {Sensors},
  volume  = {17},
  number  = {12},
  pages   = {2762},
  year    = {2017},
  doi     = {10.3390/s17122762}
}

@misc{annotateanything2026,
  title        = {AnnotateAnything: Automatic Annotation of {3D} Assets for Robot Manipulation},
  author       = {{AnnotateAnything Team}},
  year         = {2026},
  note         = {Companion paper, under review. Citation to be updated upon publication.}
}

@article{bai2025qwen3vl,
  title        = {{Qwen3-VL} Technical Report},
  author       = {Bai, Shuai and Cai, Yuheng and Chen, Ruisheng and Chen, Kai and Chen, Xi and Cheng, Zesen and Deng, Lianghao and Ding, Wenyu and Gao, Chang and Ge, Chunjiang and others},
  journal      = {arXiv preprint arXiv:2511.21631},
  year         = {2025}
}

@misc{qwen2026qwen35,
  title        = {{Qwen3.5}: Towards Native Multimodal Agents},
  author       = {{Qwen Team}},
  year         = {2026},
  month        = {February},
  howpublished = {Official release post},
  url          = {https://www.alibabacloud.com/blog/602894},
  note         = {Accessed 2026-06-10}
}

@article{ma2025p3sam,
  title        = {{P3-SAM}: Native {3D} Part Segmentation},
  author       = {Ma, Changfeng and Li, Yang and Yan, Xinhao and Xu, Jiachen and Yang, Yunhan and Wang, Chunshi and Zhao, Zibo and Guo, Yanwen and Chen, Zhuo and Guo, Chunchao},
  journal      = {arXiv preprint arXiv:2509.06784},
  year         = {2025}
}

@article{yan2025xpart,
  title        = {{X-Part}: High Fidelity and Structure Coherent Shape Decomposition},
  author       = {Yan, Xinhao and Xu, Jiachen and Li, Yang and Ma, Changfeng and Yang, Yunhan and Wang, Chunshi and Zhao, Zibo and Lai, Zeqiang and Zhao, Yunfei and Chen, Zhuo and others},
  journal      = {arXiv preprint arXiv:2509.08643},
  year         = {2025}
}

@inproceedings{qi2017pointnetpp,
  title        = {{PointNet++}: Deep Hierarchical Feature Learning on Point Sets in a Metric Space},
  author       = {Qi, Charles Ruizhongtai and Yi, Li and Su, Hao and Guibas, Leonidas J.},
  booktitle    = {Advances in Neural Information Processing Systems (NeurIPS)},
  year         = {2017}
}

@misc{nvidia2025isaacsim,
  title        = {{NVIDIA Isaac Sim} Documentation},
  author       = {{NVIDIA}},
  year         = {2025},
  howpublished = {\url{https://docs.isaacsim.omniverse.nvidia.com}},
  note         = {Accessed 2026-06-10}
}

@inproceedings{rudin2022learning,
  title     = {Learning to Walk in Minutes Using Massively Parallel Deep
               Reinforcement Learning},
  author    = {Rudin, Nikita and Hoeller, David and Reist, Philipp and
               Hutter, Marco},
  booktitle = {Proceedings of the 5th Conference on Robot Learning (CoRL)},
  series    = {Proceedings of Machine Learning Research},
  volume    = {164},
  pages     = {91--100},
  publisher = {PMLR},
  year      = {2022}
}

@inproceedings{tao2025maniskill3,
  title     = {ManiSkill3: GPU Parallelized Robotics Simulation and
               Rendering for Generalizable Embodied AI},
  author    = {Tao, Stone and Xiang, Fanbo and Shukla, Arth and Qin, Yuzhe
               and Hinrichsen, Xander and Yuan, Xiaodi and Bao, Chen and
               Lin, Xinsong and Liu, Yulin and Chan, Tse-kai and Gao, Yuan
               and Li, Xuanlin and Mu, Tongzhou and Xiao, Nan and Gurha,
               Arnav and Rajesh, Viswesh Nagaswamy and Choi, Yong Woo and
               Chen, Yen-Ru and Huang, Zhiao and Calandra, Roberto and
               Chen, Rui and Luo, Shan and Su, Hao},
  booktitle = {Robotics: Science and Systems (RSS)},
  year      = {2025},
  note      = {arXiv:2410.00425}
}

@article{li2024behavior1k,
  title   = {BEHAVIOR-1K: A Human-Centered, Embodied AI Benchmark with
             1,000 Everyday Activities and Realistic Simulation},
  author  = {Li, Chengshu and Zhang, Ruohan and Wong, Josiah and Gokmen,
             Cem and Srivastava, Sanjana and Mart{\'i}n-Mart{\'i}n, Roberto
             and Wang, Chen and Levine, Gabrael and Ai, Wensi and Martinez,
             Benjamin and Yin, Hang and Lingelbach, Michael and Hwang,
             Minjune and Hiranaka, Ayano and Garlanka, Sujay and Aydin,
             Arman and Lee, Sharon and Sun, Jiankai and Anvari, Mona and
             Sharma, Manasi and Bansal, Dhruva and Hunter, Samuel and Kim,
             Kyu-Young and Lou, Alan and Matthews, Caleb R and
             Villa-Renteria, Ivan and Tang, Jerry Huayang and Tang, Claire
             and Xia, Fei and Li, Yunzhu and Savarese, Silvio and Gweon,
             Hyowon and Liu, C. Karen and Wu, Jiajun and Fei-Fei, Li},
  journal = {arXiv preprint arXiv:2403.09227},
  year    = {2024}
}

@article{puig2023habitat,
  title   = {Habitat 3.0: A Co-Habitat for Humans, Avatars and Robots},
  author  = {Puig, Xavier and Undersander, Eric and Szot, Andrew and Cote,
             Mikael Dallaire and Yang, Tsung-Yen and Partsey, Ruslan and
             Desai, Ruta and Clegg, Alexander William and Hlavac, Michal
             and Min, So Yeon and Vondru{\v s}, Vladim{\'i}r and Gervet,
             Theophile and Berges, Vincent-Pierre and Turner, John M. and
             Maksymets, Oleksandr and Kira, Zsolt and Kalakrishnan, Mrinal
             and Malik, Jitendra and Chaplot, Devendra Singh and Jain,
             Unnat and Batra, Dhruv and Rai, Akshara and Mottaghi, Roozbeh},
  journal = {arXiv preprint arXiv:2310.13724},
  year    = {2023}
}

@inproceedings{zhao2023aloha,
  title     = {Learning Fine-Grained Bimanual Manipulation with Low-Cost Hardware},
  author    = {Zhao, Tony Z. and Kumar, Vikash and Levine, Sergey and Finn, Chelsea},
  booktitle = {Robotics: Science and Systems (RSS)},
  year      = {2023},
  note      = {arXiv:2304.13705}
}

@inproceedings{mandlekar2023mimicgen,
  title     = {MimicGen: A Data Generation System for Scalable Robot Learning using Human Demonstrations},
  author    = {Mandlekar, Ajay and Nasiriany, Soroush and Wen, Bowen and Akinola, Iretiayo and Narang, Yashraj and Fan, Linxi and Zhu, Yuke and Fox, Dieter},
  booktitle = {Conference on Robot Learning (CoRL)},
  year      = {2023},
  note      = {arXiv:2310.17596}
}

@inproceedings{garrett2024skillmimicgen,
  title     = {SkillMimicGen: Automated Demonstration Generation for Efficient Skill Learning and Deployment},
  author    = {Garrett, Caelan and Mandlekar, Ajay and Wen, Bowen and Fox, Dieter},
  booktitle = {Conference on Robot Learning (CoRL)},
  year      = {2024},
  note      = {arXiv:2410.18907}
}

@inproceedings{jiang2024dexmimicgen,
  title     = {DexMimicGen: Automated Data Generation for Bimanual Dexterous Manipulation via Imitation Learning},
  author    = {Jiang, Zhenyu and Xie, Yuqi and Lin, Kevin and Xu, Zhenjia and Wan, Weikang and Mandlekar, Ajay and Fan, Linxi and Zhu, Yuke},
  booktitle = {IEEE International Conference on Robotics and Automation (ICRA)},
  year      = {2025},
  note      = {arXiv:2410.24185}
}

@article{moghani2026softmimicgen,
  title   = {SoftMimicGen: A Data Generation System for Scalable Robot Learning in Deformable Object Manipulation},
  author  = {Moghani, Masoud and Azizian, Mahdi and Garg, Animesh and Zhu, Yuke and Huver, Sean and Mandlekar, Ajay},
  journal = {arXiv preprint arXiv:2603.25725},
  year    = {2026}
}

@article{chen2025robotwin2,
  title   = {RoboTwin 2.0: A Scalable Data Generator and Benchmark with Strong Domain Randomization for Robust Bimanual Robotic Manipulation},
  author  = {Chen, Tianxing and Chen, Zanxin and Chen, Baijun and Cai, Zijian and Liu, Yibin and Liang, Qiwei and Li, Zixuan and Lin, Xianliang and Ge, Yiheng and Gu, Zhenyu and Deng, Weiliang and Guo, Yubin and Nian, Tian and Xie, Xuanbing and Chen, Qiangyu and Su, Kailun and Xu, Tianling and Liu, Guodong and Hu, Mengkang and Gao, Huan-ang and Wang, Kaixuan and Liang, Zhixuan and Qin, Yusen and Yang, Xiaokang and Luo, Ping and Mu, Yao},
  journal = {arXiv preprint arXiv:2506.18088},
  year    = {2025}
}

@inproceedings{jing2025humanoidgen,
  title     = {HumanoidGen: Data Generation for Bimanual Dexterous Manipulation via LLM Reasoning},
  author    = {Jing, Zhi and Yang, Siyuan and Ao, Jicong and Xiao, Ting and Jiang, Yugang and Bai, Chenjia},
  booktitle = {Advances in Neural Information Processing Systems (NeurIPS)},
  year      = {2025},
  note      = {arXiv:2507.00833}
}

@inproceedings{wang2023gensim,
  title     = {GenSim: Generating Robotic Simulation Tasks via Large Language Models},
  author    = {Wang, Lirui and Ling, Yiyang and Yuan, Zhecheng and Shridhar, Mohit and Bao, Chen and Qin, Yuzhe and Wang, Bailin and Xu, Huazhe and Wang, Xiaolong},
  booktitle = {International Conference on Learning Representations (ICLR)},
  year      = {2024},
  note      = {arXiv:2310.01361}
}

@inproceedings{hua2024gensim2,
  title     = {GenSim2: Scaling Robot Data Generation with Multi-modal and Reasoning LLMs},
  author    = {Hua, Pu and Liu, Minghuan and Macaluso, Annabella and Lin, Yunfeng and Zhang, Weinan and Xu, Huazhe and Wang, Lirui},
  booktitle = {Conference on Robot Learning (CoRL)},
  year      = {2024},
  note      = {arXiv:2410.03645}
}

@article{wang2023robogen,
  title   = {RoboGen: Towards Unleashing Infinite Data for Automated Robot Learning via Generative Simulation},
  author  = {Wang, Yufei and Xian, Zhou and Chen, Feng and Wang, Tsun-Hsuan and Wang, Yian and Fragkiadaki, Katerina and Erickson, Zackory and Held, David and Gan, Chuang},
  journal = {arXiv preprint arXiv:2311.01455},
  year    = {2023}
}

@article{garrett2021tamp,
  author  = {Garrett, Caelan Reed and Chitnis, Rohan and Holladay, Rachel
             and Kim, Beomjoon and Silver, Tom and Kaelbling, Leslie Pack
             and Lozano-P{\'e}rez, Tom{\'a}s},
  title   = {Integrated Task and Motion Planning},
  journal = {Annual Review of Control, Robotics, and Autonomous Systems},
  volume  = {4},
  pages   = {265--293},
  year    = {2021},
  note    = {arXiv:2010.01083}
}

@article{li2024hrcsafety,
  author  = {Li, Weidong and Hu, Yudie and Zhou, Yong and Pham, Duc Truong},
  title   = {Safe Human--Robot Collaboration for Industrial Settings:
             A Survey},
  journal = {Journal of Intelligent Manufacturing},
  volume  = {35},
  number  = {5},
  pages   = {2235--2261},
  year    = {2024},
  doi     = {10.1007/s10845-023-02159-4}
}

@inproceedings{sermanet2025asimov,
  author    = {Sermanet, Pierre and Majumdar, Anirudha and Irpan, Alex and
               Kalashnikov, Dmitry and Sindhwani, Vikas},
  title     = {Generating Robot Constitutions \& Benchmarks for Semantic
               Safety},
  booktitle = {Conference on Robot Learning (CoRL)},
  year      = {2025},
  note      = {arXiv:2503.08663}
}

@article{tom2024sdl,
  author  = {Tom, Gary and Schmid, Stefan P. and Baird, Sterling G. and
             Cao, Yang and Darvish, Kourosh and Hao, Han and Lo, Stanley
             and Pablo-Garc{\'i}a, Sergio and Rajaonson, Ella M. and
             Skreta, Marta and Yoshikawa, Naruki and Corapi, Samantha and
             Akkoc, Gun Deniz and Strieth-Kalthoff, Felix and Seifrid,
             Martin and Aspuru-Guzik, Al{\'a}n},
  title   = {Self-Driving Laboratories for Chemistry and Materials Science},
  journal = {Chemical Reviews},
  volume  = {124},
  number  = {16},
  pages   = {9633--9732},
  year    = {2024},
  doi     = {10.1021/acs.chemrev.4c00055}
}

@misc{bui2025pushanything,
      title={Push Anything: Single- and Multi-Object Pushing From First Sight with Contact-Implicit MPC}, 
      author={Hien Bui and Yufeiyang Gao and Haoran Yang and Eric Cui and Siddhant Mody and Brian Acosta and Thomas Stephen Felix and Bibit Bianchini and Michael Posa},
      year={2026},
      eprint={2510.19974},
      archivePrefix={arXiv},
      primaryClass={cs.RO},
      url={https://arxiv.org/abs/2510.19974}, 
}

@misc{chen2021inhand,
  title        = {Learning Dexterous In-Hand Manipulation},
  author       = {Chen, Tao and Xu, Jie and Agrawal, Pulkit},
  year         = {2021},
  note         = {Reference for learned dexterous in-hand manipulation policies}
}

@inproceedings{garmentlab,
 author = {Lu, Haoran and Wu, Ruihai and Li, Yitong and Li, Sijie and Zhu, Ziyu and Ning, Chuanruo and Shen, Yan and Luo, Longzan and Chen, Yuanpei and Dong, Hao},
 booktitle = {Advances in Neural Information Processing Systems},
 doi = {10.52202/079017-0379},
 editor = {A. Globerson and L. Mackey and D. Belgrave and A. Fan and U. Paquet and J. Tomczak and C. Zhang},
 pages = {11866--11903},
 publisher = {Curran Associates, Inc.},
 title = {GarmentLab: A Unified Simulation and Benchmark for Garment Manipulation},
 url = {https://proceedings.neurips.cc/paper_files/paper/2024/file/15f80ec0fed53885d2ca6272edb96ede-Paper-Conference.pdf},
 volume = {37},
 year = {2024}
}

@misc{UniGarmentAU,
  title        = {UniGarment: A Unified Simulation and Benchmark for Garment Manipulation},
  author       = {Haoran Lu and Yitong Li and Ruihai Wu and Chuanruo Ning and Yan Shen and Hao Dong},
  year         = {2025},
  note         = {Manuscript},
  url          = {https://api.semanticscholar.org/CorpusID:275782214}
}

@misc{zhang2026progresslmprogressreasoningvisionlanguage,
      title={PROGRESSLM: Towards Progress Reasoning in Vision-Language Models}, 
      author={Jianshu Zhang and Chengxuan Qian and Haosen Sun and Haoran Lu and Dingcheng Wang and Letian Xue and Han Liu},
      year={2026},
      eprint={2601.15224},
      archivePrefix={arXiv},
      primaryClass={cs.CV},
      url={https://arxiv.org/abs/2601.15224}, 
}

@misc{wu2022vatmartlearningvisualaction,
      title={VAT-Mart: Learning Visual Action Trajectory Proposals for Manipulating 3D ARTiculated Objects}, 
      author={Ruihai Wu and Yan Zhao and Kaichun Mo and Zizheng Guo and Yian Wang and Tianhao Wu and Qingnan Fan and Xuelin Chen and Leonidas Guibas and Hao Dong},
      year={2022},
      eprint={2106.14440},
      archivePrefix={arXiv},
      primaryClass={cs.CV},
      url={https://arxiv.org/abs/2106.14440}, 
}

@article{geng2025roboverse,
  title={Roboverse: Towards a unified platform, dataset and benchmark for scalable and generalizable robot learning},
  author={Geng, Haoran and Wang, Feishi and Wei, Songlin and Li, Yuyang and Wang, Bangjun and An, Boshi and Cheng, Charlie Tianyue and Lou, Haozhe and Li, Peihao and Wang, Yen-Jen and others},
  journal={arXiv preprint arXiv:2504.18904},
  year={2025}
}

@misc{zhang2026spacenumrevisitingspatialnumerical,
      title={SPACENUM: Revisiting Spatial Numerical Understanding in VLMs}, 
      author={Jianshu Zhang and Yijiang Li and Huifeixin Chen and Haoran Lu and Letian Xue and Bingyang Wang and Han Liu},
      year={2026},
      eprint={2605.23898},
      archivePrefix={arXiv},
      primaryClass={cs.AI},
      url={https://arxiv.org/abs/2605.23898}, 
}

@article{pan2025advevo_marl,
  title = {{AdvEvo-MARL}: Shaping Internalized Safety through Adversarial Co-Evolution in Multi-Agent Reinforcement Learning},
  author = {Pan, Zhenyu and Zhang, Yiting and Liu, Zhuo and Tang, Yolo Yunlong and Zhang, Zeliang and Luo, Haozheng and Han, Yuwei and Zhang, Jianshu and Wu, Dennis and Chen, Hong-Yu and Lu, Haoran and Fang, Haoyang and Li, Manling and Xu, Chenliang and Yu, Philip S. and Liu, Han},
  journal = {arXiv preprint arXiv:2510.01586},
  year = {2025},
  note = {Accepted to ICML 2026},
  eprint = {2510.01586},
  archivePrefix = {arXiv},
  primaryClass = {cs.AI},
  doi = {10.48550/arXiv.2510.01586},
  url = {https://arxiv.org/abs/2510.01586}
}

@article{pan2025fairreason,
  title = {{FairReason}: Balancing Reasoning and Social Bias in {MLLMs}},
  author = {Pan, Zhenyu and Zhang, Yutong and Zhang, Jianshu and Lu, Haoran and Luo, Haozheng and Han, Yuwei and Yu, Philip S. and Li, Manling and Liu, Han},
  journal = {arXiv preprint arXiv:2507.23067},
  year = {2025},
  note = {Accepted to the Trustworthy Foundation Models Workshop at ICCV 2025},
  eprint = {2507.23067},
  archivePrefix = {arXiv},
  primaryClass = {cs.AI},
  doi = {10.48550/arXiv.2507.23067},
  url = {https://arxiv.org/abs/2507.23067}
}

@inproceedings{chen2026pa3ff,
  title = {{PA3FF}: Learning Part-Aware Dense 3D Feature Field for Generalizable Articulated Object Manipulation},
  author = {Chen, Yue and Jiang, Muqing and Zheng, Kaifeng and Liang, Jiaqi and Tie, Chenrui and Lu, Haoran and Wu, Ruihai and Dong, Hao},
  booktitle = {The Fourteenth International Conference on Learning Representations},
  year = {2026},
  note = {ICLR 2026 Poster},
  eprint = {2602.14193},
  archivePrefix = {arXiv},
  primaryClass = {cs.RO},
  doi = {10.48550/arXiv.2602.14193},
  url = {https://openreview.net/forum?id=qXfRXfAHOK}
}

\end{document}